\documentclass{article} 
\usepackage{collas2022_conference,times}

\usepackage{amsmath,amsfonts,bm}

\def\eqref#1{equation~\ref{#1}}

\def\1{\bm{1}}

\DeclareMathAlphabet{\mathsfit}{\encodingdefault}{\sfdefault}{m}{sl}
\SetMathAlphabet{\mathsfit}{bold}{\encodingdefault}{\sfdefault}{bx}{n}

\usepackage{hyperref}
\hypersetup{
    colorlinks=true,
    linkcolor=red,
    filecolor=magenta,      
    urlcolor=blue,
    citecolor=purple,
    pdftitle={CORA},
    pdfpagemode=FullScreen,
    }

\usepackage[utf8]{inputenc} 
\usepackage[T1]{fontenc}    
\usepackage{hyperref}       
\usepackage{url}            
\usepackage{booktabs}       
\usepackage{amsfonts}       
\usepackage{nicefrac}       
\usepackage{microtype}      
\usepackage{xcolor}         
\usepackage{graphicx}
\usepackage{wrapfig}
\usepackage{subfig}
\usepackage{float}
\usepackage{amsmath,amsthm,mathtools}
\usepackage{enumitem}
\usepackage[table]{colortbl} 

\usepackage{array}
\newcolumntype{H}{>{\setbox0=\hbox\bgroup}c<{\egroup}@{}}

\newcommand{\brackets}[1]{\langle #1 \rangle}
\newcommand{\abs}[1]{\lvert #1 \rvert}

\newcommand{\tablemarginhack}{\centering\addtolength{\leftskip}{-2cm}\addtolength{\rightskip}{-2cm}}

\newcommand{\github}{
 \url{https://github.com/AGI-Labs/continual_rl}
}

\renewcommand*\cite[1]{\citep{#1}}

\title{CORA: Benchmarks, Baselines, and Metrics\\as a Platform for\\ Continual Reinforcement Learning Agents}

\author{Sam Powers\thanks{Equal contribution} \\
Carnegie Mellon University \\
\texttt{snpowers@cs.cmu.edu}
\And 
Eliot Xing\footnotemark[1] \\
Georgia Institute of Technology \\
\texttt{exing@gatech.edu}
\AND 
Eric Kolve \\
Allen Institute for AI
\And
Roozbeh Mottaghi \\
Allen Institute for AI
\And
Abhinav Gupta \\
Carnegie Mellon University
}

\collasfinalcopy 

\begin{document}

\maketitle

\begin{abstract}
Progress in continual reinforcement learning has been limited due to several barriers to entry: missing code, high compute requirements, and a lack of suitable benchmarks. In this work, we present CORA, a platform for \textbf{Co}ntinual \textbf{R}einforcement Learning \textbf{A}gents that provides benchmarks, baselines, and metrics in a single code package. The benchmarks we provide are designed to evaluate different aspects of the continual RL challenge, such as catastrophic forgetting, plasticity, ability to generalize, and sample-efficient learning. Three of the benchmarks utilize video game environments (Atari, Procgen, NetHack). The fourth benchmark, CHORES, consists of four different task sequences in a visually realistic home simulator, drawn from a diverse set of task and scene parameters. To compare continual RL methods on these benchmarks, we prepare three metrics in CORA: Continual Evaluation, Isolated Forgetting, and Zero-Shot Forward Transfer. Finally, CORA includes a set of performant, open-source baselines of existing algorithms for researchers to use and expand on. We release CORA and hope that the continual RL community can benefit from our contributions, to accelerate the development of new continual RL algorithms.
\end{abstract}

\section{Introduction}
Over the course of the last decade, reinforcement learning (RL) has developed into a promising tool for learning a large variety of tasks, such as robotic manipulation~\cite{kober2012reinforcement, kormushev2013reinforcement, deisenroth2013survey, parisi2015reinforcement}, embodied AI~\cite{zhu2017target, shridhar2020alfred, batra2020rearrangement, szot2021habitat}, video games~\cite{mnih2015human, vinyals2019grandmaster}, and board games like Chess, Go, or Shogi~\cite{silver2017amastering}. However, these advances can be attributed to fine-tuned agents each trained to solve the specific task. For example, a robot trained to hit a baseball would not be able to play table tennis, even though both tasks involve swinging at a ball. If we were to train the agent with current learning-based methods on a new task, then it would tend to forget previous tasks and skills.

By contrast, humans continuously learn, remembering many tasks and using past experiences to help learn new tasks in different environments over extended periods of time. Developing agents capable of continuously building on what was learned previously, without forgetting knowledge obtained from the past, is crucial in order to deploy robotic agents into everyday scenarios. This capability is referred to as \textit{continual learning}~\cite{ring1994continual, ring1998child}, also known as \textit{never-ending learning}~\cite{NELL-aaai15, Chen-2013-7815}, \textit{incremental learning}, and \textit{lifelong learning}~\cite{Thrun1995LifelongRL}. 

In recent years, there has been a growing interest in building agents that continuously learn skills or tasks without forgetting previously learned behavior~\cite{khetarpal2020_crlreview}. However, unlike other areas leveraging machine learning such as computer vision and natural language processing, growth in the field of continual RL is still quite limited. Why is that? We argue there are three primary reasons, (a) missing code: research code in continual RL is not publicly available (due to the use of proprietary code for training agents at scale) and reproducibility remains difficult. This creates a high barrier to entry as any new entrant must re-implement and tune baselines, in addition to designing their own algorithm; (b) high compute barrier: the lack of publicly released baselines is compounded by the fact that extensive compute resources are required to run experiments used in prior work, which is not easily available in academic settings; (c) benchmarking gaps: there are few benchmarks and metrics used to evaluate continual RL, with no set standards. 

With this work, we aim to democratize the field of continual RL by reducing the barriers to entry and enable more research groups to develop algorithms for continual RL. To this end, we introduce CORA, a platform that includes benchmarks, baselines, and metrics for \textbf{Co}ntinual \textbf{R}einforcement Learning \textbf{A}gents. With this platform, we present three contributions to the field which we believe will support progress towards this goal. First, we present a set of benchmarks, each tailored to measure progress toward a different goal of continual learning. Our benchmarks include task sequences designed to: test generalization to unseen environment contexts (Procgen), evaluate scalability to the number of tasks being learned (MiniHack), and exercise scalability in realistic settings (CHORES), in addition to a standard, proven benchmark (Atari). Second, we provide three metrics to compare key attributes of continual RL methods on these benchmarks: Continual Evaluation, Isolated Forgetting, and Zero-Shot Forward Transfer. Finally, we release open-source implementations of previously proposed continual RL algorithms in a shared codebase, including CLEAR~\cite{rolnick2018clear}, a state-of-the-art method. We demonstrate that while CLEAR outperforms other baselines in Atari and Procgen, there is still significant room for methods to improve on our benchmarks.


In all, we present the CORA platform\footnote{\github}, designed to be a modular and extensible code package which brings all of this together. We hope our platform will be a one-stop shop for developing new algorithms, comparing to existing baselines using provided metrics, and running methods on evaluations suitable for testing different aspects of continual RL agents. We believe our contributions will help researchers conveniently develop new continual RL methods and facilitate communicating research results to the community in a standard fashion. If continual robot learning can successfully utilize household benchmarks like CHORES, then household robots or robots in the workforce may not be far distant. 
\section{Related Work}

We discuss various environments and tasks used to benchmark reinforcement learning in Appendix~\ref{appendix:extended_related_work}.

\textbf{Evaluating continual reinforcement learning} While continual learning is most commonly addressed in the context of supervised learning for image classification such as in~\cite{ruvolo2013ella, lopez2017gradient, hsu2018re, javed2019meta, hsu2018re, mai2021_online_cl}, here we focus our discussion on continual reinforcement learning, and as such, simulation environments and tasks to benchmark RL agents. For an overview on continual learning applied to neural networks in general, we refer the reader to~\citet{parisi2019continual} and~\citet{mundt2020wholistic}.

With CORA, we introduce benchmarks designed to evaluate continual RL algorithms that can be used in more challenging, realistic scenarios. Continual RL for policies or robotic agents~\cite{lesort2020continual} is more nascent, although several benchmarks have been proposed. As mentioned above, continual RL has typically been evaluated on a sequence of Atari games~\cite{kirkpatrick2017ewc, schwarz2018progress, rolnick2018clear} and we leverage these prior results to validate our baselines. Other video game-like environments proposed to evaluate continual learning include StarCraft~\cite{schwarz2018towards} and VizDoom~\cite{lomonaco2020continual}. 

Procgen~\cite{cobbe2020procgen} and MiniHack~\cite{kuettler2020nethack, samvelyan2021minihack}, two of our other benchmarks, are procedurally-generated, like Jelly Bean World~\cite{platanios2020jelly} which is a procedurally-generated 2D gridworld proposed as a testbed for continual learning agents.~\citet{nekoei2021continuous} present \textit{Lifelong Hanabi} for continual learning in a multi-agent RL environemnt. Beyond game-like environments,~\citet{wolczyk2021continual} evaluate continual RL using task boundaries in a multi-task robot manipulation environment, while~\citet{khetarpal2018environments} discuss home simulations as a potentially suitable environment to benchmark continual RL. In this work we present CHORES in AI2-THOR~\cite{kolve2017ai2}: task sequences for an agent in a home simulation to evaluate continual RL methods in the visually realistic scenes offered. 




Our work is conceptually similar to \texttt{bsuite}~\cite{osband202S0_bsuite}, which curates a collection of toy, diagnostic experiments to evaluate different capabilities of a standard, non-continual RL agent. Concurrent with our work, Sequoia~\cite{normandin2021sequoia} introduces a software framework with baselines, metrics, and evaluations aimed at unifying research in continual supervised learning and continual reinforcement learning. While both are valuable benchmarking tools, they focus predominantly on simpler tasks like MNIST and CartPole. The most complex environments that Sequoia uses are Meta-World, with simple state-based manipulation tasks, and MonsterKong, composed of 8 hand-designed platformer levels. In contrast to both, CORA presents challenging task sequences for vision-based, procedurally-generated environments that evaluate generalization and scability for continual RL. Also concurrent with our work, Avalanche RL~\cite{lucchesi2022avalanche} introduces a library for continual RL, whose functionality will be merged into Avalanche~\cite{lomonaco2021avalanche}, is a popular library for continual learning. However, Avalanche RL does not present any experimental results on baseline methods. In this paper, we evaluate several continual RL methods across four different environments.

\section{Task Sequences for Benchmarking Continual RL}
\label{section:benchmarks}

The goal of continual reinforcement learning is to develop an agent that can learn a variety of different tasks in non-stationary settings. To this end, prior work has primarily focused on preventing catastrophic forgetting~\cite{kirkpatrick2017ewc, rolnick2018clear, schwarz2018progress} and maintaining plasticity~\cite{mermillod2013_stabilityplasticity} so that the agent can learn new tasks. While simple tasks are useful for debugging, skill on them does not necessarily translate to more complex tasks. We believe that the field has matured enough and is ready for more ambitious goals. In particular, we believe continual RL methods should address the following problems: (a) showing positive forward transfer by leveraging past experience; (b) generalizing to unseen environment contexts; (c) learning similar tasks through provided goal specifications; (d) improving sample efficiency; in addition to (e) mitigating catastrophic forgetting; and (f) maintaining plasticity.


While a single benchmarking environment that suitably deals with each of these features may be ideal, over the course of development we have found this to be impractical with the tools currently available. For example, visually-realistic, physics-based environments are generally not fast enough for the longer sequences of tasks that we use to test resilience to forgetting. Furthermore, it may be overbearing for new algorithms to sufficiently address every continual RL goal, whereas a modular set of evaluations allows for researchers to focus on areas to best highlight particular contributions of their new methods. 
Instead, we present four benchmarks which continual RL reseachers may utilize:

\begin{itemize}[itemsep=0.05em, topsep=-0.1em]
    \item Atari~\cite{bellemare2013arcade}, 6 task sequence: A standard, proven benchmark used by~\citet{schwarz2018progress} and~\citet{rolnick2018clear}, particularly to demonstrate resilience to catastrophic forgetting.
    \item Procgen~\cite{cobbe2020procgen}, 6 task sequence: Designed to test resilience to forgetting and in-distribution generalization to unseen contexts in procedurally-generated, visually-distinct environments.
    \item MiniHack~\cite{kuettler2020nethack}, 15 task sequence, based on NetHack~\cite{samvelyan2021minihack}: Designed to train agents on a long sequence of tasks in environments that are stochastic, procedurally-generated, and visually-similar, in order to demonstrate resilience to forgetting, maintenance of plasticity, forward transfer, and out-of-distribution generalization (extrapolation along different environment factors).
    \item 4 different CHORES, utilizing ALFRED~\cite{shridhar2020alfred} and AI2-THOR~\cite{kolve2017ai2}: Designed to test agents in a visually realistic domain where sample efficiency is key. Unlike other environments where different tasks may be easily identified visually, CHORES tasks explicitly provide a goal image. CHORES also present an opportunity to test forward transfer due to task similarity. For example, the ability to pick up a hand towel ideally should transfer from one bathroom to another.
\end{itemize}

We direct the reader to Appendix~\ref{appendix:background} for formalism and background on the continual RL setting, including more precise definitions of generalization for these benchmarks and how continual RL applies to these task sequences.

Task selection and ordering is still an open area of research~\cite{jiang2020_prioritizedlevelreplay}, and we did not tune task ordering. We use the Atari task sequence as demonstrated in prior work, as well as use the implicit ordering in which Procgen and MiniHack presented their tasks. Selection of tasks for CHORES was more involved, and is described in Appendix~\ref{sec:task_selection}. 

Our goals include reducing the compute costs of continual RL experiments for the new benchmarks, as compared to the Atari experiments. Indeed, we observed a speedup of 7x for MiniHack, 6x for Procgen, and 2x for CHORES; details are available in Appendix~\ref{sec:experiment_durations}.

\subsection{Atari tasks}
\label{sec:atari_tasks}

Building off the work of~\citet{kirkpatrick2017ewc} that evaluated a random set of ten Atari~\cite{bellemare2013arcade} games, recent work in continual reinforcement learning~\cite{schwarz2018progress, rolnick2018clear} evaluates continual learning on six Atari games: [0-SpaceInvaders, 1-Krull, 2-BeamRider, 3-Hero, 4-StarGunner, 5-MsPacman]. They train agents on each of the six tasks for 50M frames, cycling through the sequence 5 times, for a total of 1500M frames seen. This results in 250M frames per task, which is five times as many frames as is standard in the single task setting. The primary focus of algorithms that were developed and evaluated on this Atari task sequence was to reduce catastrophic forgetting. This setting is particularly suitable for catastrophic forgetting due to the lack of overlap between tasks, in regards to both observations and skills required. More details are given in Appendix~\ref{results:atari}.

Following~\cite{kirkpatrick2017ewc, schwarz2018progress, rolnick2018clear}, we use the original Atari settings, meaning that the games are deterministic. Modifications such as ``sticky actions''~\cite{machado2018revisiting} have become more standard to overcome simulator determinism, and are an option for increasing task difficulty in future work. In this work, we use Atari to validate our baseline implementations, preferring procedurally-generated environments (Procgen, MiniHack) to test generalization. We note that different Atari game modes may also be used to produce variation and assess generalization capability, as proposed by~\citet{farebrother2018generalization}.


\subsection{Procgen tasks}


We use Procgen~\cite{cobbe2020procgen} to define a new sequence of video game tasks, with the intention of replacing the Atari tasks used previously to evaluate continual learning methods. We choose Procgen because its procedural generation allows for evaluating generalization on unseen levels, unlike Atari. Like Atari however, the tasks are all visually distinct and the task sequence is well-suited to evaluating catastrophic forgetting. As with Atari, this is due to a general lack of overlap between tasks. From the full set of available Procgen environments, the specific set of tasks was chosen by~\citet{igl2021transient} to ensure the existence of a nontrivial generalization gap and to ensure generalization actually improves during training. 

Procgen is also significantly faster to run (our experiments take several days for Procgen vs. weeks on Atari). To improve sample efficiency and reduce compute costs, we use the easy distribution mode for these Procgen games. We use a sequence of six tasks [0-Climber, 1-Dodgeball, 2-Ninja, 3-Starpilot, 4-Bigfish, 5-Fruitbot], training for 5M frames on each task with 5 learning cycles. This results in 25M frames per task and 125M frames total. Note that we are not increasing the number of training frames per task compared to the original paper, unlike the Atari task sequence.

The observation space is (64, 64) RGB images and is not framestacked. The 15-dim action space is the same across Procgen tasks. As recommended in the original paper, we train the agent on 200 levels, while evaluation uses the full distribution of levels that Procgen can procedurally generate. What is randomized varies depending on the game environment, but covers textures, enemies, objects, and room layouts.  See Appendix~\ref{appendix:initial_obs}, Figure~\ref{fig:figure_crl_env_viz_procgen} for a visualization of the observations an agent may receive for each task in Procgen.

\subsection{MiniHack's NetHack tasks}



Most prior continual RL work evaluates on a relatively small number of tasks, but the recently introduced MiniHack~\cite{samvelyan2021minihack} environment is fast enough to enable scaling up. MiniHack is based on the NetHack Learning Environment~\cite{kuettler2020nethack}, a setting that is procedurally generated like Procgen and has stochastic dynamics (such as when attacking monsters). As with Procgen, the variation over which levels are randomized differs by environment, but includes objects, enemies, start \& goal locations, and room layouts. The larger number of tasks enables MiniHack to more extensively test an agent's ability to prevent forgetting and to maintain plasticity. Additionally, while the Procgen tasks are easy to tell apart visually, the MiniHack tasks use the same texture assets and are more challenging to distinguish. This makes task identification and boundary detection more difficult.  

To create the MiniHack task sequence, we define 15 (train, test) task pairs with a total of 27 different navigation-type tasks. The training environments are the easier versions, and we evaluate the agent on the harder environment variant. We select from the navigation-type tasks introduced by MiniHack, ordering by how MiniHack presents their tasks, and only omit the tasks that require episodic memory and deep exploration. We provide the full MiniHack task sequence we use in Appendix~\ref{appendix:full_minihack_sequence}. Three evaluation environments are each used twice, because each has two related training tasks, the impact of which we discuss further in Section~\ref{section:minihack_results}. 
MiniHack also provides skill acquisition tasks, which could be used in future work for an even more challenging task sequence. 


When reporting results on this task sequence in Figure~\ref{fig:minihack_results}, we use the training environment name to refer to each task. We use only the pixel-based input for the agent. MiniHack renders an (80, 80) RGB image which we zero-pad to (84, 84) for convenience. See Appendix~\ref{appendix:initial_obs}, Figure~\ref{fig:figure_crl_env_viz_procgen} for a visualization of observations an agent may receive for each task in MiniHack. All tasks share an 8-dim action space.

\subsection{CHORES benchmark suite using ALFRED and AI2-THOR}



\begin{figure}[h]
    \centering
    \includegraphics[trim=0 2em 0 1.5em, clip, width=1.0\textwidth]{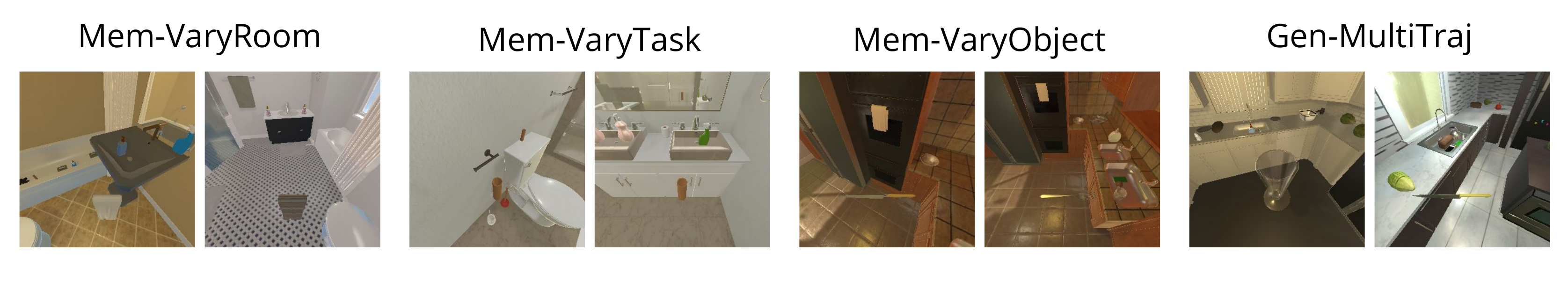}
    \caption{Examples for CHORES that show the variation within each task sequence.}
    \label{fig:ex_chores}
\end{figure}

AI2-THOR~\cite{kolve2017ai2} is a visually realistic simulation environment that provides a variety of rooms for an agent to act in, with 30 layouts each of bedrooms, living rooms, kitchens, and bathrooms. ALFRED~\cite{shridhar2020alfred} is a benchmark for embodied vision-and-language agents which provides demonstrations for extended sequences of complex, tool-based tasks defined using AI2-THOR.  

Using the demonstration trajectories and task definitions from ALFRED, we define a set of environments and task sequences for continual RL, which we refer to as \textbf{C}ontinual \textbf{H}ousehold \textbf{R}obot \textbf{E}nvironment \textbf{S}equences (CHORES). We do not provide ALFRED demonstration trajectories to the agent, as learning from demonstrations is beyond the scope of this paper. Instead, we leverage the demonstration data to initialize an AI2-THOR environment and generate subgoal images for the agent, which communicate the intended task for the agent to perform. The initial state of the environment is set to the initial state of the demonstration trajectory. The usage of these demonstrations enables us to have a variety of initializations for robot location, object locations, and room instance without explicitly setting the simulation parameters or hand-defining distributions over these parameters. ALFRED also defines reward functions for its tasks based on achieving its subgoals, which we use. Figure~\ref{fig:chores_subgoals} in Appendix~\ref{sec:chores_design_objectives} visualizes an example of a full set of CHORES subgoals. 




Our CHORES benchmark extends continual RL into a visually realistic domain, where sample efficiency is key and where tasks bear similarities that make forward transfer particularly useful. Sample efficiency is critical because we  designed CHORES to use a tight frame budget, as an initial attempt to mirror what would be feasible in the real world. 

We first define three CHORES that shift the environment context in well-defined ways: \textbf{Mem-VaryRoom} changes the room scene, \textbf{Mem-VaryTask} changes the task type, and \textbf{Mem-VaryObject} changes the object with which the agent interacts. The fourth CHORES, \textbf{Gen-VaryTraj}, is considerably harder than the first three: it varies both the object and the scene, in addition to testing generalization on unseen contexts from heldout demo trajectories. Figure~\ref{fig:ex_chores} visualizes CHORES and shows examples of variation within each task sequence. In Appendix~\ref{sec:chores_design_objectives}, we discuss design objectives and compute constraints used while creating the set of CHORES we introduce in this work. Appendix~\ref{sec:chores_experiment_details} describes the four CHORES proposed in more detail. We note that the CHORES protocol is not exclusive to AI2-THOR and can also be applied using any home simulation with a diverse dataset of demonstrations.

We use an action space of 12 discrete actions (e.g. LookDown, MoveAhead, SliceObject, PutObject, etc.). For an action that interacts with an object, we take the action with the correct task-relevant object. Note that this differs from agents evaluated in ALFRED originally, which generate interaction masks to select one object from those in view to interact with. We use an observation size of (64, 64, 6), with 3 channels for the current RGB image observation and 3 channels for an RGB goal image. 

\section{Metrics}
\label{section:metrics}

We refer the reader to Appendix~\ref{appendix:background} for background and full review of the continual RL setting. We assume $N$ tasks are presented as a sequence $\mathcal{S}_N:=(\mathcal{T}_0 \ldots \mathcal{T}_{N-1})$. The agent trains on task $\mathcal{T}_i$ at timesteps in the interval $[A_i, B_i)$, where $A_i$ and $B_i$ are the task boundaries denoting the start and end, respectively, of task $\mathcal{T}_i$. We cycle through the tasks $M$ times, so the full task sequence $\mathcal{S}_{NM}$ has length $N\cdot M$.
\\ \\
At each timestep $t$, the policy $\pi$ receives an observation, reward, and indicator of whether the episode is done, and takes an action based on what it has observed. In this section, we define episode return as the undiscounted sum of rewards received over an episode. We train the agent $s$ different times on the task sequence, with each run using a different initial random seed. We consider several expected episode returns to be used when defining our metrics:
\newcommand*\phantomrel[1]{\mathrel{\phantom{#1}}}
\begin{align}
    r_{i, \cdot, t=K}&:=\bold{R}_{t=K}(\pi, \mathcal{T}_i) & \text{expected return achieved on task $\mathcal{T}_i$ at timestep $K$}\\
    r_{i, j, end}&:=\bold{R}_{t=B_j}(\pi, \mathcal{T}_i) & \text{expected return achieved on task $\mathcal{T}_i$ after training on task $\mathcal{T}_j$}\\
    r_{i, all, max}&:=\max_{K\in[A_0, B_{N-1})} \bold{R}_{t=K}(\pi, \mathcal{T}_i) & \text{maximum (over all timesteps) expected return achieved on task $\mathcal{T}_i$}\\[-0.9em]
    & & \phantomrel{=} {} \textrm{after training on all tasks} \nonumber
\end{align}
Using these definitions, we discuss metrics for measuring different attributes of continual RL agents. Before proceeding, we describe how we estimate expected returns. A run involves training one instance of an agent on a task sequence. We pause each run every $n$ timesteps and evaluate $E=10$ episodes worth of data for every task in the sequence, and record the means. We further smooth the returns by using a moving average with a rolling window of size $w$ to get an estimate of $r$.


\subsection{Continual Evaluation, Isolated Forgetting, and Zero-Shot Forward Transfer}
\label{sec:metrics}

We use three metrics to evaluate the performance of an algorithm on our benchmarks. The first is the standard Continual Evaluation metric as used by prior work~\cite{rolnick2018clear, schwarz2018progress}. The second metric, Isolated Forgetting, measures how much an agent may forget an old task while learning a new task. The third metric, Zero-Shot Forward Transfer, measures how much an old task may contribute to the learning of a new task. 

Continual Evaluation presents the opportunity to compute our definitions of Forgetting and Transfer, which isolate the effects that training on task $\mathcal{T}_j$ has on the performance of task $\mathcal{T}_i$, without requiring single-task models to be trained and evaluated separately, as in~\cite{chaudhry2018_forgetting_intrans_metrics}. We present our Forgetting and Transfer metrics in two forms. The first form is the summary statistic, which provides a high-level overview of performance and is shown for our benchmarks in Table~\ref{tab:summary_metrics}. The second form is as a diagnostic table that describes how training on task $\mathcal{T}_j$ (column) impacts each task $\mathcal{T}_i$ (row). Appendix~\ref{section:metrics_tables} contains the full diagnostic tables for all benchmarks. 

\textbf{Continual Evaluation  ($\mathcal{C}$):} The Continual Evaluation metric, presented as a set of graphs, evaluates performance on all tasks periodically during training. This provides an understanding at any point of how the agent performs on every task $\mathcal{T}_i$ at timestep $K$. This is essentially a Monte Carlo estimate, $\mathcal{C}_{i}(t=K) \approx r_{i, \cdot, t=K}$. 


While the agent is asked to pause for evaluation every $n$ timesteps, in practice there is some variation due to the asynchronous implementation of the agents. To align the evaluation data across runs, we linearly interpolate to a common interval, then compute the mean and standard error over $s$ seeds. We graph $\mathcal{C}_i$ for each task; an example can be seen in Figure \ref{fig:procgen_results_res}. We also provide the final performance mean and standard error for each method in a table format, for convenient reference. 


\textbf{Isolated Forgetting ($\mathcal{F}$):} Isolated Forgetting, originally inspired by~\cite{lopez2017gradient, wang2021wanderlust}, represents how much is forgotten from a learned task during later tasks. It compares the expected return achieved for earlier task $\mathcal{T}_i$ before and after training on later task $\mathcal{T}_j$, where $i < j$:
\begin{gather}
    \mathcal{F}_{i, j} = \frac{ r_{i, j-1, end} - r_{i, j, end} }{ \abs{r_{i, all, max}} }
\end{gather}

When $\mathcal{F}_{i, j}>0$, the agent has become worse at past task $\mathcal{T}_i$ while training on new task $\mathcal{T}_j$, indicating forgetting has occurred. Conversely, when $\mathcal{F}_{i, j}<0$, the agent has become better at task $\mathcal{T}_i$, indicating \textit{backward transfer}~\cite{lopez2017gradient} has been observed. We normalize by the absolute value of the maximum expected return observed for task $\mathcal{T}_i$ within the run. Tasks can have varying reward scales, and normalization helps for comparing between tasks. 

Unlike~\citet{chaudhry2018_forgetting_intrans_metrics}, we do not use the max value observed, but rather the value right before training on a task, which corresponds with~\citet{lopez2017gradient}. This allows us to isolate the impact of that particular task, rather than looking at the cumulative effect to that point. We believe this makes the results easier to understand overall, and the metric more useful.





\textbf{Zero-Shot Forward Transfer ($\mathcal{Z}$):} Forward transfer considers how much prior tasks aid in the learning of new tasks. The Intransgience metric as defined by~\citet{chaudhry2018_forgetting_intrans_metrics} measures forward transfer by comparing the maximum expected return for a task trained independently to the expected return achieved while it was trained sequentially. While this might be the most accurate way to evaluate forward transfer, computing independent performance effectively doubles the amount of compute required. 

As one of our goals is to minimize the compute requirements of this benchmark, we instead propose what we refer to as the Zero-Shot Forward Transfer metric. 
It compares the expected return achieved for later task $\mathcal{T}_i$ before and after training on earlier task $\mathcal{T}_j$, where $i > j$: 
\begin{gather}
    \mathcal{Z}_{i, j} = \frac{ r_{i, j, end} - r_{i, j-1, end} }{ \abs{r_{i, all, max}} }
\end{gather}
When $\mathcal{Z}_{i, j}>0$, the agent has become better at later task $\mathcal{T}_i$ having trained on earlier task $\mathcal{T}_j$, indicating forward transfer has occurred by \textit{zero-shot} learning~\cite{lopez2017gradient, diaz2018don}. When $\mathcal{Z}_{i, j}<0$, the agent has become worse at task $i$, indicating negative transfer has occurred. We normalize by the absolute value of the maximum expected return observed for task $\mathcal{T}_i$ within the run, as was done for Forgetting.


%
%


\textbf{Additional details} (a) We only consider one cycle of the task sequence to compute $\mathcal{F}$ and $\mathcal{Z}$. We do this for interpretability and ease of understanding, though these metrics could be averaged across cycles with additional assumptions. (b) The diagnostic tables and summary statistics are given using the unseen testing environments when available. We use the same task indices when referring to the training and testing environments. (c) We scale by 10 for readability\footnote{For instance, ``4.2'' uses 1 less character than ``0.42'', so this scaling helps fit the metric tables horizontally in the paper.}, and average across seeds. (d) Summary statistics are computed as an average across tasks: $\overline{\mathcal{F}} =\sum_{i < j} \frac{10}{s} \sum_s \mathcal{F}_{i, j}$ and $\overline{\mathcal{Z}} =\sum_{i > j} \frac{10}{s} \sum_s \mathcal{Z}_{i, j}$. Additionally, we compute the standard error of the mean; details are given in Appendix~\ref{sec:compute_sem}. 





\textbf{Why use offline evaluation?} While a continual agent operating in the real world would not pause for evaluation, offline evaluation is a useful tool enabled by simulation to understand agent performance. Offline evaluation helps answer questions such as: ``How does the agent currently perform on tasks learned in the past?'', ``How does experience the agent has acquired help it learn new tasks?'', or ``How well would the agent generalize if it were asked to perform the task in a unseen environment?''.
%
\section{CORA: A Platform for Continual Reinforcement Learning Agents}
\subsection{Baselines}
\label{sec:baselines}

We re-implemented four continual RL methods with baseline results on the Atari sequence from Section~\ref{sec:atari_tasks}, which are not publicly available to the best of our knowledge.  We prioritized methods which had been demonstrated on Atari before, in order to reference such results and appropriately validate our implementations. The continual RL methods were selected to cover the categorizations described by~\citet{parisi2019continual, lesort2020continual, mundt2020wholistic}: elastic weight consolidation (EWC)~\cite{kirkpatrick2017ewc}, online EWC~\cite{schwarz2018progress}, Progress and Compress (P\&C)~\cite{schwarz2018progress}, Continual Learning with Experience and Replay (CLEAR)~\cite{rolnick2018clear}. EWC is a Regularization approach, P\&C is an Architectural approach, and CLEAR is a Rehearsal approach. Note that CLEAR is task-agnostic, while EWC and P\&C require explicit task boundaries. 

Our implementations of these baselines all build off the IMPALA~\cite{espeholt2018impala} architecture and use the open-source TorchBeast code~\cite{torchbeast2019}. We discuss implementation details in Appendix~\ref{appendix:implementation_differences}. In Appendix~\ref{results:atari}, we validate the performance of our baseline implementations compared to that of the original implementations on Atari.

\subsection{Code package}
\label{section:code_package}

We release our \texttt{continual\_rl} codebase\footnote{
Our code: \github
} as a convenient way to run continual RL baselines on the benchmarks we outlined in Section~\ref{section:benchmarks} and to use the continual RL evaluation metrics we defined in Section~\ref{section:metrics}. The package is designed modularly, so any component may be used separately elsewhere, and new benchmarks or algorithms may be integrated in. We provide more details on design and usage of the \texttt{continual\_rl} package in Appendix~\ref{sec:code_structure}. Hyperparameters for all experiments are made available as configuration files in the codebase, and also detailed in Appendix~\ref{appendix:hyperparameters}.

\setlength\tabcolsep{1.5pt} 
\begin{table}[H]
    \tiny
    \centering
    \subfloat[Forgetting $(\mathcal{F})$ summary statistics for all experiments.]{
    \begin{tabular}{lccccc}
              \toprule
              &  &  & Online & &  \\
              & IMPALA & EWC & EWC & P\&C & CLEAR \\
              \midrule
        Atari  & \cellcolor{red!9} 2.3 ± 0.1 & \cellcolor{red!1} 0.3 ± 0.3 & \cellcolor{red!6} 1.6 ± 0.1 & \cellcolor{red!7} 1.8 ± 0.1 & \cellcolor{red!2} 0.7 ± 0.1\\
        Procgen & \cellcolor{red!4} 1.2 ± 0.0 & \cellcolor{red!2} 0.7 ± 0.0 & \cellcolor{red!4} 1.1 ± 0.0 & \cellcolor{red!2} 0.5 ± 0.0 & \cellcolor{red!0} -0.0 ± 0.0 \\
        MiniHack & \cellcolor{red!1} 0.3 ± 0.0 &  - & - & - & \cellcolor{red!0} 0.1 ± 0.0 \\
        C-VaryRoom & - & \cellcolor{green!2} -0.6 ± 0.6 & - & \cellcolor{red!0} 0.0 ± 0.0 & \cellcolor{green!6} -1.7 ± 1.7 \\
        C-VaryTask & - & \cellcolor{red!8} 2.1 ± 1.4 & - & \cellcolor{green!8} -2.2 ± 2.2 & \cellcolor{red!6} 1.5 ± 0.2 \\
        C-VaryObj & - & \cellcolor{green!4} -1.0 ± 1.1 & - & \cellcolor{red!8} 2.0 ± 2.0 & \cellcolor{red!13} 3.4 ± 0.2 \\
        C-MultiTraj & - & \cellcolor{red!2} 0.7 ± 1.2 & - & \cellcolor{green!0} -0.1 ± 0.1 & \cellcolor{green!1} -0.3 ± 2.1 \\
        \bottomrule
    \end{tabular}
    }
    \hspace{3em}
    \subfloat[Transfer $(\mathcal{Z})$ summary statistics for all experiments.]{
    \begin{tabular}{lccccc}
              \toprule
              &  &  & Online & &  \\
              & IMPALA & EWC & EWC & P\&C & CLEAR \\
              \midrule
        Atari  &  \cellcolor{green!0} 0.1 ± 0.0 &  \cellcolor{green!0} 0.1 ± 0.2 &  \cellcolor{red!0} -0.0 ± 0.0 & \cellcolor{red!0} -0.0 ± 0.1 & \cellcolor{red!0} 0.0 ± 0.0 \\
        Procgen & \cellcolor{red!0} -0.1 ± 0.0 & \cellcolor{red!0} -0.2 ± 0.1 & \cellcolor{red!0} -0.1 ± 0.1 & \cellcolor{green!0} 0.1 ± 0.1 & \cellcolor{red!0} -0.1 ± 0.0 \\
        MiniHack & \cellcolor{green!2} 0.6 ± 0.0 & - & - & - & \cellcolor{green!2} 0.5 ± 0.1 \\
        C-VaryRoom & - & \cellcolor{red!0} -0.0 ± 0.0 & - & \cellcolor{green!12} 3.2 ± 1.9 & \cellcolor{red!4} -1.1 ± 1.1 \\
        C-VaryTask & - & \cellcolor{red!16} -4.0 ± 2.6 & - & \cellcolor{green!0} 0.2 ± 0.1 & \cellcolor{red!12} -3.2 ± 0.0 \\
        C-VaryObj & - & \cellcolor{green!10} 2.6 ± 2.9 & - & \cellcolor{green!21} 5.4 ± 1.3 & \cellcolor{red!18} -4.6 ± 0.8 \\
        C-MultiTraj & - & \cellcolor{red!16} -4.0 ± 0.5 & - & \cellcolor{green!1} 0.4 ± 0.1 & \cellcolor{red!18} -4.7 ± 0.7 \\
        \bottomrule
    \end{tabular}
    }
    \caption{Summary statistics for all benchmarks and for all methods evaluated on them.}
    \label{tab:summary_metrics}
\end{table}

\section{Experimental Results}
\label{section:experimental_results}

In this section, we present results on Procgen (Section~\ref{section:procgen_results}), MiniHack (Section~\ref{section:minihack_results}), and CHORES (Section~\ref{section:chores_results}). \textbf{For Atari results, see Appendix~\ref{results:atari}}. Metric summary statistics for all methods can be seen in Table~\ref{tab:summary_metrics}, and metric diagnostic tables are available in Section~\ref{section:metrics_tables}. Final performance tables are also available in Appendix~\ref{sec:final_perf_tables}, Tables~\ref{tab:final_perf_atari_train} through~\ref{tab:final_perf_chores}. To estimate expected return and compute metrics, we use the following values for parameters described in Section~\ref{section:metrics}: Procgen: $n=0.25e6$, $w=20$, $s=20$; MiniHack: $n=1e6$, $w=20$, $s=10$, CHORES: $n=5e4$, $w=5$, $s=3$, Atari: $n=0.25e6$, $w=20$, $s=5$. 

On the Continual Evaluation plots, solid lines represent evaluation on unseen testing environments, while dashed lines show evaluation on the training environments. Shaded grey rectangles are used to indicate which task is being trained during the indicated interval. We plot the mean as each line and the standard error as the surrounding, shaded region.

In each Forgetting table, we show negative values (representing backwards transfer) in green and positive in red, darker in proportion to the magnitude of $\mathcal{F}$. Values close to zero are unshaded. In contrast, in each Transfer table, we show positive values (indicating forward transfer) in green shades and negative values in red.

We proceed to discuss experimental results with CORA using two perspectives. First, from the viewpoint of benchmark analysis, we empirically discuss what each benchmark is evaluating and give examples of how the metric tables may be used. Second, from the view of algorithm design, we examine the performance of the baselines to identify axes which can be improved on by future algorithms. We frame this section through these two lenses in order to show how CORA may be used by end-users.

\subsection{Procgen results}
\label{section:procgen_results}

\begin{figure}[h]
    \centering
    \hspace{2.5em} 
    \includegraphics[trim=0 4em 20em 0, clip, width=0.25\textwidth]{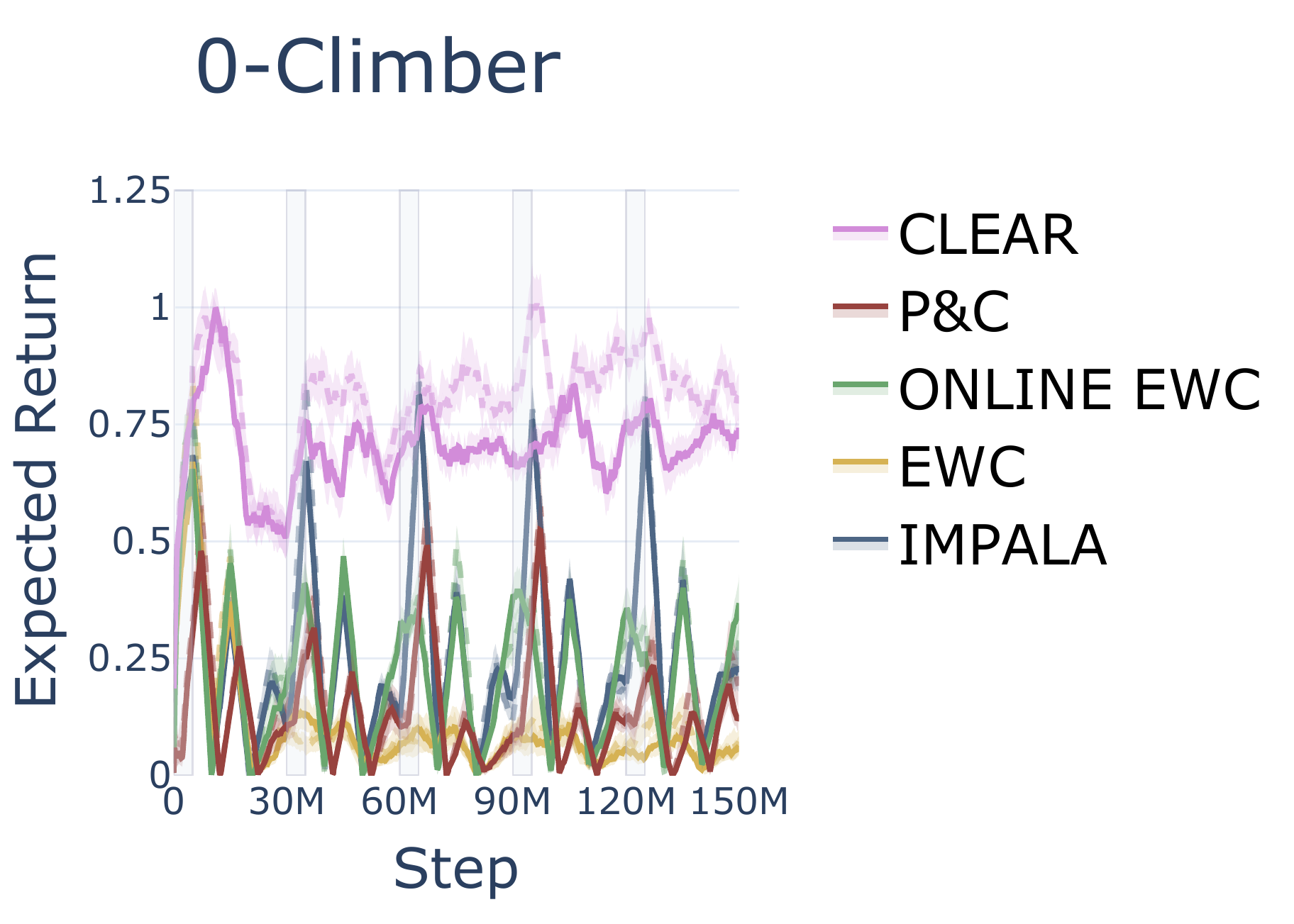}
    \includegraphics[trim=0 4em 20em 0, clip, width=0.25\textwidth]{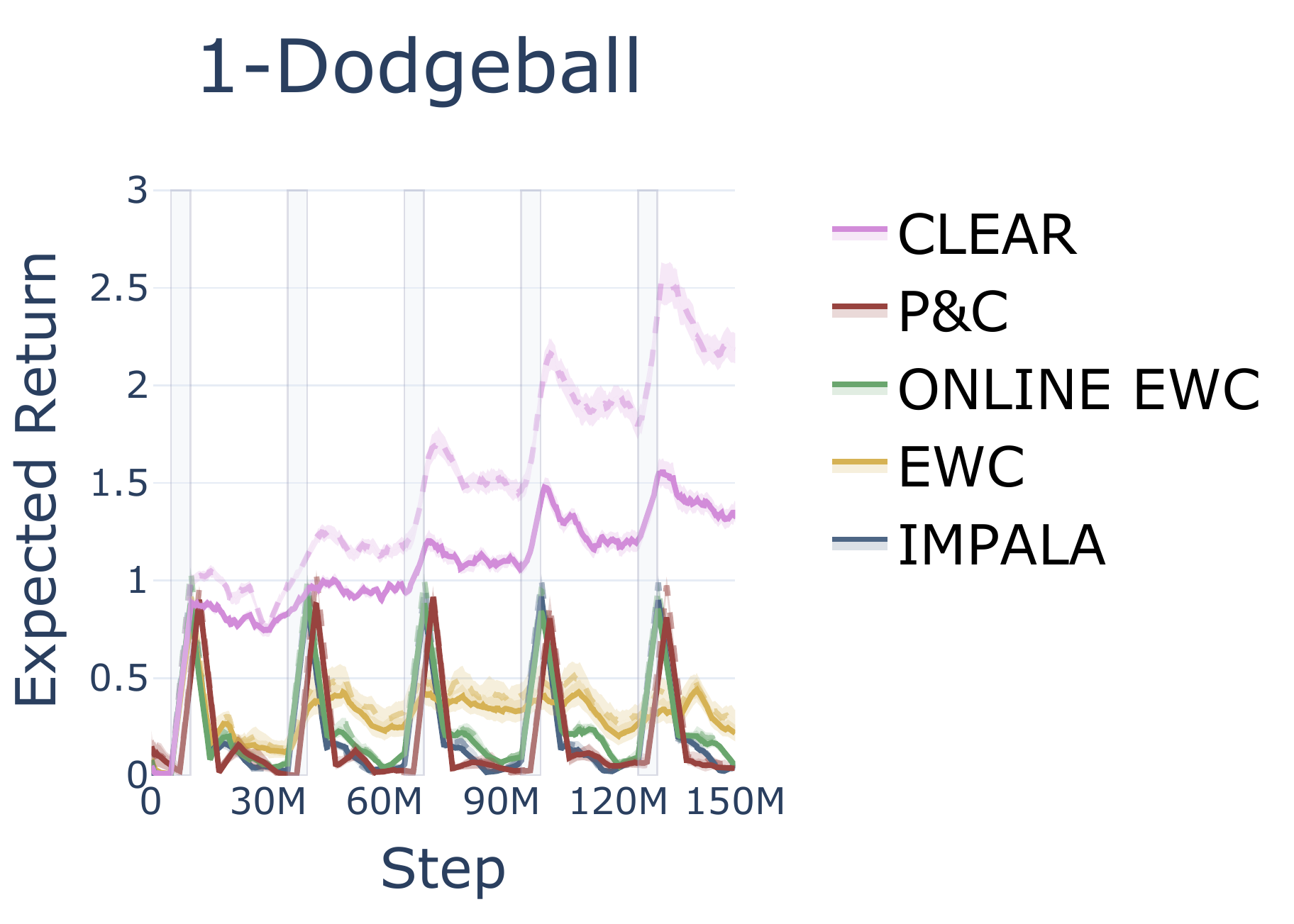}
    \includegraphics[trim=0 4em 0em 0, clip, width=0.40\textwidth]{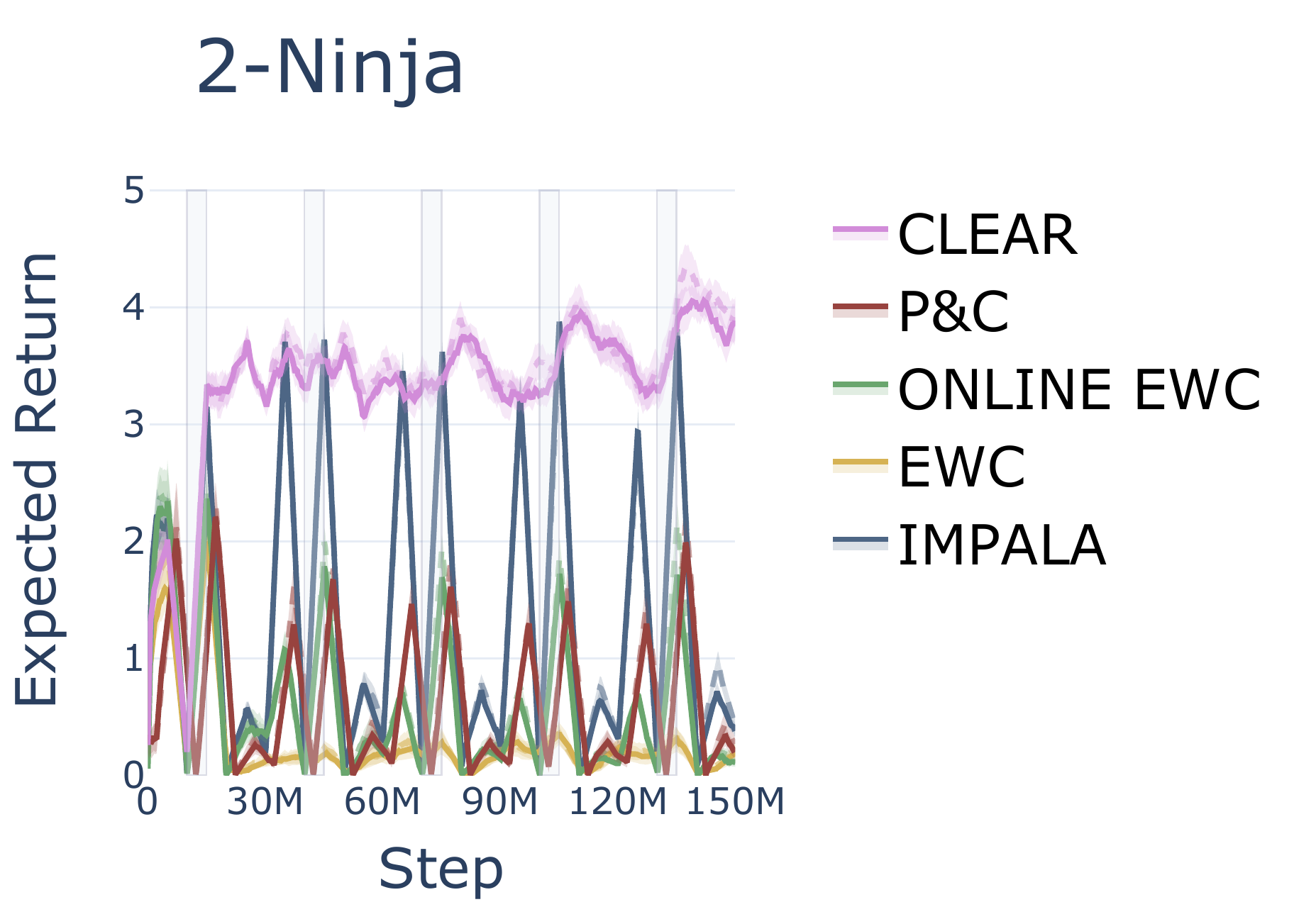} \\
    \includegraphics[trim=0 0em 20em 0, clip, width=0.25\textwidth]{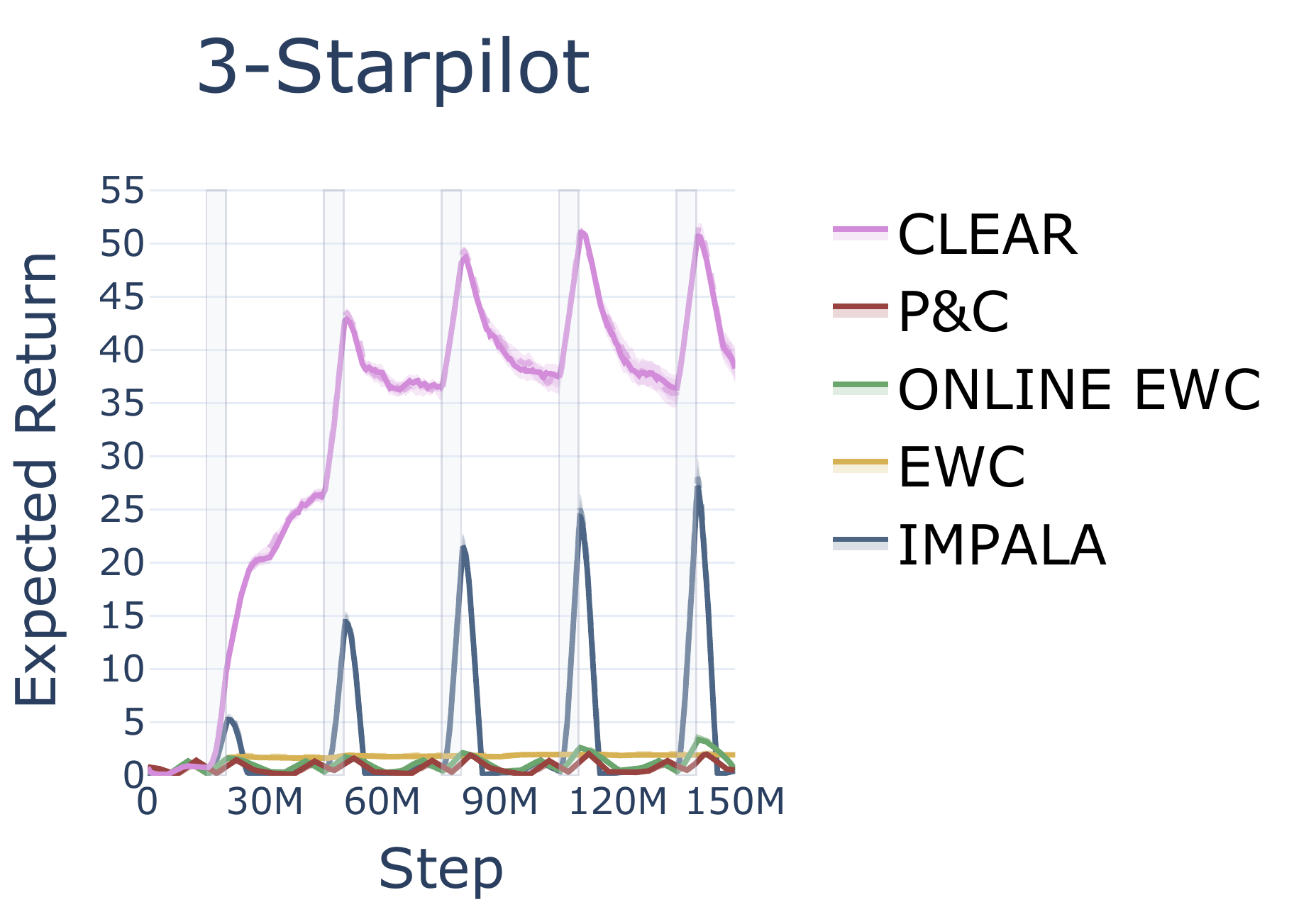}
    \includegraphics[trim=0 0em 20em 0, clip, width=0.25\textwidth]{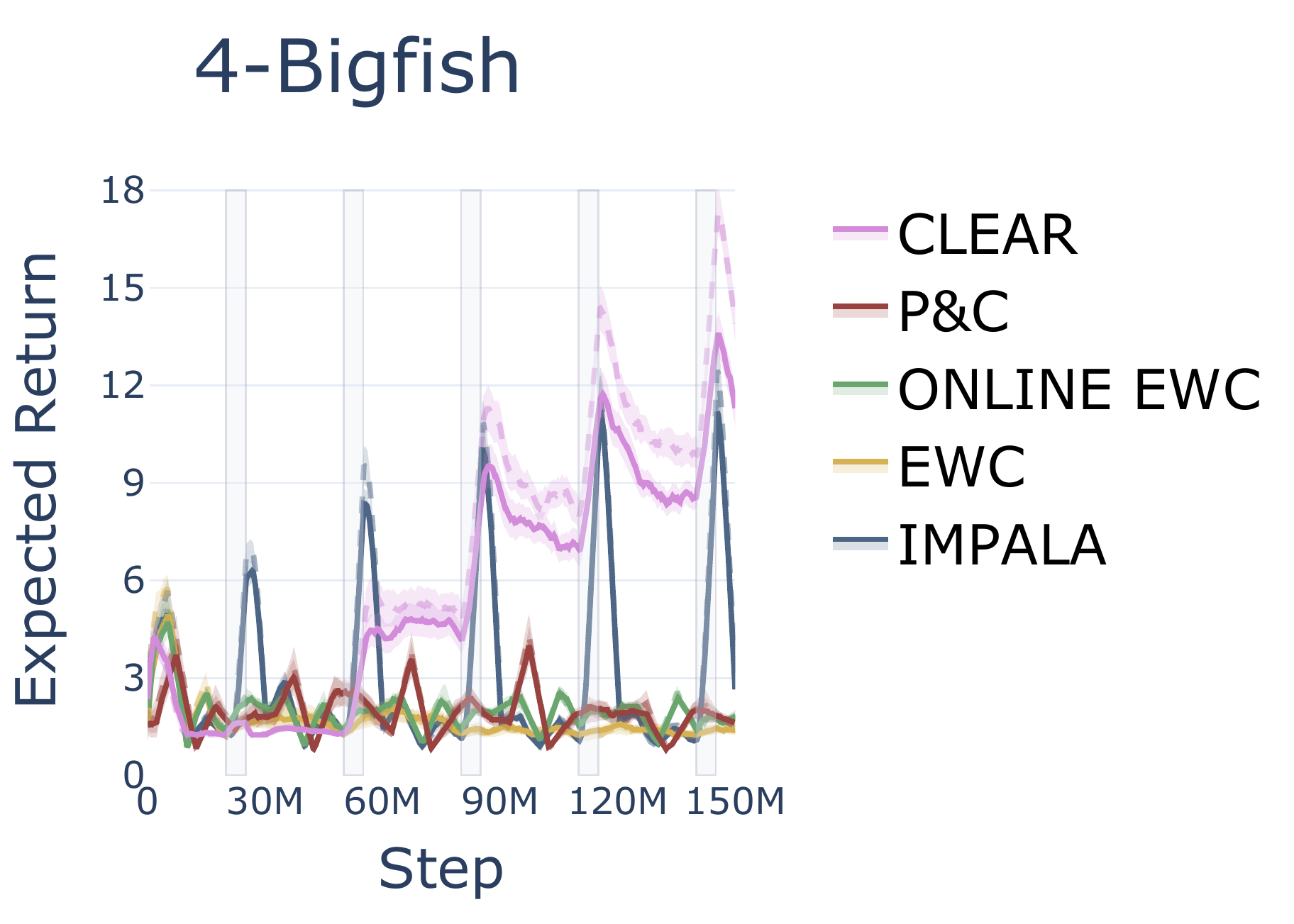}
    \includegraphics[trim=0 0em 20em 0, clip, width=0.25\textwidth]{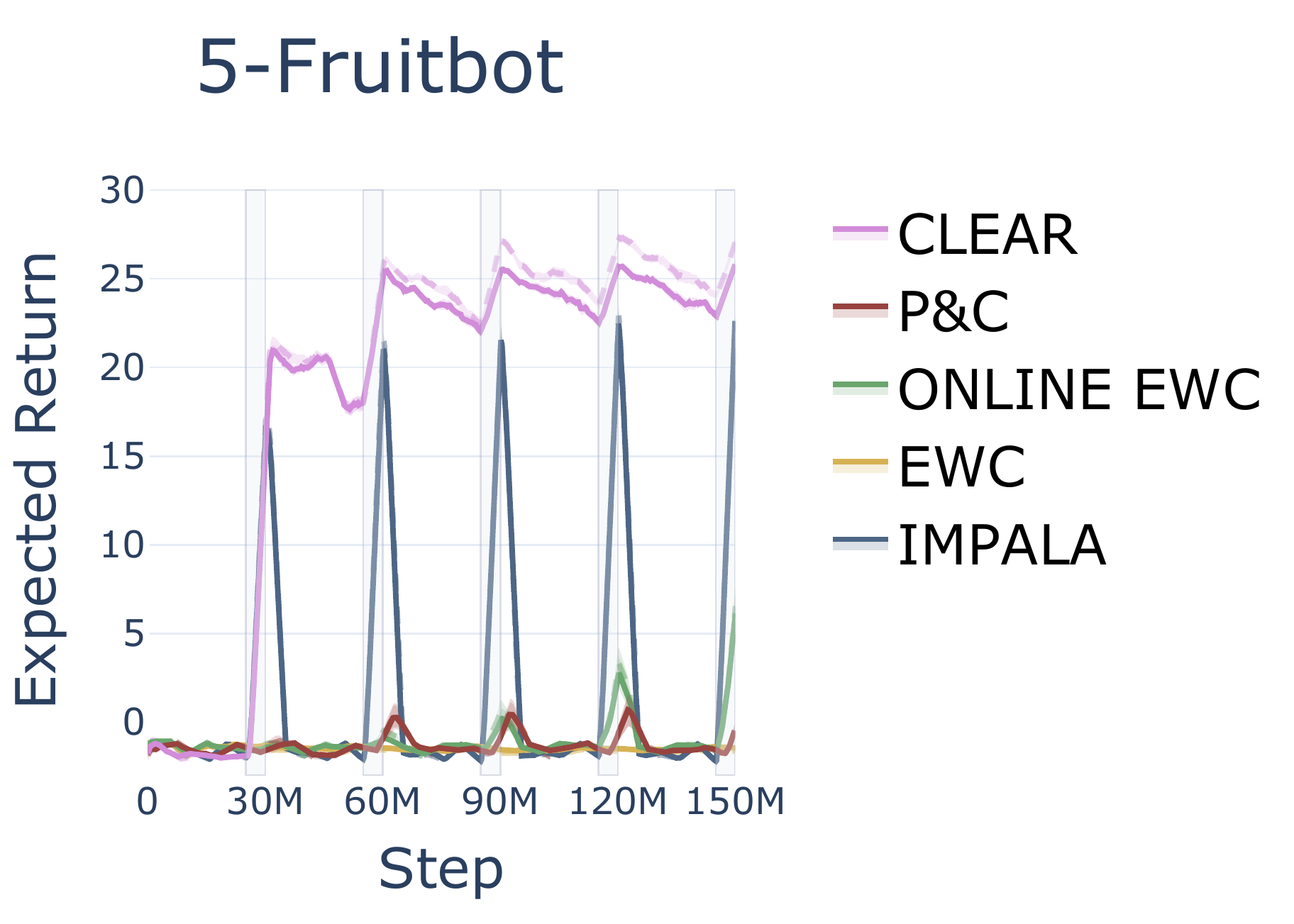}
    \hspace{4em}
    \caption{Results for Continual Evaluation $(\mathcal{C})$ on the 6 Procgen tasks, based on recommendations by~\cite{igl2021transient}. The solid line shows evaluation on unseen testing environments; the dashed line shows evaluation on training environments. Gray shaded rectangles show when the agent trains on each task. For readability, Figure~\ref{fig:procgen_results_noclear} in Appendix~\ref{appendix:old_procgen_results} presents an alternative version of this figure without CLEAR results.}
    \label{fig:procgen_results_res}
\end{figure}

\textbf{Benchmark analysis:} From the summary statistics in Table~\ref{tab:summary_metrics}, we can see that Procgen tests for catastrophic forgetting, but shows little forward transfer overall, which aligns with our expectations for this benchmark. Using the Transfer metric diagnostic tables in Appendix~\ref{sec:procgen_transfer}, we see that that forward transfer is not uniform across tasks. For example, we observe that 0-Climber transfers reasonably well to 2-Ninja and 4-Bigfish. Intuitively, as 0-Climber and 2-Ninja are both platformer games, transfer is expected. Transfer to 4-Bigfish is less obvious but may be explained by both games using side-view perspectives or by sharing useful skills like object gathering. In particular, 0-Climber involves collecting stars, while 4-Bigfish tasks the agent with eating other fish. 

\textbf{Algorithm design:} From the Continual Evaluation results in Figure~\ref{fig:procgen_results_res}, we observe that CLEAR is a strong baseline for avoiding catastrophic forgetting on all tasks, reliably outperforming every other method. However, there is still room for improvement: maximum scores obtained by CLEAR fall significantly short of the maximum achievable scores reported in Appendix C of the Procgen paper~\cite{cobbe2020procgen}, particularly on 0-Climber (1 vs 12.6), 1-Dodgeball (2.5 vs 19), 2-Ninja (4 vs 10), and 4-Bigfish (18 vs 40). 
Additionally, by comparing the training (dashed) and testing (solid) lines, we observe that CLEAR generalizes well to unseen contexts on all tasks, except 1-Dodgeball. The summary statistics in Table~\ref{tab:summary_metrics} show that transfer is overall low for Procgen. Using the more detailed diagnostic tables in Appendix~\ref{sec:procgen_transfer}, we can see that this varies by task. For instance with EWC, training on 1-Dodgeball improves performance on 3-Starpilot but reduces performance on all other tasks. CLEAR shows some transfer from 0-Climber to 2-Ninja, but essentially none anywhere else, even showing negative transfer from 1-Dodgeball to 2-Ninja. These failures represent opportunities for investigation and for new algorithms to improve on.

\subsection{MiniHack results}
\label{section:minihack_results}

\begin{figure}[h]
    \centering
    \hspace{2.25em} 
    \includegraphics[trim=0 3.5em 13em 0, clip, width=0.17\textwidth]{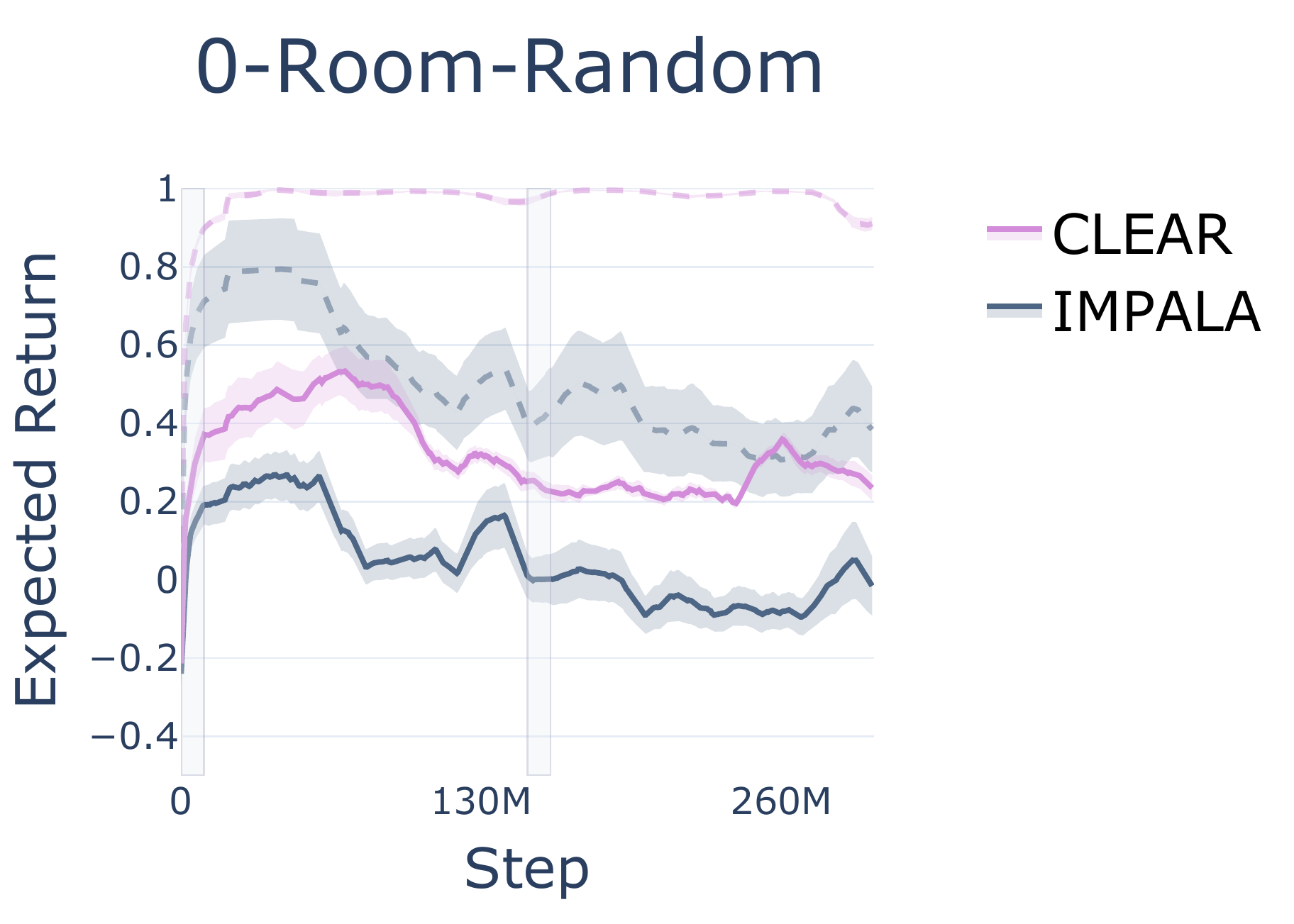}
    \includegraphics[trim=0 3.5em 13em 0, clip, width=0.17\textwidth]{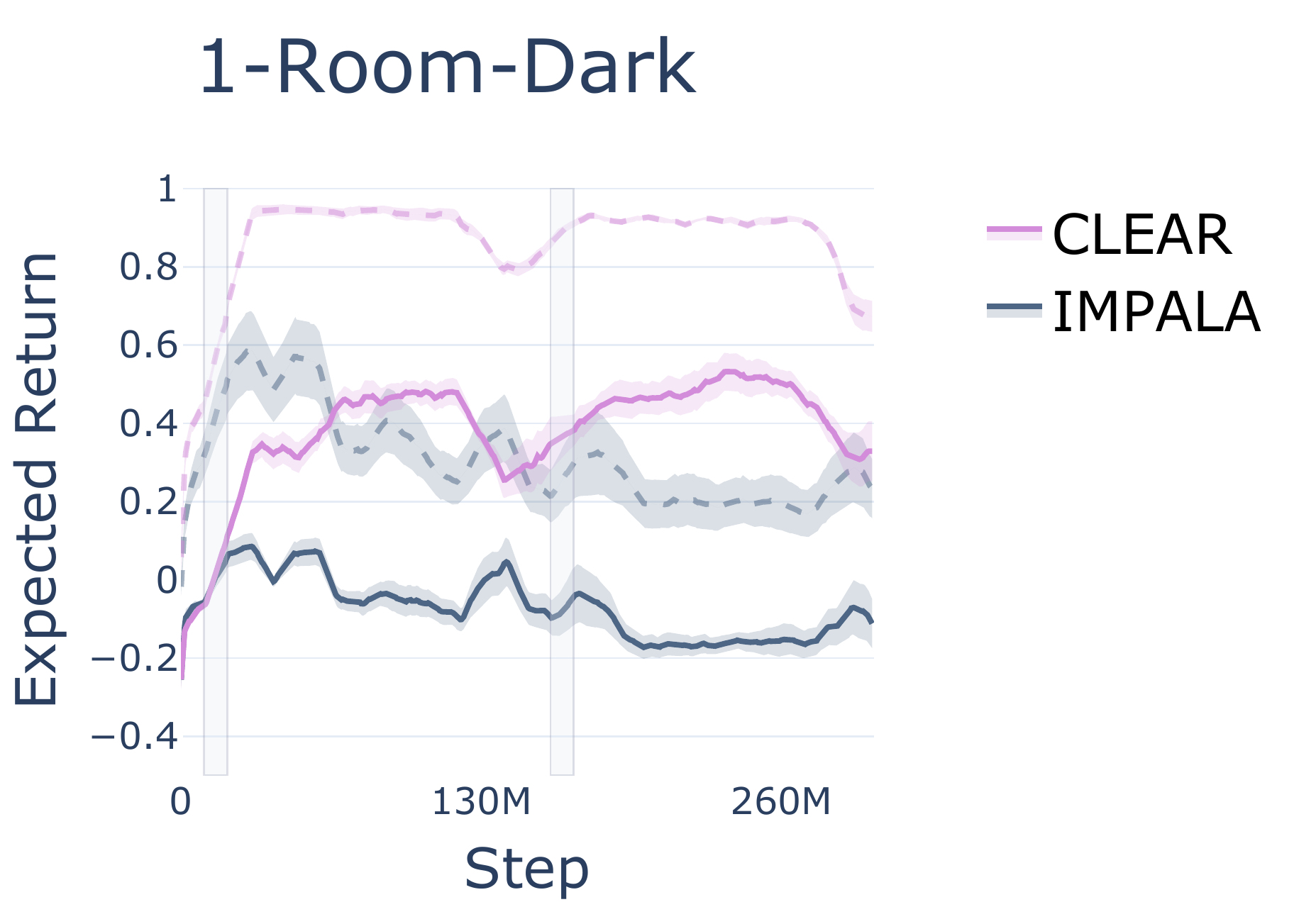}
    \includegraphics[trim=0 3.5em 13em 0, clip, width=0.17\textwidth]{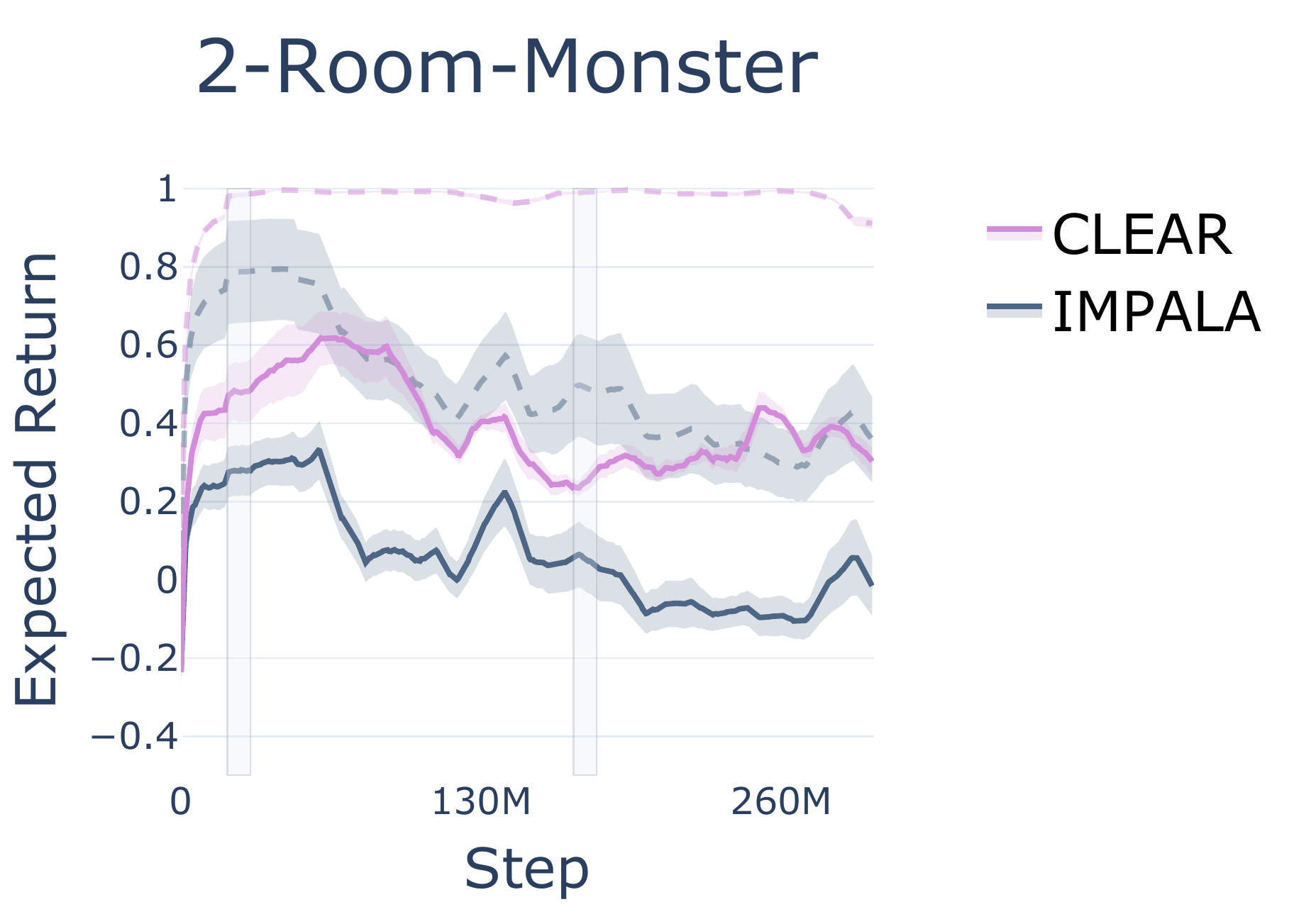}
    \includegraphics[trim=0 3.5em 13em 0, clip, width=0.17\textwidth]{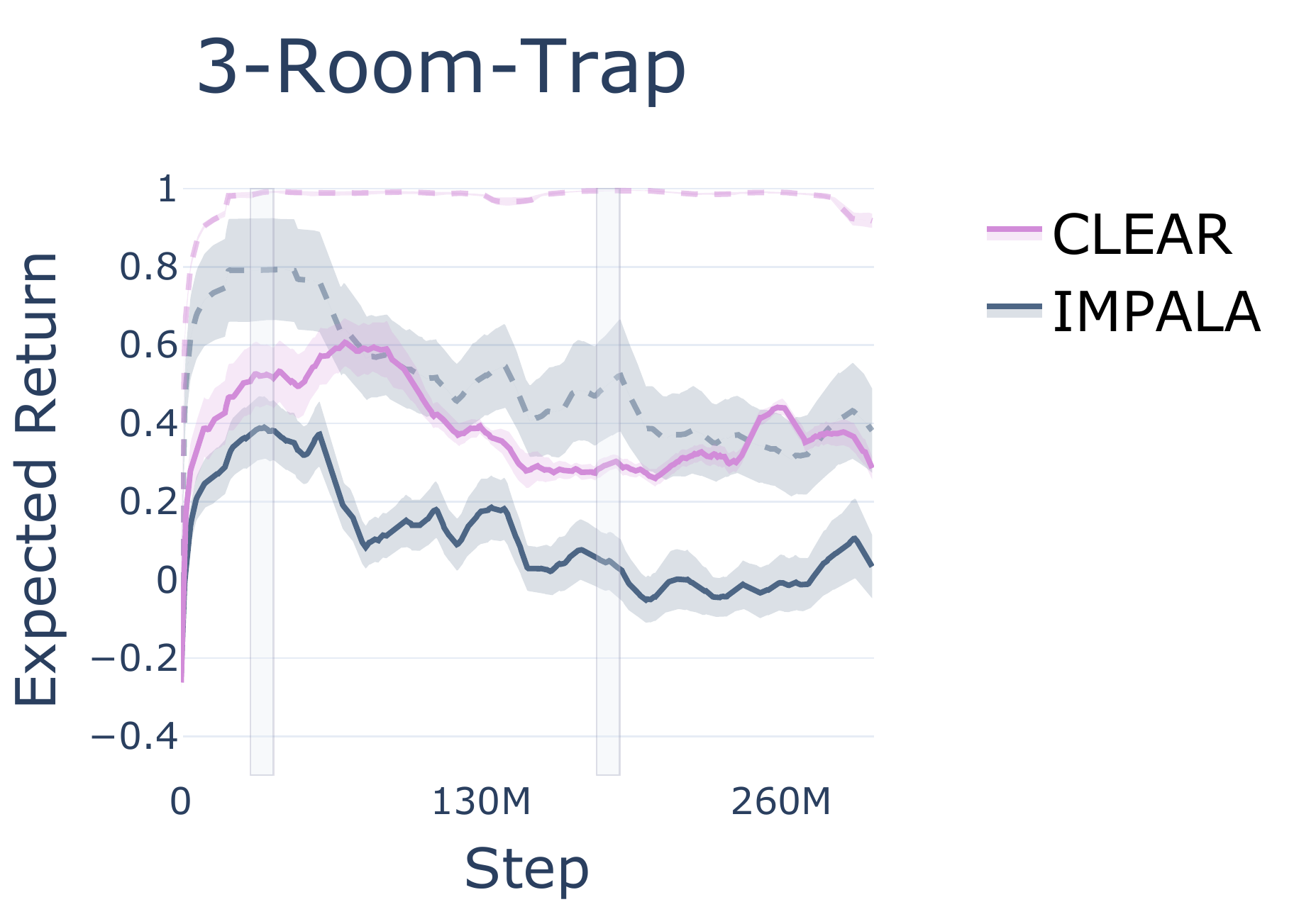} 
    \includegraphics[trim=0 3.5em 0em 0, clip, width=0.225\textwidth]{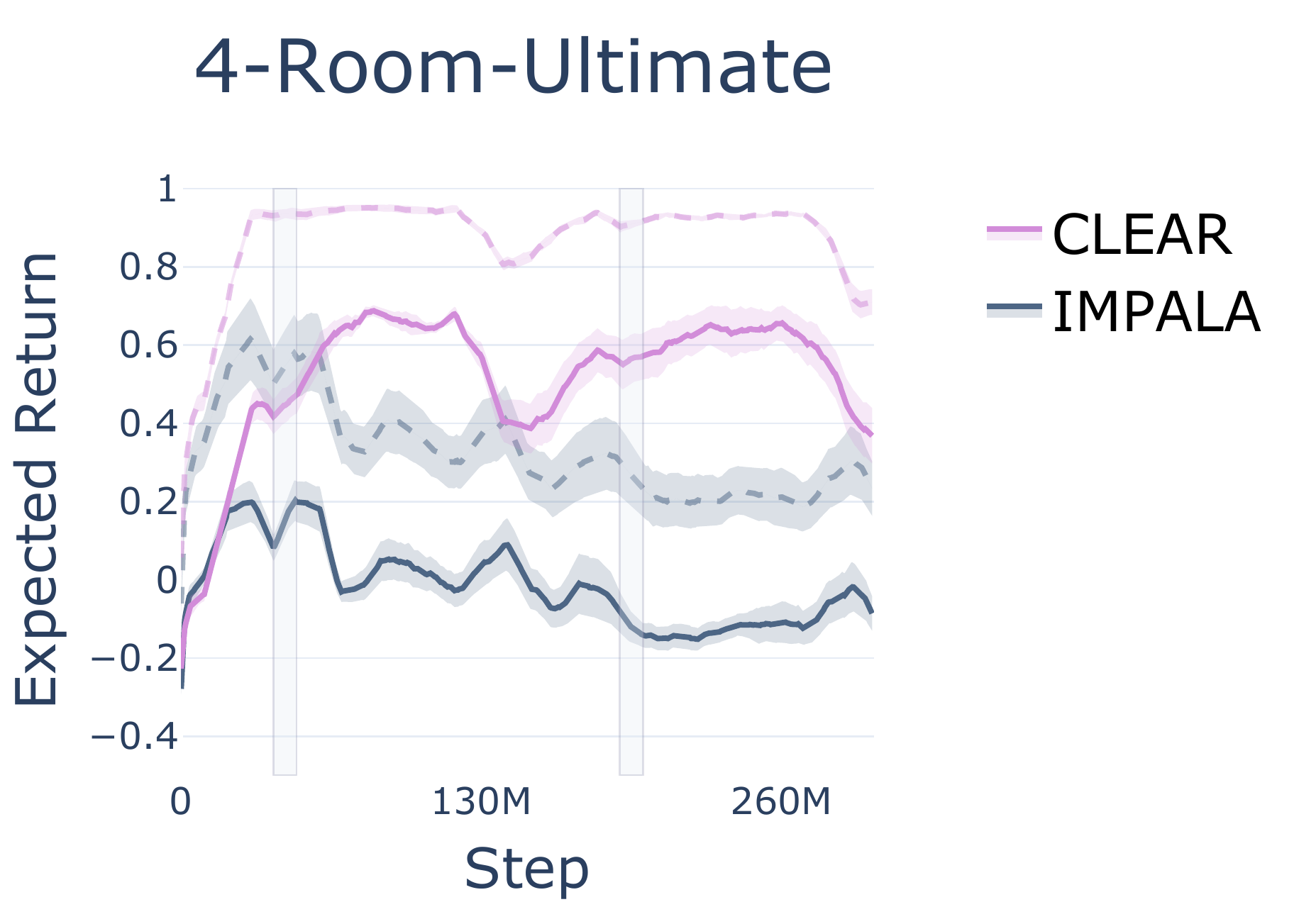} \\
    \includegraphics[trim=0 3.5em 13em 0, clip, width=0.17\textwidth]{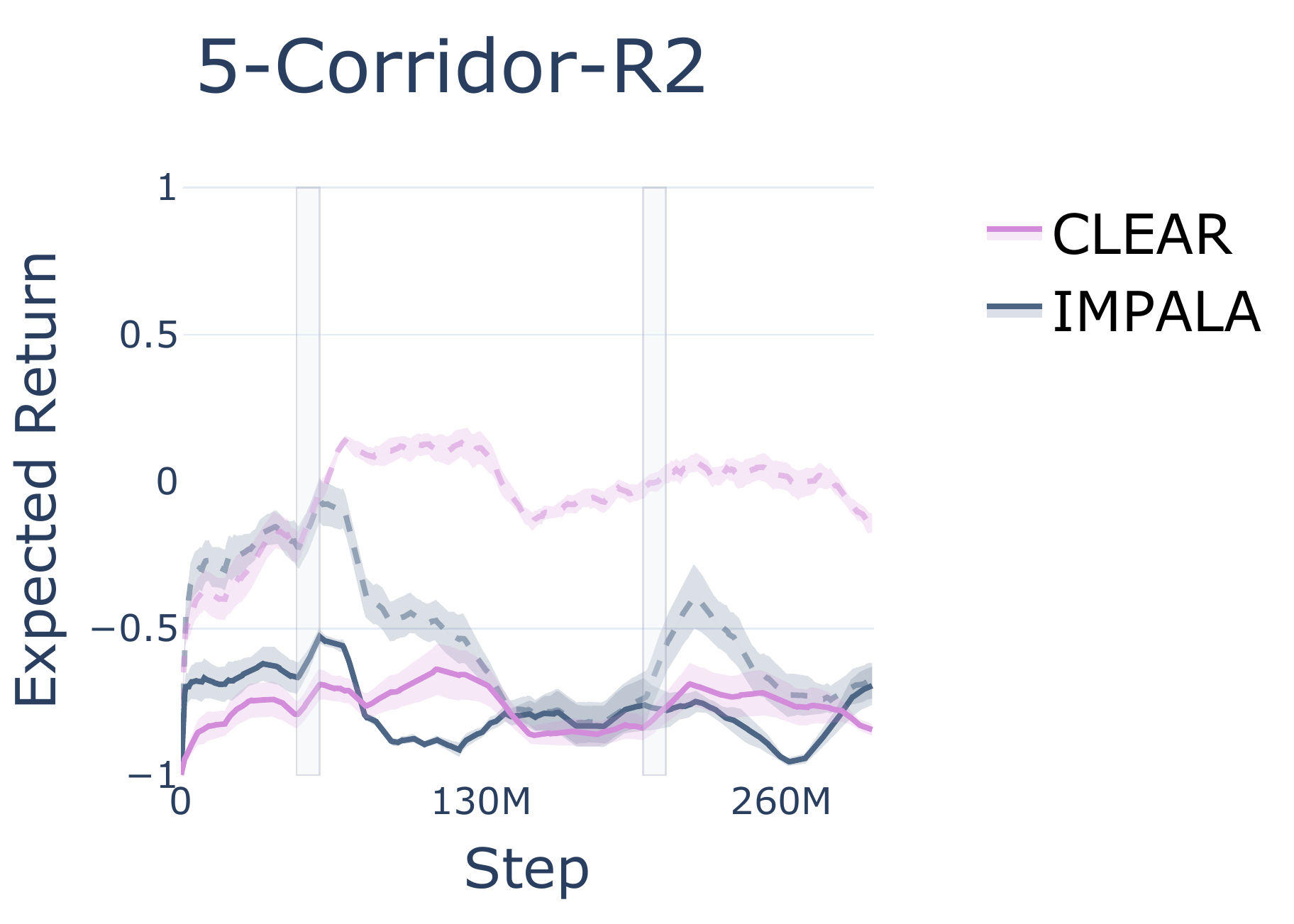}
    \includegraphics[trim=0 3.5em 13em 0, clip, width=0.17\textwidth]{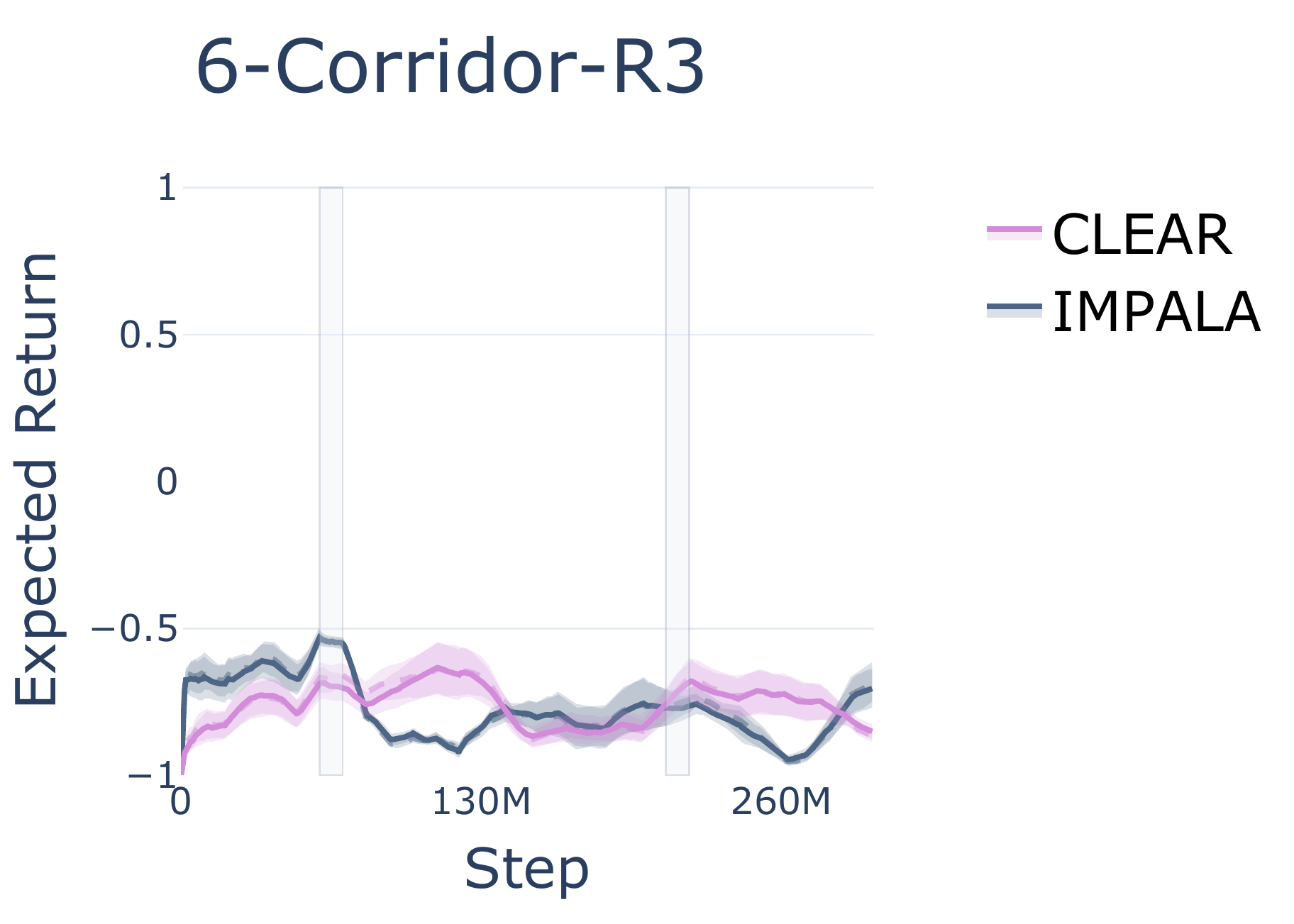} 
    \includegraphics[trim=0 3.5em 13em 0, clip, width=0.17\textwidth]{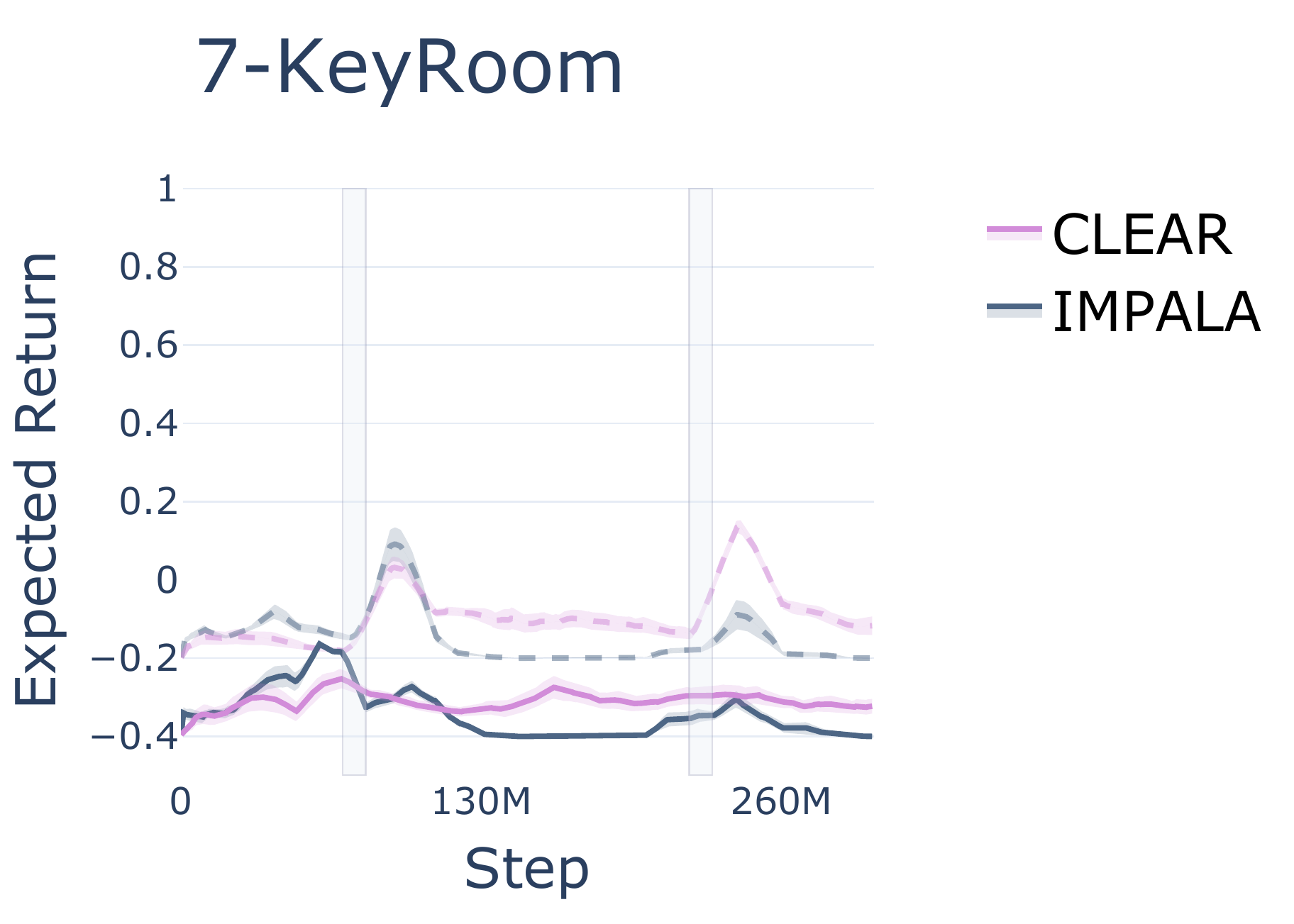} 
    \includegraphics[trim=0 3.5em 13em 0, clip, width=0.17\textwidth]{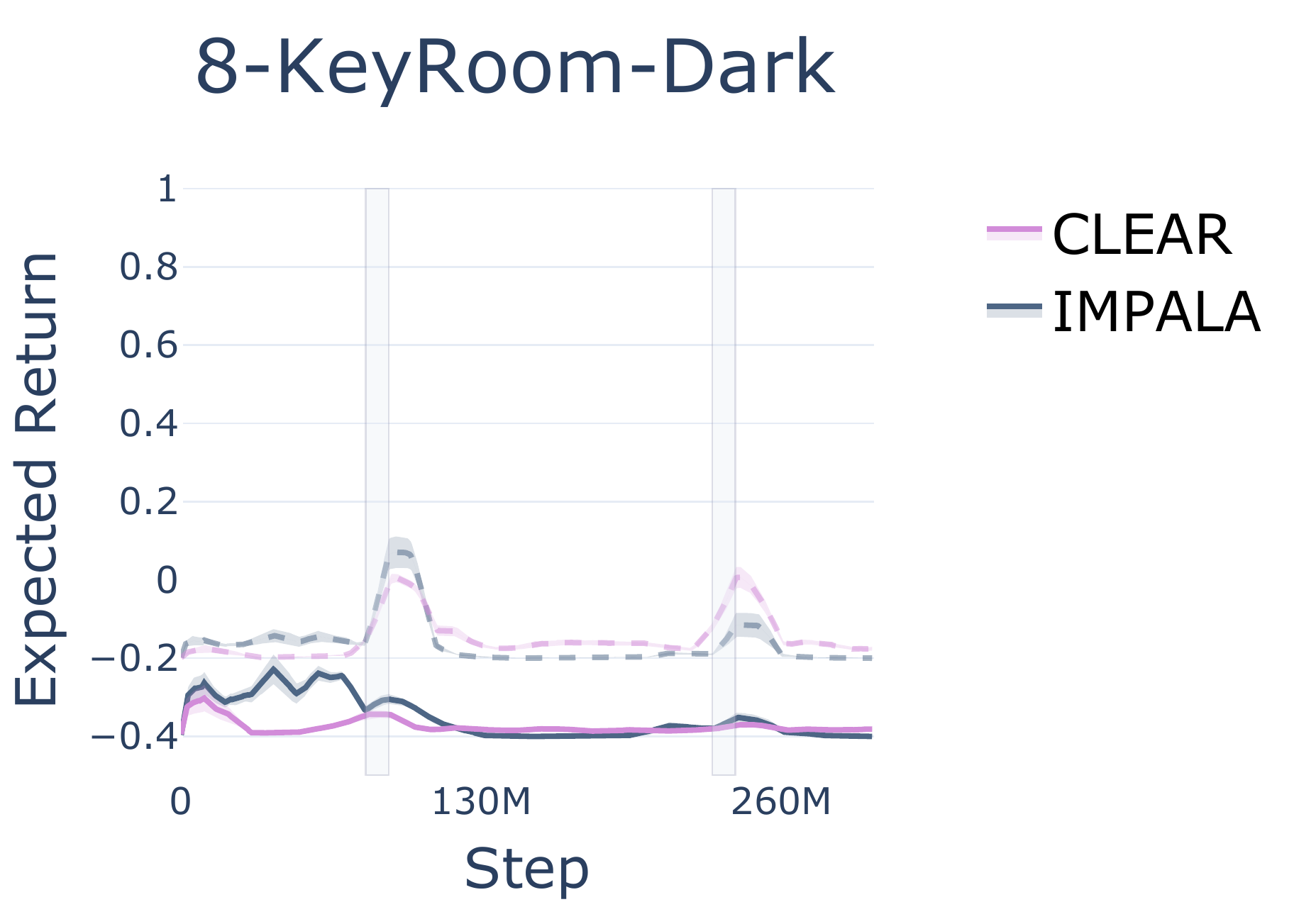} 
    \includegraphics[trim=0 3.5em 13em 0, clip, width=0.17\textwidth]{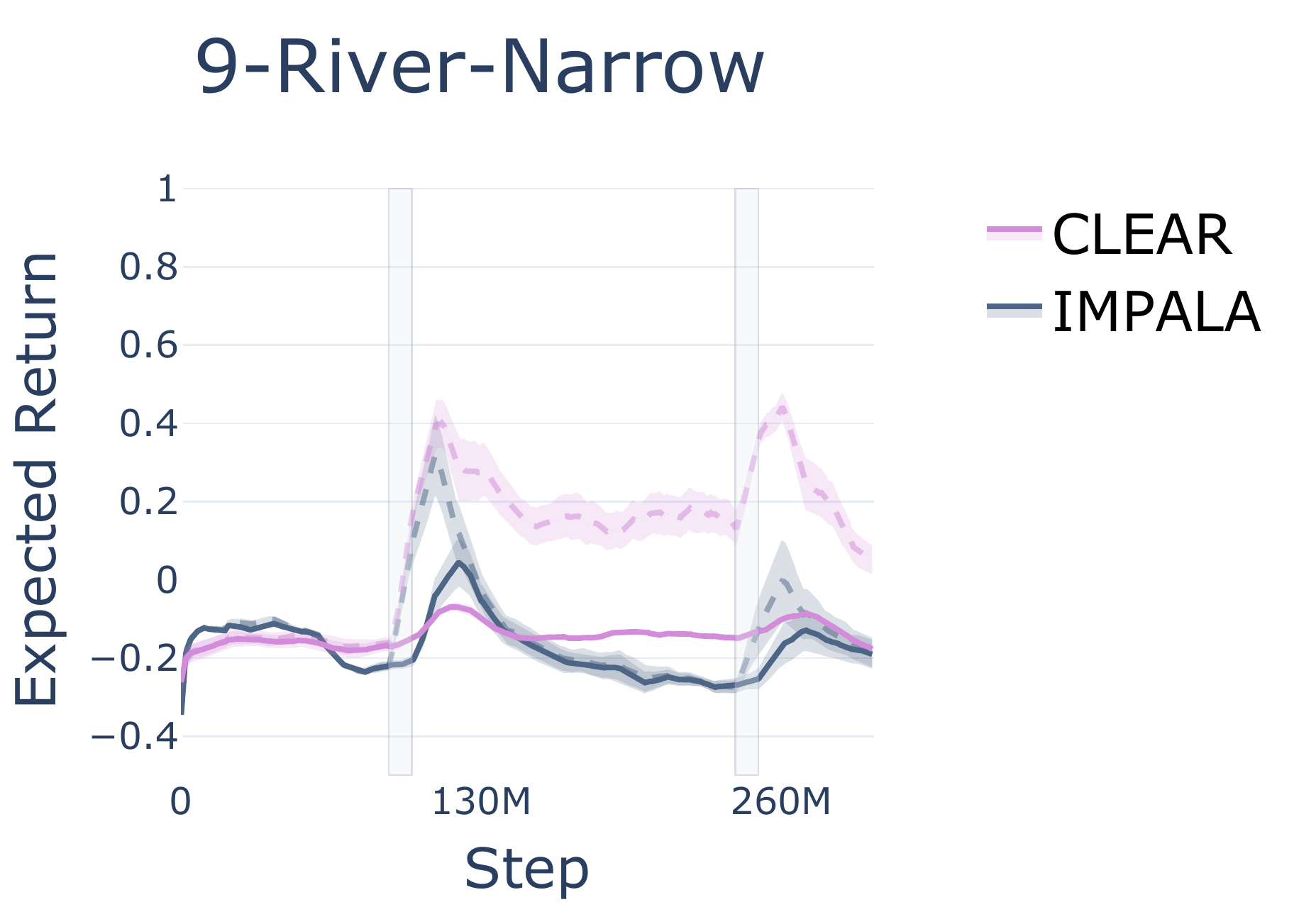} \\
    \includegraphics[trim=0 0em 13em 0, clip, width=0.17\textwidth]{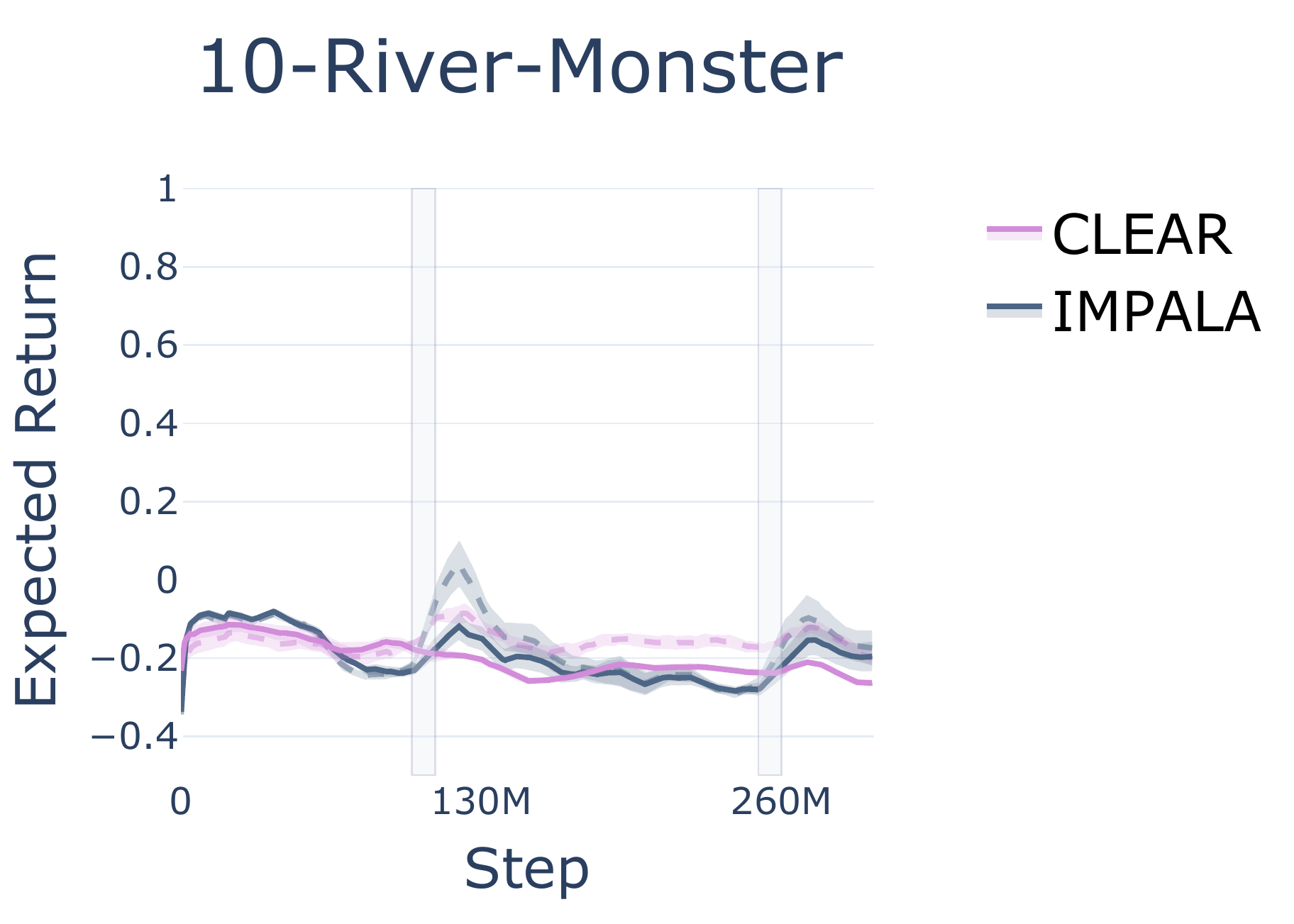} 
    \includegraphics[trim=0 0em 13em 0, clip, width=0.17\textwidth]{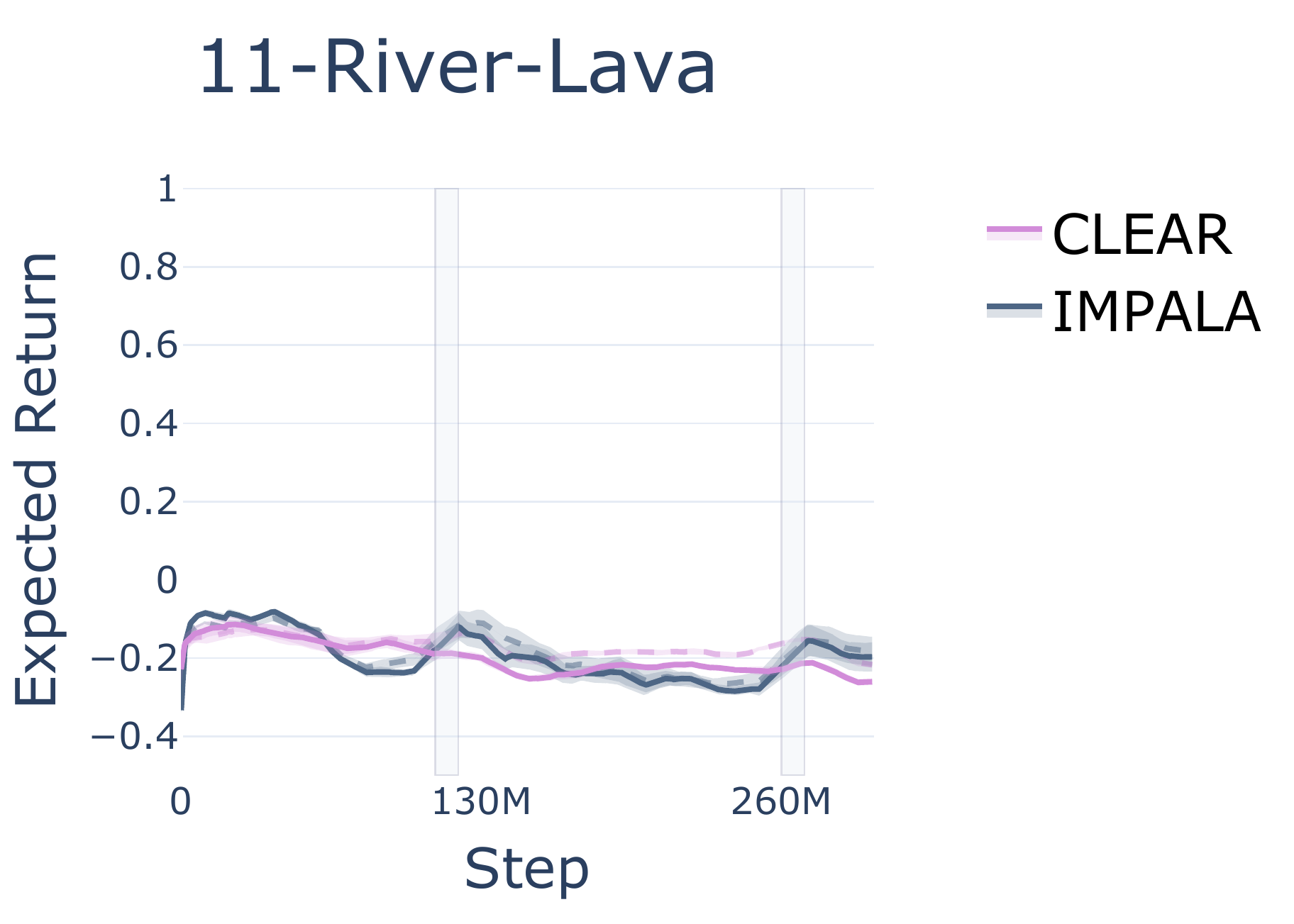}
    \includegraphics[trim=0 0em 13em 0, clip, width=0.17\textwidth]{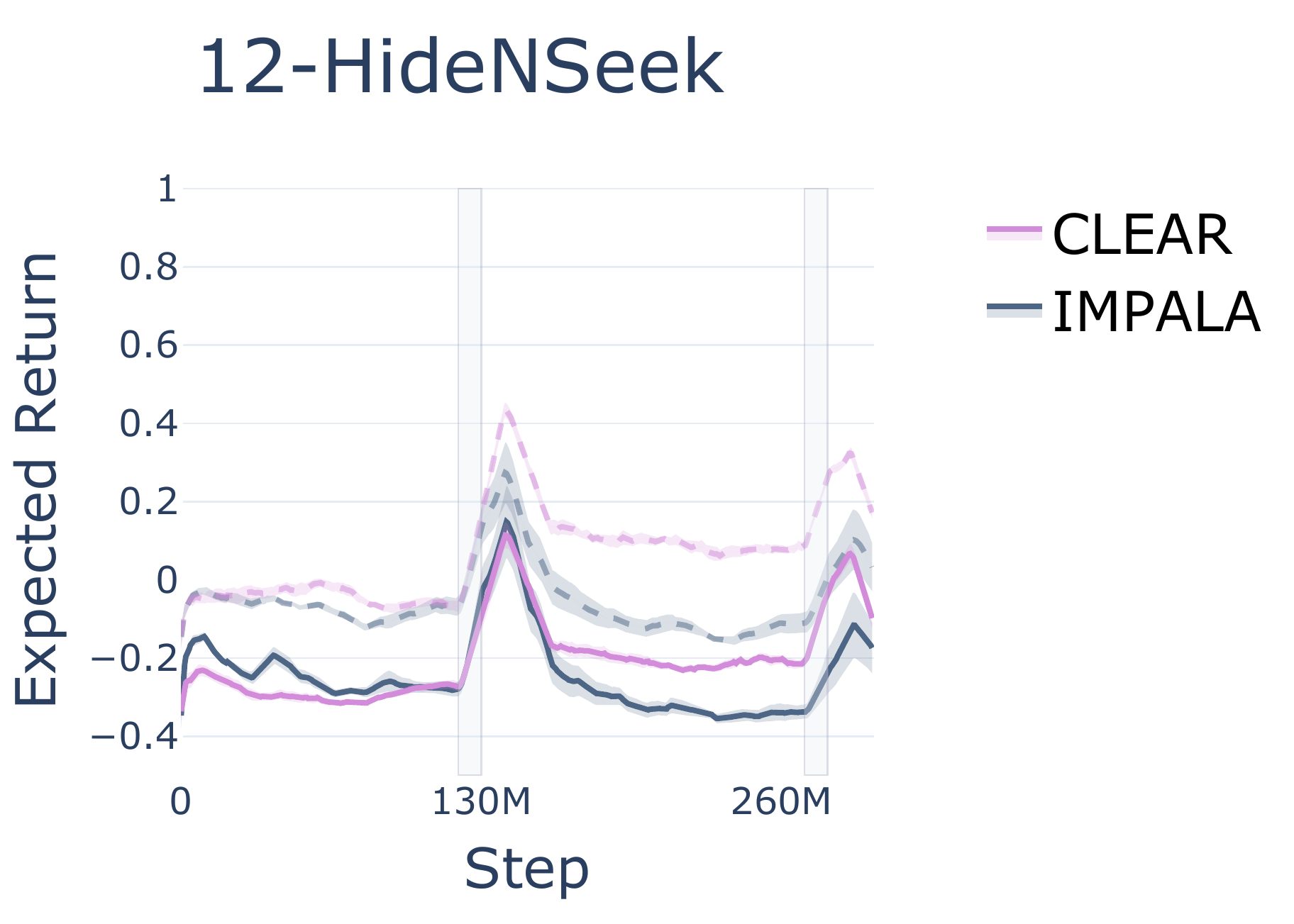}
    \includegraphics[trim=0 0em 13em 0, clip, width=0.17\textwidth]{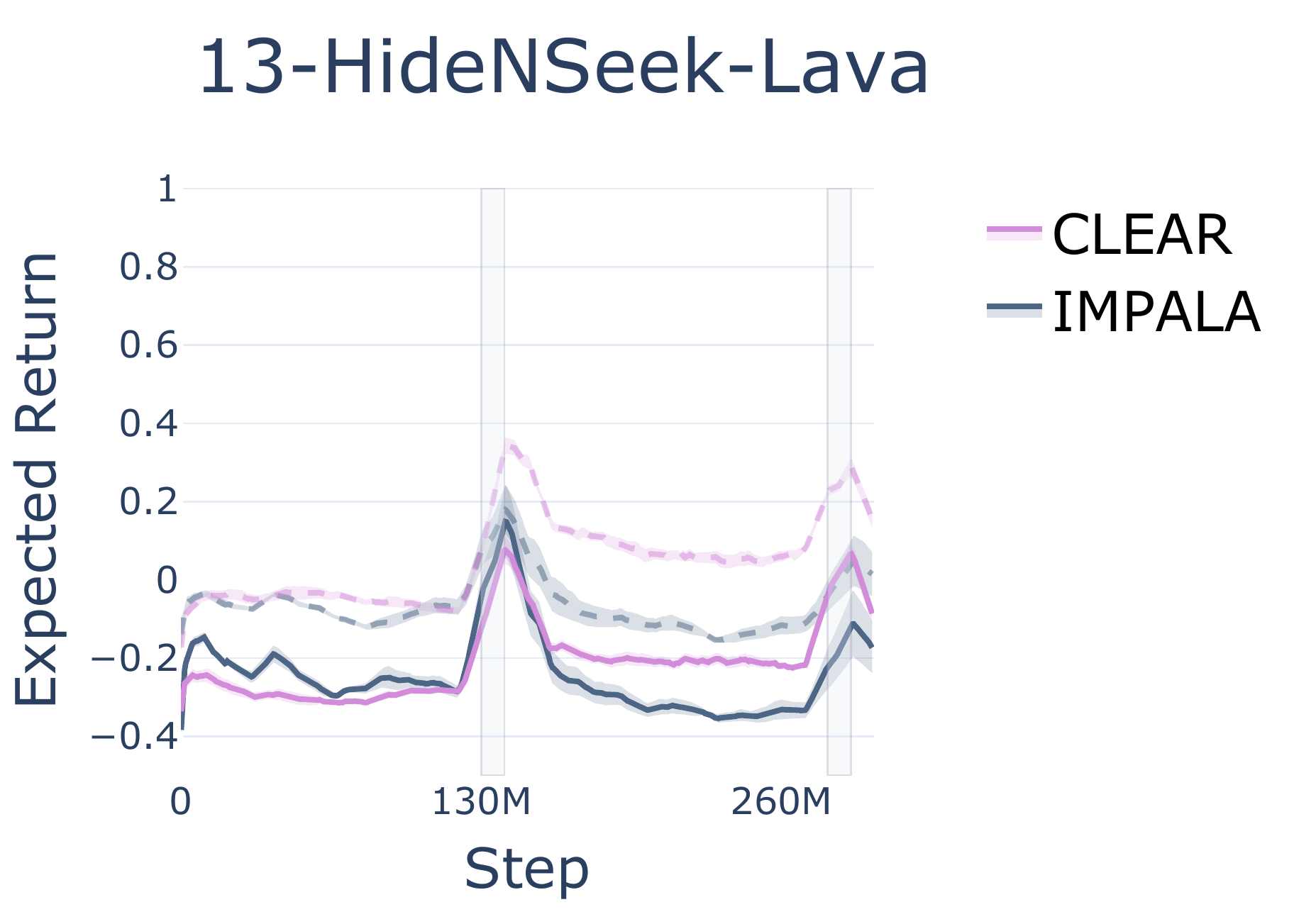}
    \includegraphics[trim=0 0em 13em 0, clip, width=0.17\textwidth]{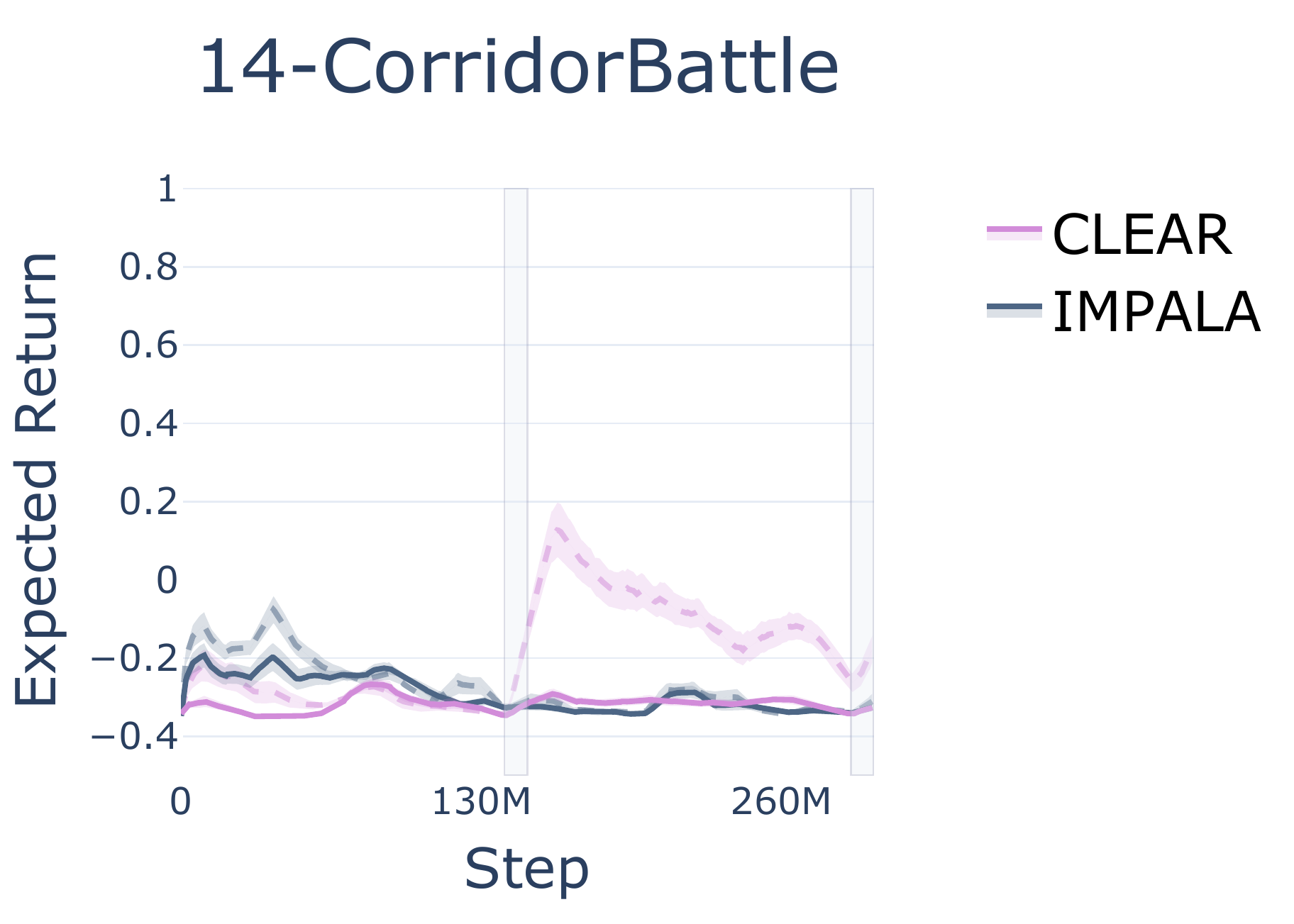}
    \caption{Results for Continual Evaluation $(\mathcal{C})$ on the 15 MiniHack task pairs sequence. The solid line shows evaluation on unseen testing environments; the dashed line shows evaluation on training environments. Gray shaded rectangles show when the agent trains on each task.}
    \label{fig:minihack_results}
\end{figure}

\textbf{Benchmark analysis:} From the summary statistics in Table~\ref{tab:summary_metrics}, we observe that IMPALA and CLEAR show minor transfer on MiniHack tasks compared to Atari and Procgen. Looking at the Transfer metric diagnostic table in Appendix~\ref{sec:minihack_transfer}, we can see that the first task 0-Room-Random transfers significantly to all other tasks, which can be interpreted as the agent learning the basics of moving around a MiniHack environment. Furthermore, we observe transfer from environments to others of the same type. For example, the Room environments generally positively transfer to each other, while mostly negatively transferring to the later River and HideNSeek environments. When training tasks share the same testing task, such as 12-HideNSeek and 13-HideNSeek-Lava, the transfer metric is noticeably high, as expected. Finally, from the Continual Evaluation results in Figure~\ref{fig:minihack_results}, we observe that MiniHack effectively tests for plasticity as well, as later experiments fail to learn effectively. 



\textbf{Algorithm design:} From the Continual Evaluation results in Figure~\ref{fig:minihack_results}, we can see CLEAR generally performs well at learning tasks and mitigating catastrophic forgetting for the first five tasks. However, we observe that the agent struggles to learn later tasks (fails to maintain plasticity), and that there is a significant out-of-distribution generalization gap, in performance on test (solid) compared to train (dashed) environments for all tasks. Additionally, inspecting the forgetting metric diagnostic tables in Appendix~\ref{sec:minihack_forgetting}, we see that the HideNSeek tasks exhibit particularly high forgetting. These results and shortcomings present important areas for new algorithms to pursue.

\subsection{CHORES results}
\label{section:chores_results}

\begin{figure}[h]
    \centering
    \subfloat[Mem-VaryRoom sequence]{
    \includegraphics[trim=0 4em 11.5em 0, clip, width=0.15\textwidth]{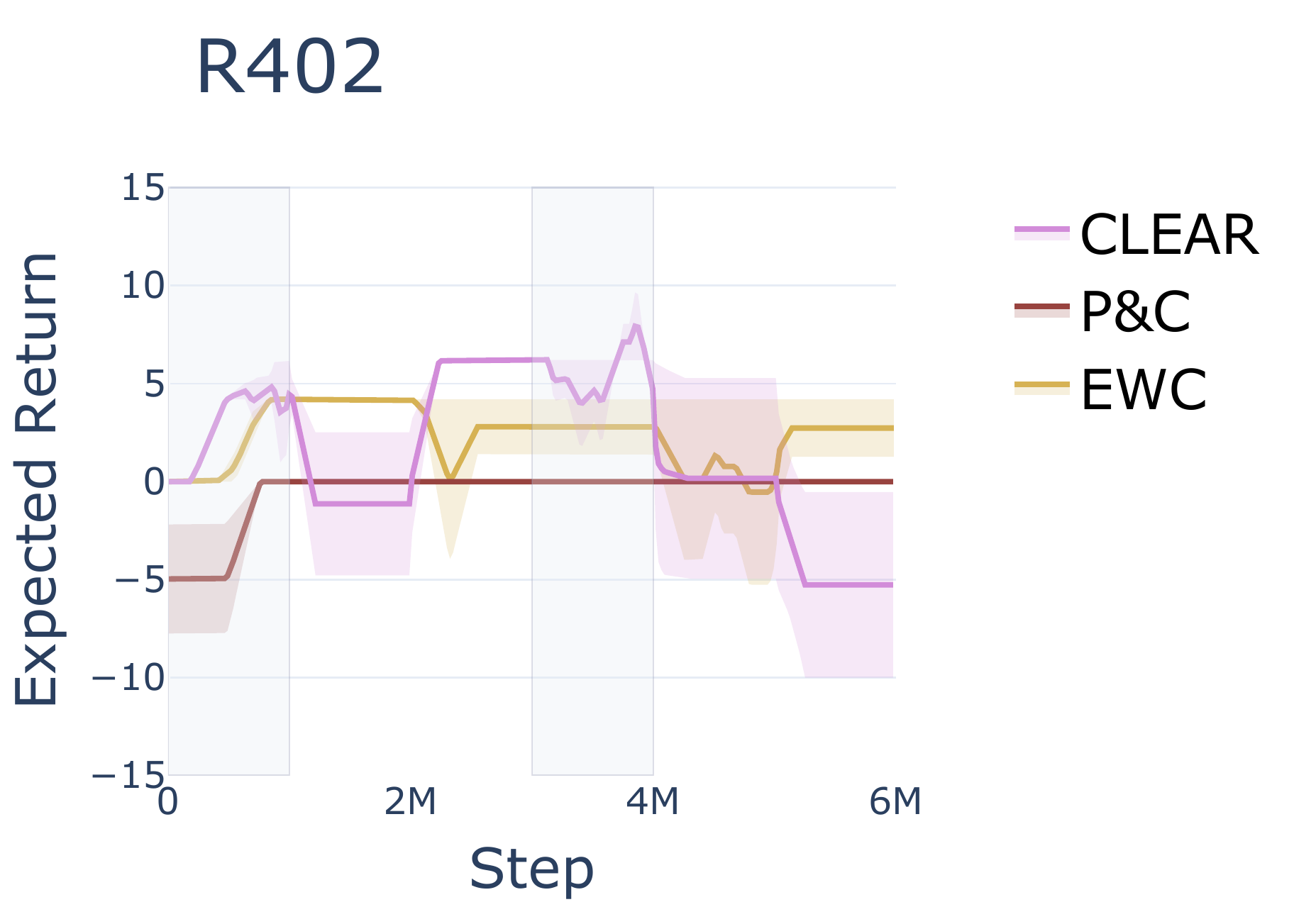}
    \includegraphics[trim=0 4em 11.5em 0, clip, width=0.15\textwidth]{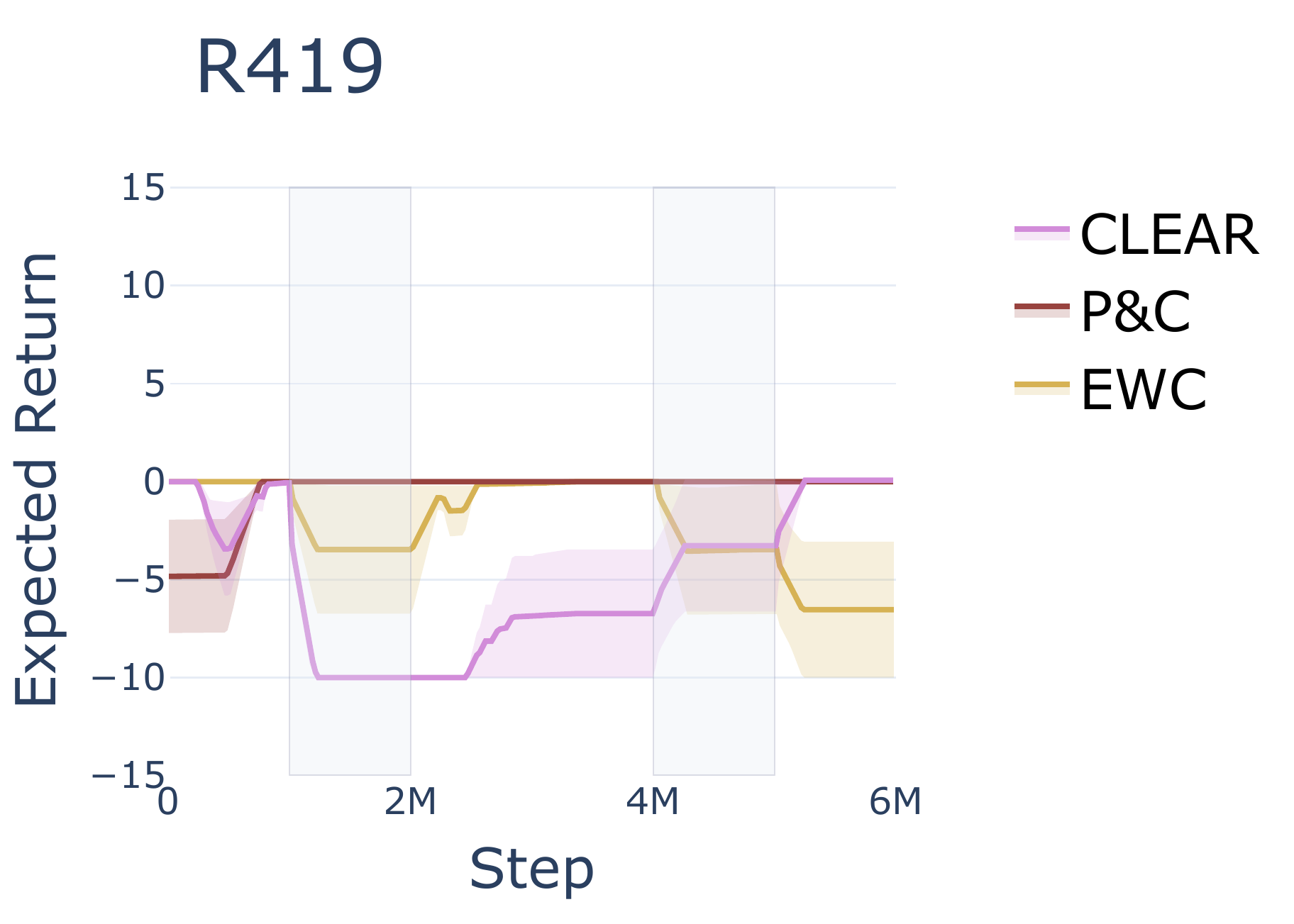}
    \includegraphics[trim=0 4em 11.5em 0, clip, width=0.15\textwidth]{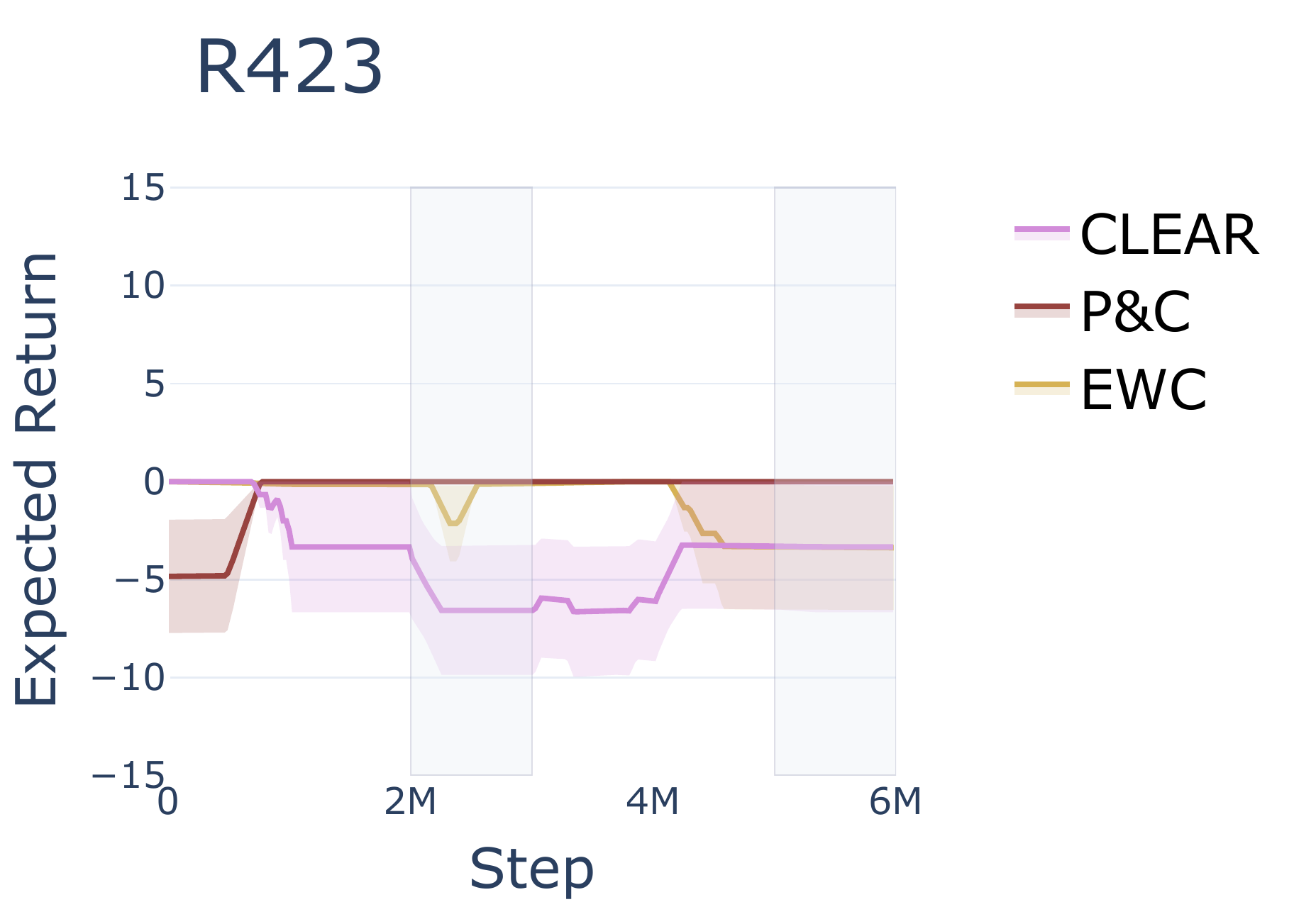}
    }
    \subfloat[Mem-VaryTask sequence]{
    \includegraphics[trim=0 4em 11.5em 0, clip, width=0.15\textwidth]{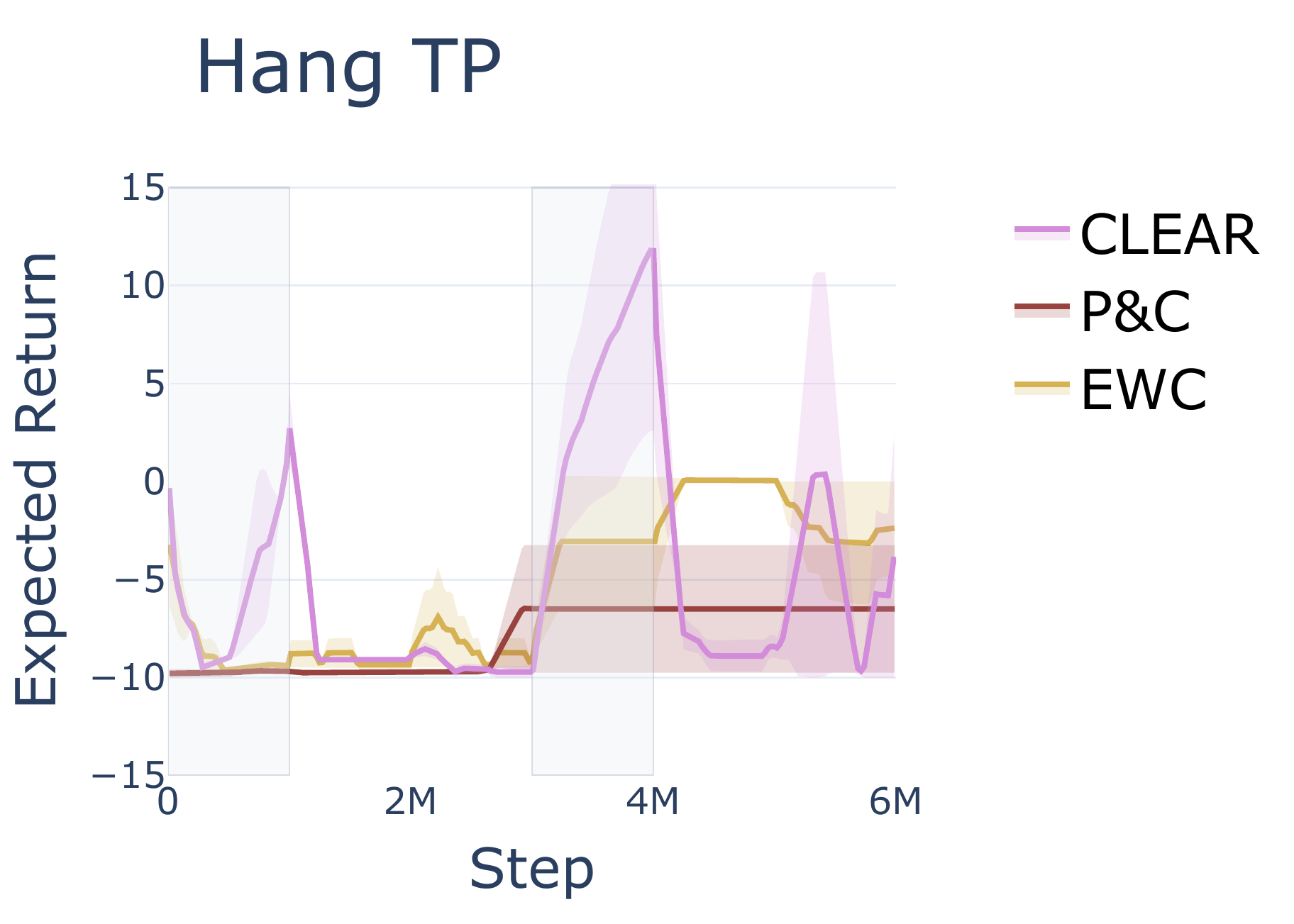}
    \includegraphics[trim=0 4em 11.5em 0, clip, width=0.15\textwidth]{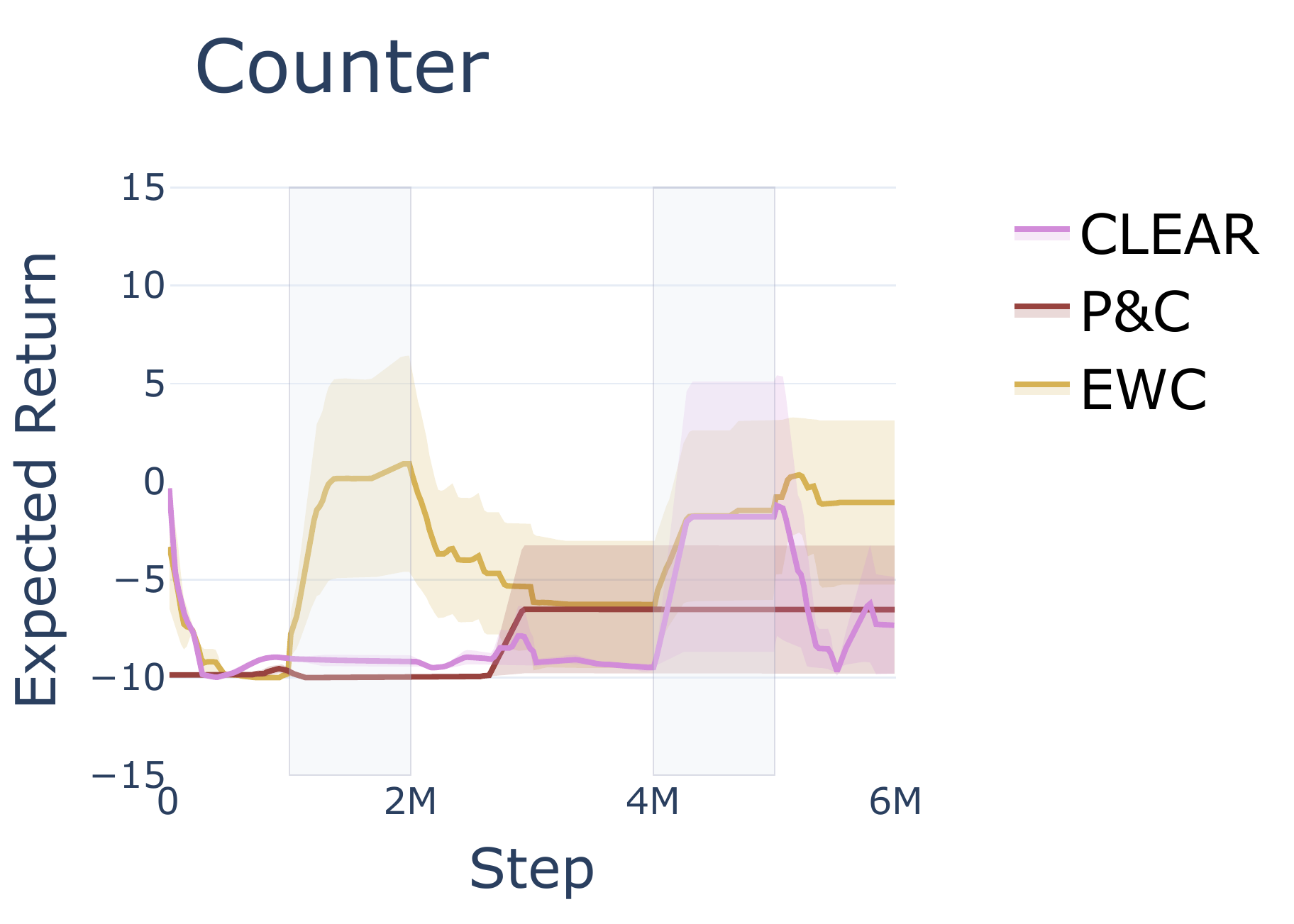}
    \includegraphics[trim=0 4em 0 0, clip, width=0.19\textwidth]{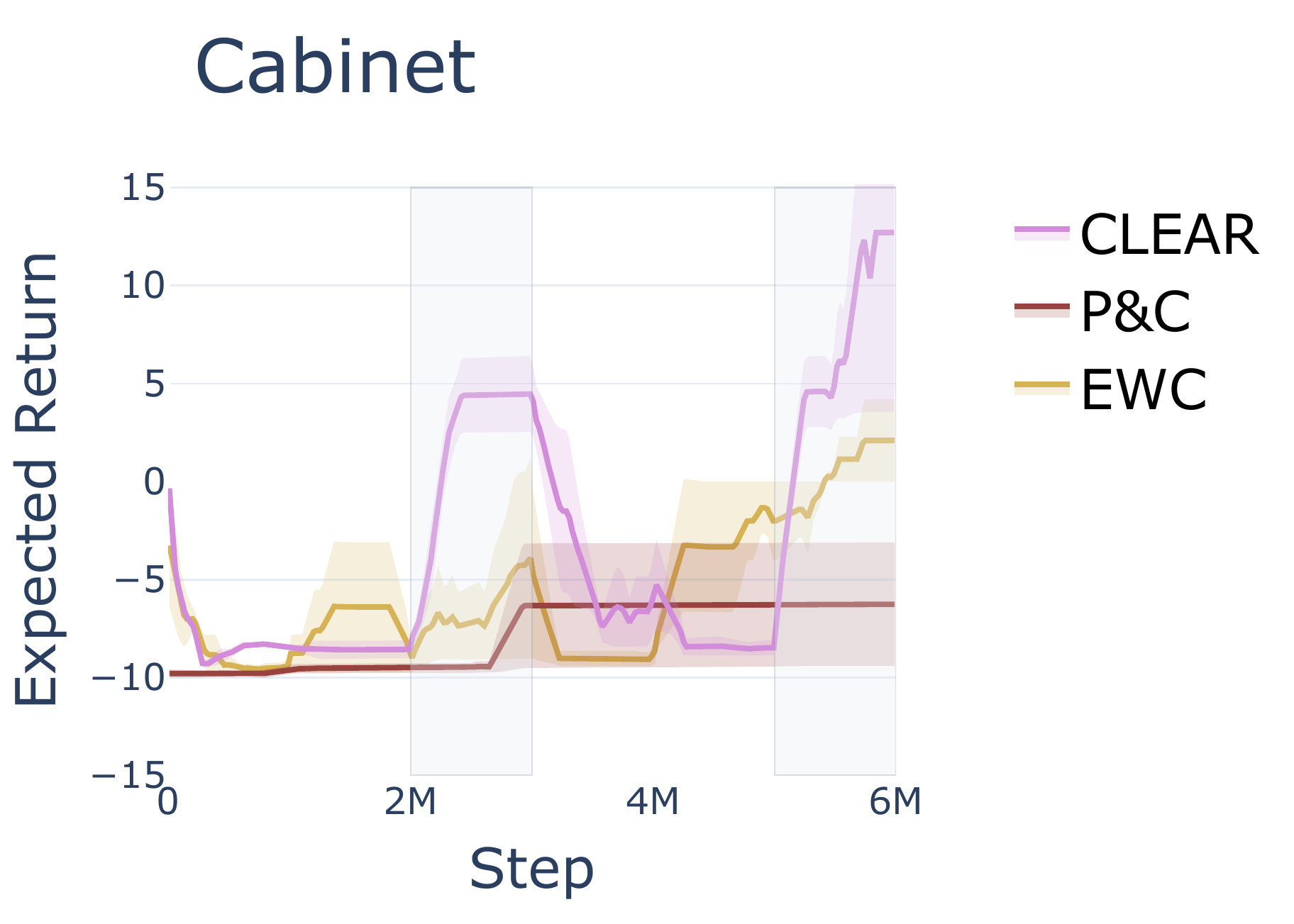}
    }
    
    \subfloat[Mem-VaryObject sequence]{
    \includegraphics[trim=0 1em 11.5em 0, clip, width=0.15\textwidth]{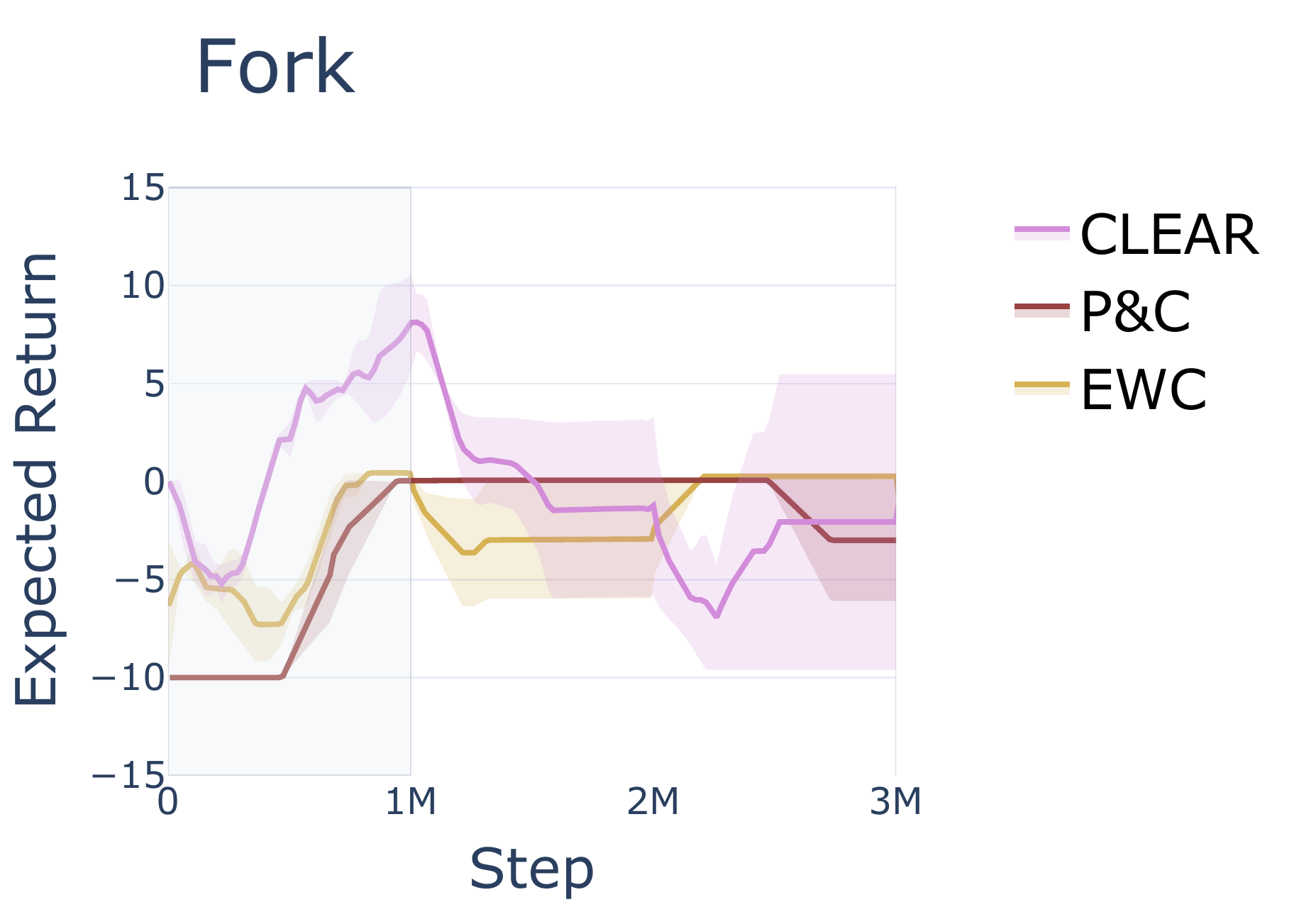}
    \includegraphics[trim=0 1em 11.5em 0, clip, width=0.15\textwidth]{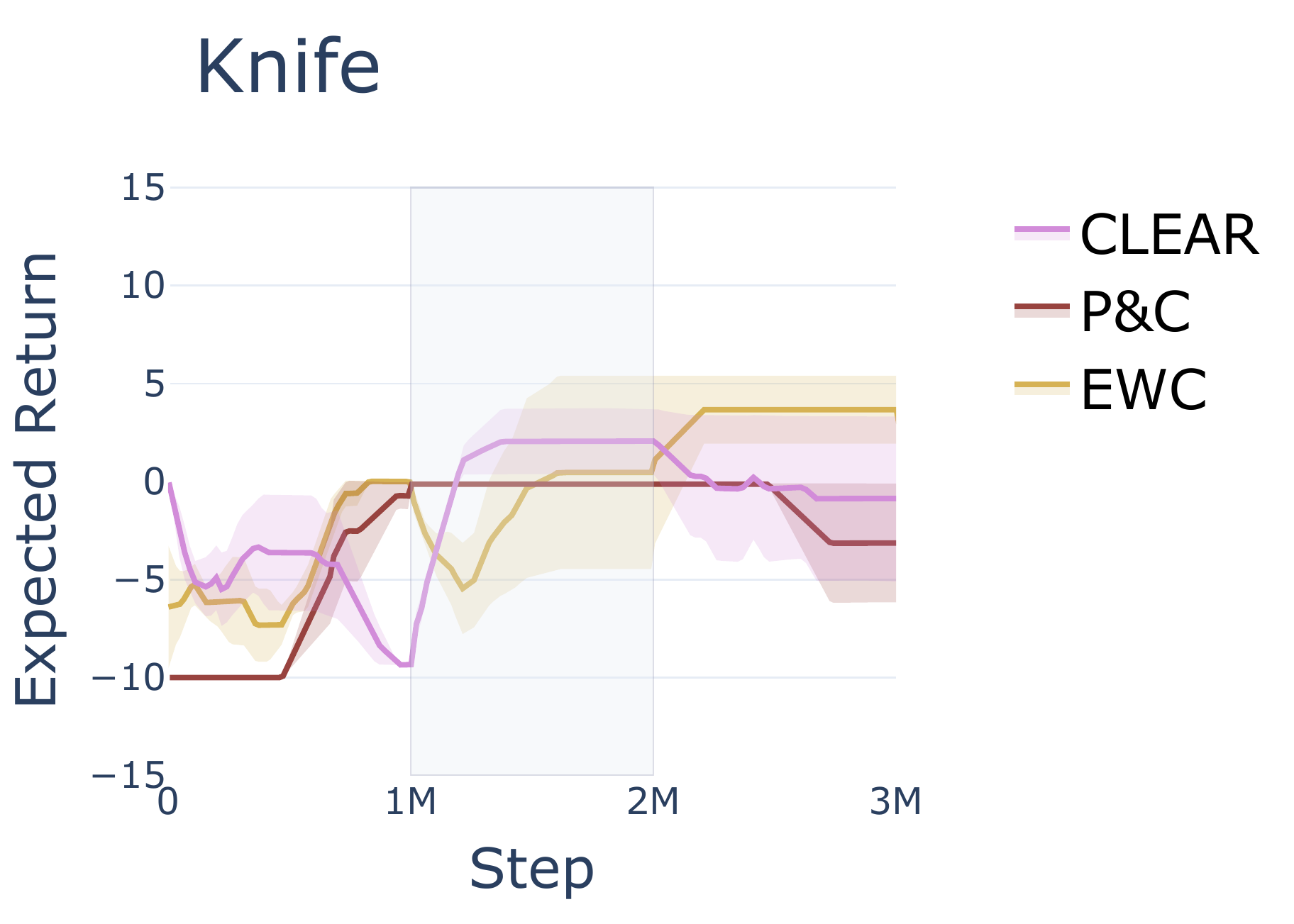}
    \includegraphics[trim=0 1em 11.5em 0, clip, width=0.15\textwidth]{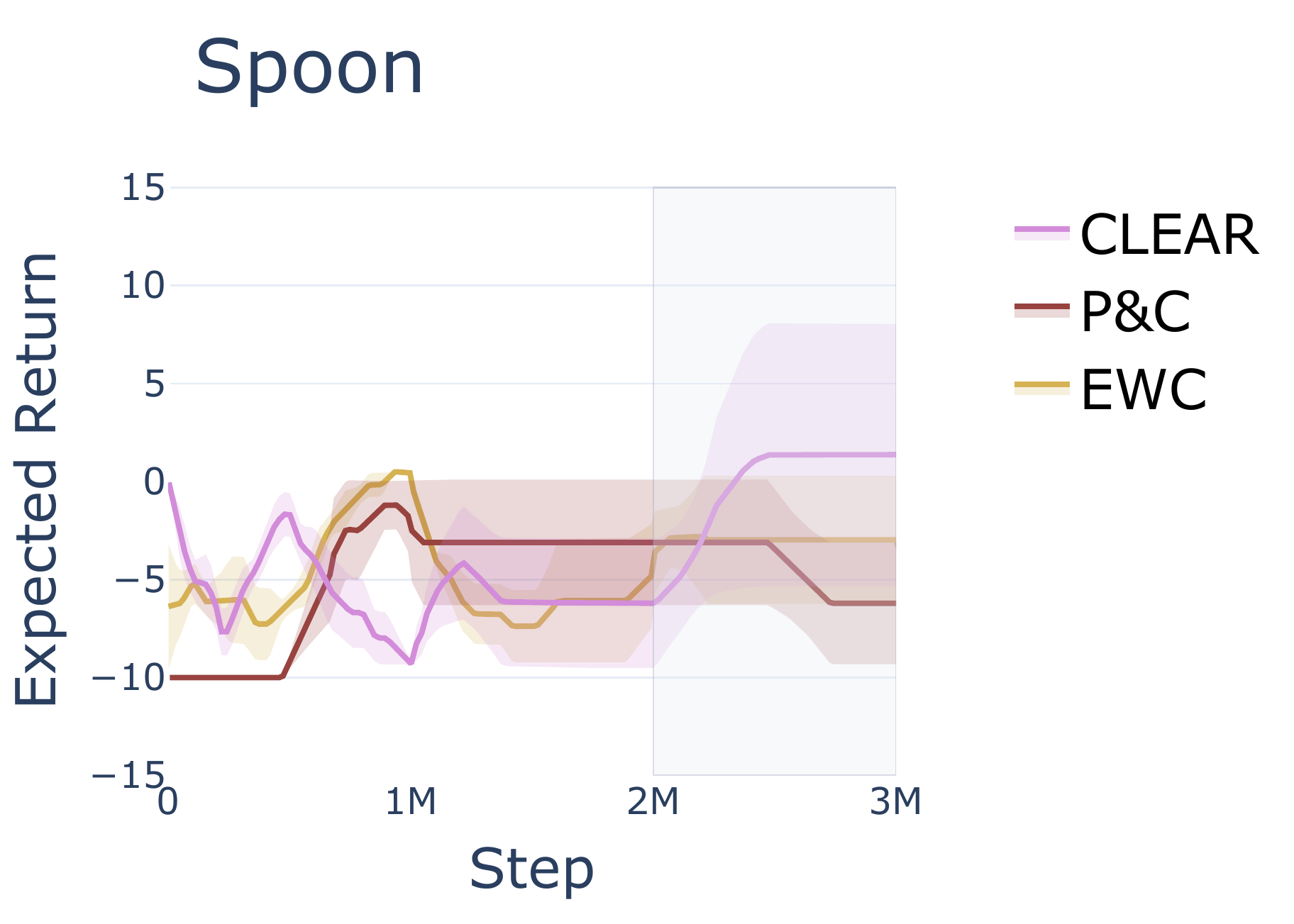}
    }
    \subfloat[Gen-MultiTraj sequence]{
    \includegraphics[trim=0 1em 11.5em 0, clip, width=0.15\textwidth]{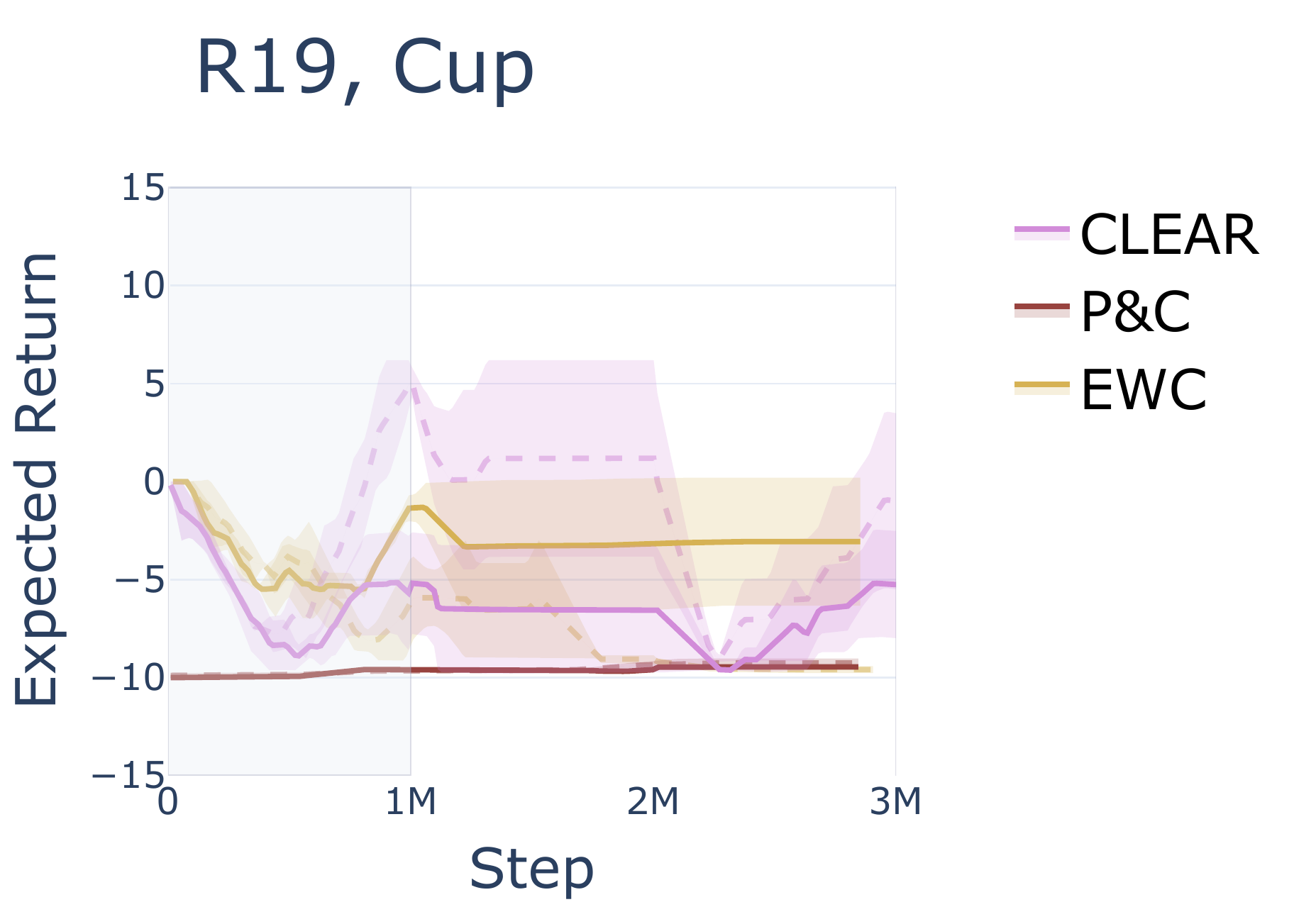}
    \includegraphics[trim=0 1em 11.5em 0, clip, width=0.15\textwidth]{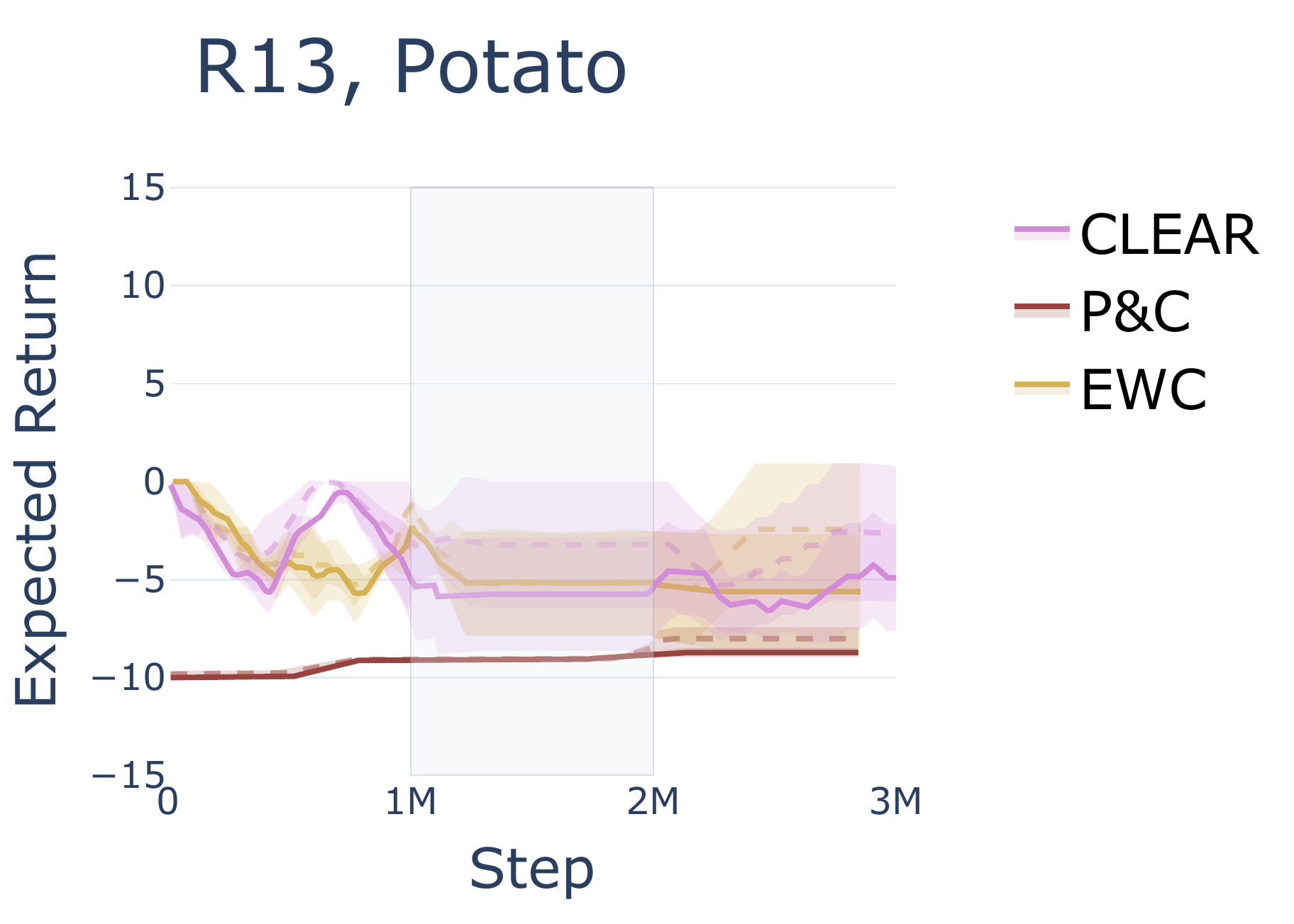}
    \includegraphics[trim=0 1em 0 0, clip, width=0.19\textwidth]{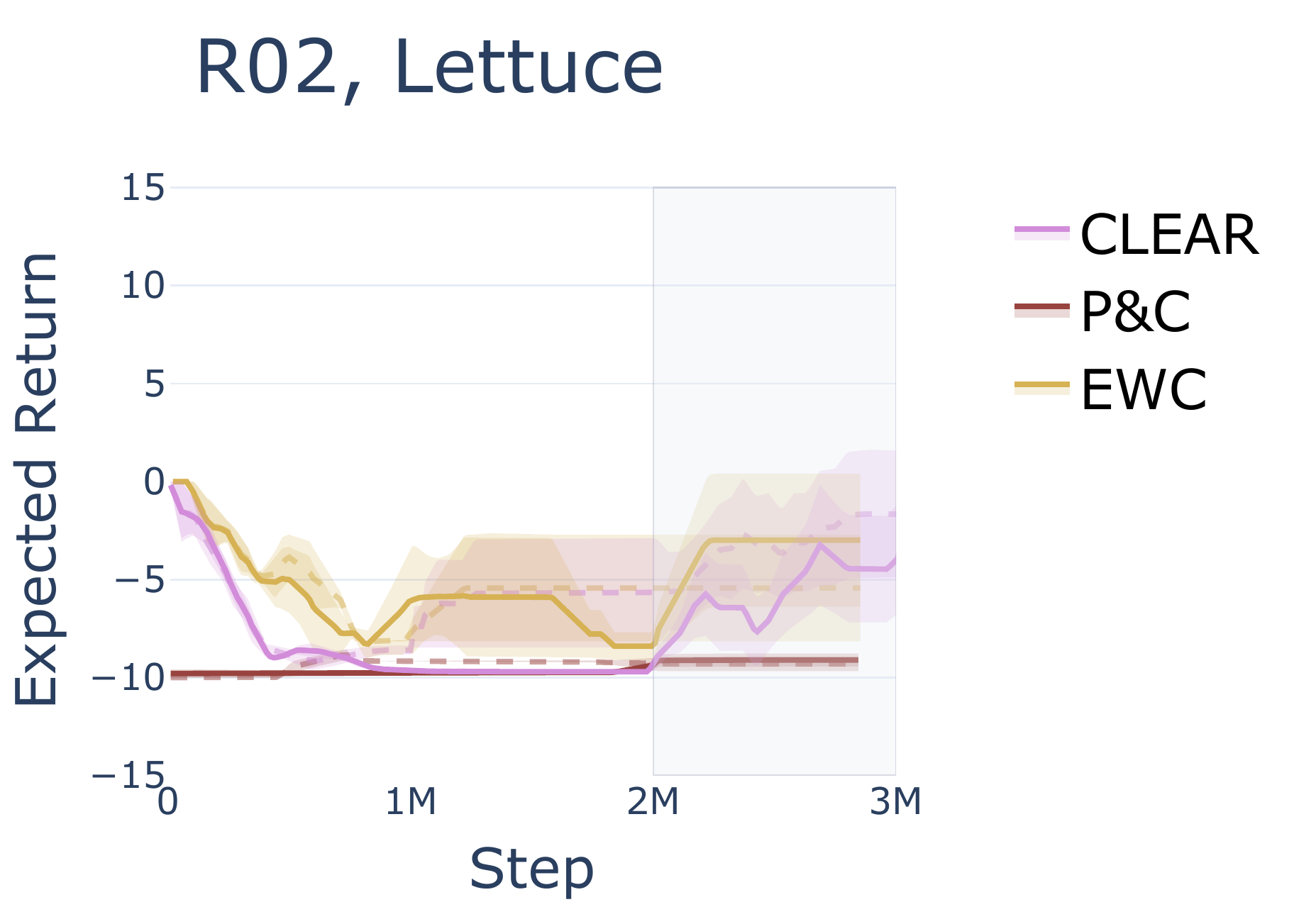}
    }
    \hspace{4em}
    \caption{Results for Continual Evaluation $(\mathcal{C})$ on the CHORES suite of benchmarks. For (d) Gen-MultiTraj, the solid line shows evaluation on unseen testing environments; the dashed line shows evaluation on training environments. Gray shaded rectangles show when the agent trains on each task.}
    \label{fig:chore_results}
\end{figure}

We show Continual Evaluation results for the four CHORES in Figure~\ref{fig:chore_results}. Forgetting and Transfer metrics diagnostic tables are available in Appendix~\ref{sec:chores_forgetting} and Appendix~\ref{sec:chores_transfer}. On the memorization sequences, the testing environment is the same as the training environment, and evaluation is represented with a solid line. In our generalization experiment, the solid line represents evaluation on held-out testing environments, and the dotted line represents performance on the training environments. We report 2 cycles for each of (a) Mem-VaryRoom and (b) Mem-VaryTask, and 1 for each of (c) Mem-VaryObject and (d) Gen-MultiTraj, due to time constraints.

\textbf{Benchmark analysis:} From the Continual Evaluation results, we find that CHORES are challenging, and current agents achieve low returns overall. Some learning occurs for the first task of every sequence, but nearly none in later tasks, with the exception of (a) Mem-VaryTask. We also observe no generalization to unseen contexts from held-out demo trajectories in (d) Gen-MultiTraj. The Transfer and Forgetting metrics are less meaningful with low returns, but there is some indication of forward transfer, particularly on (c) Mem-VaryObj. Taken together, we can see CHORES as the reach-goal; a set of tasks current methods cannot solve, and that will truly test the sample efficiency of future methods.

\textbf{Algorithm design:} CLEAR achieves the highest returns overall, but there is significant room to improve learning these tasks. We observe significant Forgetting, particularly on (c) Mem-VaryObj, illustrating one such area for improvement. Advances in sample efficiency and exploration are likely required for agents to make progress on this challenge.

\section{Conclusion}

In this paper, we present CORA, a platform designed to reduce the barriers to entry for continual reinforcement learning. CORA provides a set of benchmarks, open-sourced implementations of several baselines, evaluation metrics, and the modular \texttt{continual\_rl} package to contain it all. Each benchmark is designed to exercise different aspects of continual RL agents: a standard, proven Atari benchmark for catastrophic forgetting and sample efficiency; a Procgen benchmark to test forgetting and generalization to unseen environment contexts; a MiniHack benchmark to test generalization, plasticity, and transfer; and the new, challenging, CHORES benchmark to test capability in a visually-realistic environment where sample-efficiency is key. With these benchmarks, we demonstrate the strengths and weaknesses of the current state-of-the-art continual RL method, CLEAR. While CLEAR generally outperforms the other baselines at learning tasks and mitigating catastrophic forgetting, significant improvements are needed for generalization, forward transfer, and maintaining plasticity over a long sequence of tasks. We are excited to introduce the community to CORA, and hope CORA can aid in the development, testing, and understanding of new methods in the field of continual RL.


\textbf{Limitations} We study task sequences that share a high-dimensional observation space (images) and have a discrete action space. These RL tasks are finite-horizon, with episodic resets. In this work, we also primarily study video-game environments, with procedurally-generated variation. These assumptions are shaped by our perspective of the continual RL problem and current state of the field. We acknowledge that different points of view exist, backed by design choices which may differ from the conditions we study. For instance, we consider task cycling in this work, which may favor replay-based methods such as CLEAR that can retain data from all tasks in the sequence, after the first cycle. This protocol does not apply to a pure online learning setup, where no assumptions may be made on the structure and similarities of incoming data. We intend to relax these assumptions as new methods are developed and CORA evolves.

\clearpage

\section*{Acknowledgements}
We thank Jonathan Schwarz and David Rolnick for discussion on implementing baselines, and helpful feedback on this project. We additionally thank Abhinav Shrivastava with help running experiments. This work was supported by ONR MURI, ONR Young Investigator Program, and DARPA MCS.

\bibliography{   
    continual_refs,
    rl_refs,
    refs
}
\bibliographystyle{collas2022_conference}

\clearpage

\appendix
\section{Extended Related Work}
\label{appendix:extended_related_work}

\textbf{Environments and tasks} Historically, RL agents were evaluated on simple control tasks with state-based inputs from the OpenAI Gym~\cite{brockman2016gym} or DeepMind Control Suite~\cite{tassa2018deepmind}. Some of these tasks have been shown to be easily solvable by random search algorithms~\cite{mania2018simple} and thus should not be considered as sufficiently difficult for comparing algorithms. Leveraging physics simulators, many environments have been proposed that involve robots, fixed in place, for object manipulation tasks of varying complexity~\cite{plappert2018multi, rajeswaran18dapg, gupta2020relay, xing2021kitchenshift, robogym2020, kannan2021robodesk, james2020rlbench, robosuite2020, yu2020meta, lee2021ikea}. Learning policies for robot manipulation is challenging, compounded by the exploration difficulty of the task, continuous action spaces, and the sample inefficiency of RL algorithms.

In this work, we choose environments with discrete action spaces, in order to use a single output policy across multiple tasks, rather than use separate output heads for different tasks. We make this decision considering the capabilities of current methods. For environments with continuous action spaces, task-agnostic single-headed architectures may be infeasible currently, as in~\cite{wolczyk2021continual}, so we leave this for future work. Furthermore, it is beneficial for continual RL for tasks to share a consistent observation space, allowing for the creation of common policies that can leverage task similarities. If environments rely on state-based inputs such as the positions of objects, it is usually the case that the state space changes for different tasks. To ensure a consistent observation space, we pick environments which offer image-based observation spaces.

RL is also frequently evaluated on video game-like environments, most commonly Atari~\cite{bellemare2013arcade}, among others~\cite{boisvert2018minigrid, boisvert2018gym_miniworld, vinyals2017starcraft, beattie2016deepmindlab, kempka2016vizdoom, juliani2019obstacle, guss2019minerl, cobbe2020procgen, kuettler2020nethack}. 
In this work, we reproduce prior continual RL results on Atari~\cite{schwarz2018progress, rolnick2018clear}. We also define new task sequences using Procgen~\cite{cobbe2020procgen} and MiniHack~\cite{samvelyan2021minihack}. MiniHack is designed to isolate more tractable subproblems in the highly challenging NetHack~\cite{kuettler2020nethack} environment. We believe Procgen and MiniHack, with fast, procedurally-generated, stochastic environments to be better testbeds for continual RL onwards.

Beyond video game environments, many home environment simulators have been proposed recently~\cite{brodeur2017home, savva2017minos, puig2018virtualhome, yan2018chalet, gao2019vrkitchen, savva2019habitat} which offer visually-realistic scenes for evaluating Embodied AI. Among these types of environments, we highlight AI2-THOR~\cite{kolve2017ai2}, Habitat 2.0~\cite{szot2021habitat}, iGibson~\cite{shen2020igibson}, Sapien~\cite{xiang2020sapien}, and ThreeDWorld~\cite{gan2021threedworld}, which feature a wide range of household objects and scene-level interaction tasks. In particular, AI2-THOR, Habitat 2.0, and iGibson provide multiple home scenes based on real-world data. These different scenes are useful for applying realistic domain shift between tasks and evaluating forward transfer when learning later tasks. We choose AI2-THOR as a simulation environment in this benchmark because it offers a higher-level discrete action space, compared to Habitat 2.0 or iGibson at the time of development, along with a diverse set of demonstrations released in ALFRED~\cite{shridhar2020alfred}.
We also note that recent work using AI2-THOR has done evaluations with object manipulation tasks~\cite{shridhar2020alfred, batra2020rearrangement, ehsani2021manipulathor}, paving the way for more complex action spaces.

\section{Background}
\label{appendix:background}

Formally, we consider each task $\mathcal{T}$ as a finite, discrete-time Markov decision process (MDP), represented by a tuple $\brackets{S, A, T, r, \rho_0}$, with state space $S$, action space $A$, state transition probability function $T$, reward function $r$, and probability distribution $\rho_0$ on the initial states $S_0 \subset S$. In the standard reinforcement learning setup, the goal is to learn a policy $\pi(a | s)$ which maximizes the expected return $\mathcal{R}$, where the return of a state $s$ is defined as the sum of (discounted) rewards over a finite-length episode from state $s$.
\\ \\
We refer the reader to~\cite{zhang2018study, ghosh2021generalization, kirk2021survey}, on which we base our discussion of generalization in RL using a Contextual MDP (CDMP)~\cite{hallak2015contextual}. A state $s\in S$ can be decomposed as $(c, s')\in S_C$, where $s'\in S'$ is the underlying state and $c\in C$ is the context, such that $S=C \times S'$. We assume that the context is not observed by the agent, so the CDMP is a partially observable MDP (POMDP) with observation space $O$. 
\\ \\
The context $c$ remains fixed throughout an episode, and determines the environment variation. The initial state distribution may be factorized as $\rho_0 (s) := p(c) \rho_0 (s' | c)$, where $p(c)$ is called the context distribution, used to determine collections of training and testing environments. Formally, we consider context sets $C_{train}$ and $C_{test}$, where the policy is trained on training context-set CMDP $\mathcal{T}\vert_{C_{train}}$ and evaluated on the testing context-set CMDP $\mathcal{T}\vert_{C_{test}}$. We denote the expected return in the CMDP as $\mathbb{R}(\pi, \mathcal{T}):=\mathbb{E}_{c\sim p(c)}[\mathcal{R}(\pi, \mathcal{T}\vert_{c})]$. The objective for the policy $\pi$ is to maximize the expected return on the testing context set, $\bold{R}(\pi, \mathcal{T}\vert_{C_{test}})$. Furthermore, the generalization gap between train and test performance can be measured by $\bold{R}(\pi, \mathcal{T}\vert_{C_{train}}) - \bold{R}(\pi, \mathcal{T}\vert_{C_{test}})$. 
\\ \\
In Procgen, we consider \textit{in-distribution} generalization on unseen environment contexts, where $C_{train}$ and $C_{test}$ are disjoint sets composed of i.i.d samples of $C$. In particular, $c$ is a random seed which determines how the game level procedurally generates. For the easy difficulty setting of Procgen, $C_{train}$ is composed of 200 fixed seeds, while $C_{test}=C$ is uniform over all seeds. 
\\ \\
In MiniHack, we consider \textit{out-of-distribution} generalization, namely extrapolation along different environment factors. In addition to a random seed, any MiniHack environment instance is also determined by its \texttt{des-file}, which controls map layout as well as placement of environment features, monsters, and objects. For example, the MiniHack task Corridor-R5 has one \texttt{des-file} associated with it, while KeyRoom-S5 has defined its own separate one. Thus, each (train, test) task pair tests extrapolation along environment variations such as room size, number of rooms, obstacles, or lighting. We refer the reader to Appendix C of the MiniHack paper for further details on variation~\cite{samvelyan2021minihack}, and Appendix~\ref{appendix:full_minihack_sequence} of our paper for the full list of MiniHack task pairs we use. Similarly, CHORES Gen-MultiTraj evaluates out-of-distribution generalization on unseen factors such as room scene and object of interest.
\\ \\
For a continual reinforcement learning setup, we further consider a sequence of $N$ tasks, $\mathcal{S}_N:=(\mathcal{T}_0 \ldots \mathcal{T}_{N-1})$. The agent trains on task $\mathcal{T}_i$ at timesteps in the interval $[A_i, B_i)$, where $A_i$ and $B_i$ are the task boundaries denoting the start and end, respectively, of task $\mathcal{T}_i$. We cycle through the tasks $M$ times, so the full task sequence $\mathcal{S}_{NM}$ has length $N\cdot M$. We assume that the tasks are drawn from some world collection $\mathcal{W}$, where the dimensions of the observation space and action space are consistent across all tasks from $\mathcal{W}$. This enables us to train one model, with a single output policy layer, over the task sequence. In this work, we consider $\mathcal{W}$ to be Atari games, Procgen games, NetHack dungeons for MiniHack, or AI2-THOR scenes for CHORES. We further assume that tasks in $\mathcal{W}$ have some shared structure, for instance related to dynamics (ie. enemies hurt players) or rewards (ie. survive longer), which humans would also learn to exploit in order to perform all tasks in $\mathcal{W}$.
\\ \\
There are a couple ways to view the benefits of defining the learning problem in this sequential manner, compared to training an expert for each task or training via multi-task learning. The first is through the lens of a robotic agent, operating in the real world in a way similar to humans. Such an agent should learn and adapt to new settings as they are encountered, without forgetting prior learned behavior. Training multiple tasks in parallel or training on each task individually with human-specified boundaries makes additional assumptions, which encourages the development of methods that are ill-suited to the needs of lifelong robotic agents in the real-world.

The second way to view the benefits of continual RL is that sequentially learning tasks is a simple and effective way to induce a \textit{non-stationary} learning process. Any component of the MDP may change on task switch, and the agent should be capable of handling such distribution shifts. While our benchmarks are not imbued with all the ways in which the real world may change, we see this way of modeling the learning problem as a step towards the final goal of real-world embodied agents, for future work to build off.



\clearpage

\section{CORA Details}
\label{section:appendix}

All code, including hyperparameters, is available here: \github.

We divide additional details for CORA in this section into: design objectives for CHORES (Appendix~\ref{sec:chores_design_objectives}), details of each CHORES task sequence (Appendix~\ref{sec:chores_experiment_details}), list of MiniHack task pairing (Appendix~\ref{appendix:full_minihack_sequence}), examples of initial observations for video-game task sequences (Appendix~\ref{appendix:initial_obs}), experiment runtimes (Appendix~\ref{sec:experiment_durations}), baseline implementation details (Appendix~\ref{appendix:implementation_differences}), hyperparameters (Appendix~\ref{appendix:hyperparameters}), Atari experiment results (Appendix~\ref{results:atari}), additional Procgen figures (Appendix~\ref{appendix:old_procgen_results}), final performance tables on all environments (Appendix~\ref{sec:final_perf_tables}), and avenues for future work (Appendix~\ref{sec:future_work}).

\subsection{CHORES design objectives}
\label{sec:chores_design_objectives}

\begin{table}[h]
    \tiny
    \centering
    \begin{tabular}{lcccccc}
         \toprule
          & & & Num traj. & & & \\
          & Difficulty & Test Type & per task & Scene & Task & Object  \\
         \midrule
         Mem-VaryScene & easier & memorization & 1 & $\Delta$, bath & put in bathtub & hand towel \\
         Mem-VaryTask & easier & memorization & 1 & bath, Room 402 & $\Delta$ & toilet paper (TP) \\
         Mem-VaryObject & easier & memorization & 1 & kitchen, Room 24 & clean object & $\Delta$ \\
         Gen-MultiTraj & harder & generalization & 3 & $\Delta$, kitchen & Cool \& put in sink & $\Delta$\\
         \bottomrule
    \end{tabular}
    \vspace{1em}
    \\
    \begin{tabular}{lccc}
         \toprule
           & Task A & Task B & Task C\\
         \midrule
         Mem-VaryScene & Room 402 ($r=12$) & Room 419 ($r=12$) & Room 423 ($r=12$) \\
         Mem-VaryTask & hang TP ($r=12$) & put 2 TP in cabinet ($r=24$) & put 2 TP on counter ($r=24$) \\
         Mem-VaryObject & fork ($r=18$) & knife ($r=18$) & spoon ($r=18$) \\
         Gen-MultiTraj & Room 19, cup ($r=18$) & Room 13, sliced potato ($r=31$) & Room 2, sliced lettuce ($r=31$) \\
         \bottomrule
    \end{tabular}
    \vspace{1em}
    \caption{Summary of the four CHORES benchmarks. The first three are memorization tasks, and are evaluated on the training environment. The fourth is a harder generalization task, with 3 trajectories per task to initialize the scene and task parameters. We also summarize which scene each task is in, what task it performs, and what objects it utilizes. We categorize each CHORES by what the task sequences varies. The $r$ values in parentheses show the minimum return for solving the task.}
    \label{tab:chore_summary}
\end{table}

\textbf{Goal communication} All CORA benchmarks other than CHORES use video game environments, where the visual differences between the tasks may have been sufficient for the agent to know what they are supposed to do, in order to receive reward. For instance in Atari, 0-SpaceInvaders is distinct enough in appearance from 2-BeamRider that no further task specification is required, see Appendix~\ref{appendix:initial_obs} Figure~\ref{fig:figure_crl_env_viz_atari}. However, since all CHORES take place in a fixed set of rooms, the observation that the agent receives on its own is insufficient to distinguish task boundaries with. 
In this work, we use subgoal images in CHORES to communicate task intentions to the agent. In real-world settings, this could be achieved by a human demonstrating a task and taking pictures at critical points during the task to give to a robotic agent. Future work may leverage the language annotations ALFRED provides with each demonstration trajectory for alternate as more convenient forms of communication would be useful for robotic agents to employ.

\textbf{Task constraints}
To make the benchmark as accessible for the community, our aim was for each task used by CHORES to be individually solvable in under five hours using a machine with 16 vCPUs, 64 GB of RAM, and a Titan X GPU. Given the nature of simulating realistic environments, this corresponds to a budget of around 1 million frames per task. Additionally, since continual RL ultimately should be deployed onto robotic agents in the real world, modest sample budgets align with what will likely be feasible with real world learning.

Most existing policies may not be sample efficient enough to learn complex tasks in this amount of time. However, by providing sequences of simple tasks that are at the edge of what is currently achievable, we hope to move beyond this boundary and encourage the development of algorithms that are successful under these conditions. We also provide one complex task as an example of what is possible moving forward and for what we hope will be achievable in future CHORES benchmarking.

\textbf{Task selection}
\label{sec:task_selection}
The CHORES tasks were (by necessity) somewhat more hand-picked. These were selected in the following way:
\begin{enumerate}
    \item[1.] We used ALFRED to generate a new set of trajectories for the latest AI2-THOR version (needed for headless rendering to use on our cluster) using ALFRED's defined set of tasks.
    \item[2.] Based on our defined axes of variation (e.g. varying objects), we filtered successfully generated tasks into clusters that met our criteria.
    \item[3.] From this filtered set, we selected tasks to maximize diversity (e.g. more than just pick-and-place). 
\end{enumerate}
The selection process was done more out of necessity than the ideal, but we believe the tasks cover the desired goals of the benchmark more than adequately.


\subsection{CHORES details}
\label{sec:chores_experiment_details}

\begin{figure}[h]
    \centering
    \subfloat[Go to counter]{
    \includegraphics[width=0.2\textwidth]{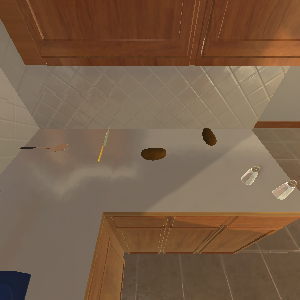}}
    \hspace{1em}
    \subfloat[Pick up knife]{
    \includegraphics[width=0.2\textwidth]{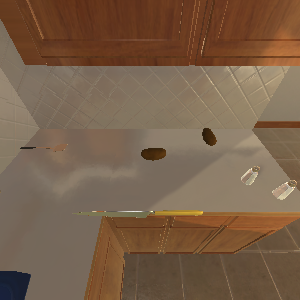}}
    \hspace{1em}
    \subfloat[Slice potato]{
    \includegraphics[width=0.2\textwidth]{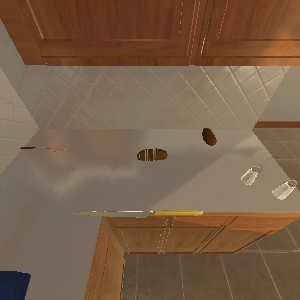}}
    \hspace{1em}
    \subfloat[Go to sink]{
    \includegraphics[width=0.2\textwidth]{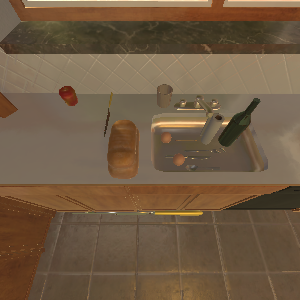}}\\
    
    \subfloat[Put knife in sink]{
    \includegraphics[width=0.2\textwidth]{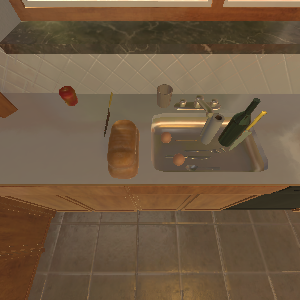}}
    \hspace{1em}
    \subfloat[Go to counter]{
    \includegraphics[width=0.2\textwidth]{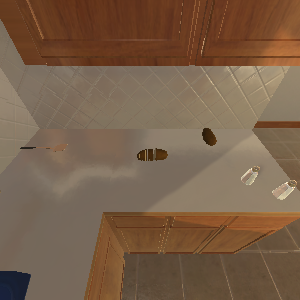}}
    \hspace{1em}
    \subfloat[Pick up potato]{
    \includegraphics[width=0.2\textwidth]{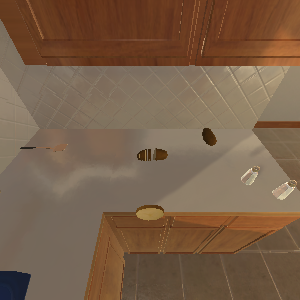}}
    \hspace{1em}
    \subfloat[Go to fridge]{
    \includegraphics[width=0.2\textwidth]{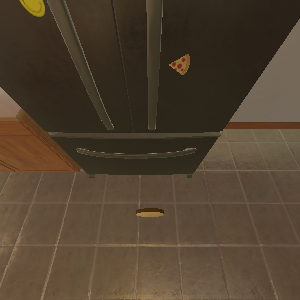}} \\
    
    \subfloat[Cool potato]{
    \includegraphics[width=0.2\textwidth]{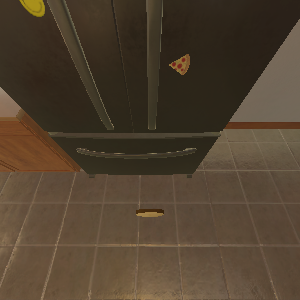}}
    \hspace{1em}
    \subfloat[Go to sink]{
    \includegraphics[width=0.2\textwidth]{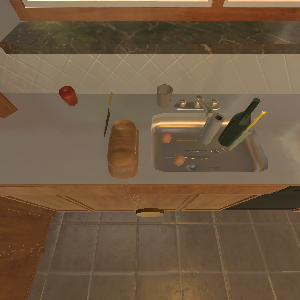}}
    \hspace{1em}
    \subfloat[Put potato in sink]{
    \includegraphics[width=0.2\textwidth]{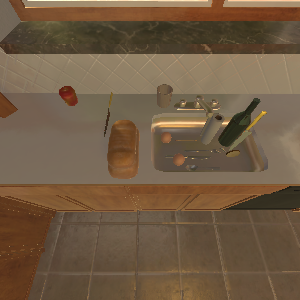}}
    \hspace{2em}
    \caption{Visualization of one ALFRED trajectory used to define a task in CHORES.}
    \label{fig:chores_subgoals}
\end{figure}

To start, we propose three memorization-based CHORES, each of which tests an agent's robustness to a particular type of domain shift. Each task within these three CHORES is intended to be relatively easy, using only one trajectory to set the environment parameters and evaluating in the same scene as during training. We additionally propose one harder CHORES to evaluate generalization. This last CHORES has more complex tasks, with a set of three trajectories per task to initialize the environment from, and evaluation is also done in unseen settings from a different 3 trajectories. These benchmarks are summarized in Table~\ref{tab:chore_summary} in the Appendix. In all cases, the locations of moveable, interactable objects are randomized between trajectories. 



The first task sequence, which we refer to as \textbf{Mem-VaryRoom}, keeps the task type and task object the same, while changing the room scene the agent interacts in to different bathrooms. The agent is trained to find a hand towel and place it in the bath tub of Room 402 for 1M steps, then in Room 419, then in Room 423. We then cycle through the environments again, to evaluate how much faster learning each environment is the second time.

The second task sequence, \textbf{Mem-VaryTask}, follows the same pattern but holds the current room and object constant, while changing the task. The agent is trained in the same bathroom to change a roll of toilet paper on a hanger, then to put two rolls of toilet paper in the cabinet, and finally to place two rolls on the countertop.

The third task sequence, \textbf{Mem-VaryObject}, holds the current room and task constant but changes the object. In kitchen 24 the agent is tasked to clean a fork, then clean a knife, then clean a spoon. Cleaning is done by putting an object under running water from a faucet. For the first two tasks, after cleaning the agent must put the object on the counter top, and in the third it must put it in the cabinet. 

The fourth task sequence, \textbf{Gen-MultiTraj}, uses a task where an agent takes an object, puts it in the fridge to cool it, removes it, and then places it in the sink. With this base task, the task sequence is as follows: (a) in kitchen 19, the agent performs the task with a cup; (b) in kitchen 13, the agent must slice a potato, then perform the task with the sliced potato; (c) in kitchen 2, the agent must slice lettuce, then perform the task with the sliced lettuce. The key difference from the previous task sequences is that each task in the fourth CHORES is evaluated on unseen settings initialized from three possible heldout demonstrations trajectories, testing an agent's ability to generalize.

In Figure \ref{fig:chores_subgoals}, we visualize all subgoal images for one trajectory of the Gen-MultiTraj potato task (task 2).

\textbf{Reward details}
Unlike ALFRED which reports the number of subgoals achieved, we report the episode returns for consistency with the other benchmarks, clipped to a minimum value of -10. Extremely negative values occur when the agent performs a particularly suboptimal action for the duration of the episode, until the maximum step limit of 1000 is hit. Without clipping, this occasional negative behavior completely drowns out the agent's successes, both in visualization and metrics.

\subsection{MiniHack task sequence}
\label{appendix:full_minihack_sequence}

The MiniHack (train, test) paired task sequence we use is:
\allowdisplaybreaks 
\begin{align*}
     0 & & \textrm{(Room-Random-5x5, Room-Random-15x15)} \\
     1 & & \textrm{(Room-Dark-5x5, Room-Dark-15x15)} \\
     2 & & \textrm{(Room-Monster-5x5, Room-Monster-15x15)} \\
     3 & & \textrm{(Room-Trap-5x5, Room-Trap-15x15)} \\
     4 & & \textrm{(Room-Ultimate-5x5, Room-Ultimate-15x15)} \\
     5 & & \textrm{(Corridor-R2, Corridor-R5)} \\
     6 & & \textrm{(Corridor-R3, Corridor-R5)} \\
     7 & & \textrm{(KeyRoom-S5, KeyRoom-S15)} \\
     8 & & \textrm{(KeyRoom-Dark-S5, KeyRoom-Dark-S15)} \\
     9 & & \textrm{(River-Narrow, River)} \\
    10 & & \textrm{(River-Monster, River-MonsterLava)} \\
    11 & & \textrm{(River-Lava, River-MonsterLava)} \\
    12 & & \textrm{(HideNSeek, HideNSeek-Big)} \\
    13 & & \textrm{(HideNSeek-Lava, HideNSeek-Big)} \\
    14 & & \textrm{(CorridorBattle, CorridorBattle-Dark)} 
\end{align*}

This task sequence defines 15 pairs of (train, eval) environments and uses a total of 27 different environments. Some of the evaluation environments are used multiple times because that those test environments have more than one related train environment. In particular, this impacts task pairs 5 and 6; 10 and 11; 12 and 13. 


\subsection{Examples for initial observations of the video game benchmarks}
\label{appendix:initial_obs}

\begin{figure}[H]
    \centering
    \includegraphics[width=1.0\textwidth,trim={0 0 0 0},clip,scale=1.0]{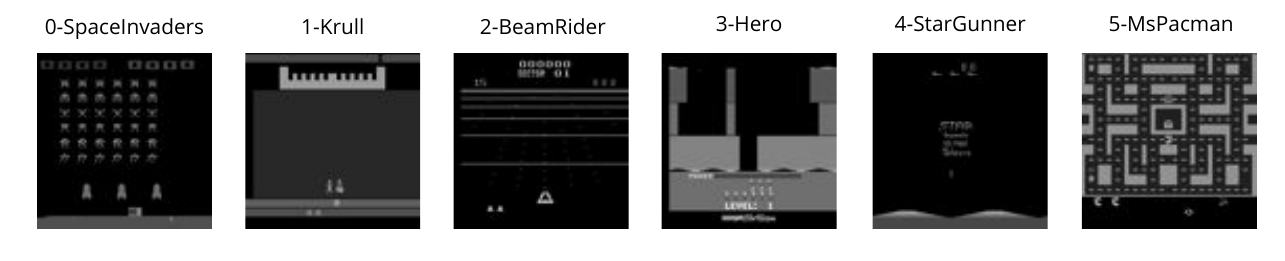}
    \caption{Examples of initial observations for each task in the 6 task Atari sequence.}
    \label{fig:figure_crl_env_viz_atari}
\end{figure}

\begin{figure}[H]
    \centering
    \includegraphics[width=1\textwidth,trim={0 0 0 0},clip,scale=1.0]{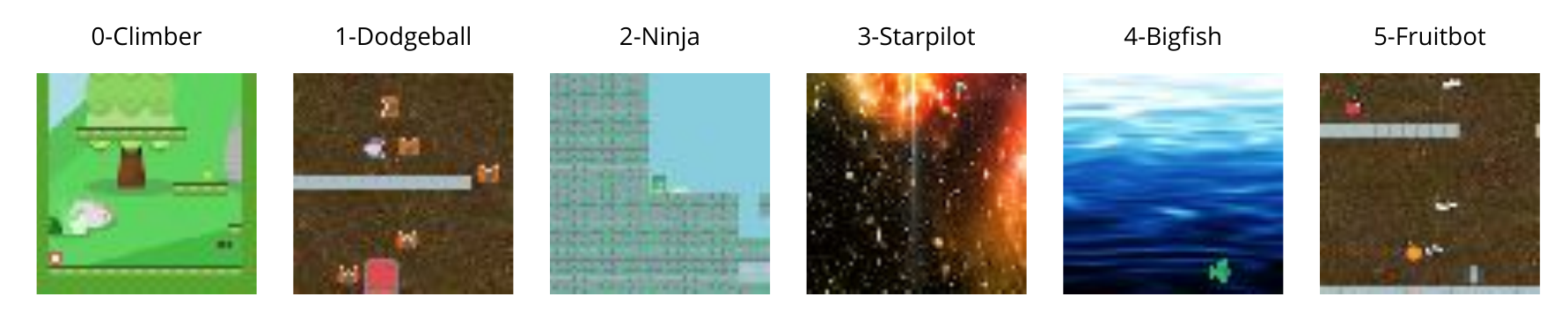}
    \caption{Examples of initial observations for each task in the 6 task Procgen sequence}
    \label{fig:figure_crl_env_viz_procgen}
\end{figure}

\begin{figure}[H]
    \centering
    \includegraphics[width=0.8\textwidth,trim={0 0 0 0},clip,scale=1.0]{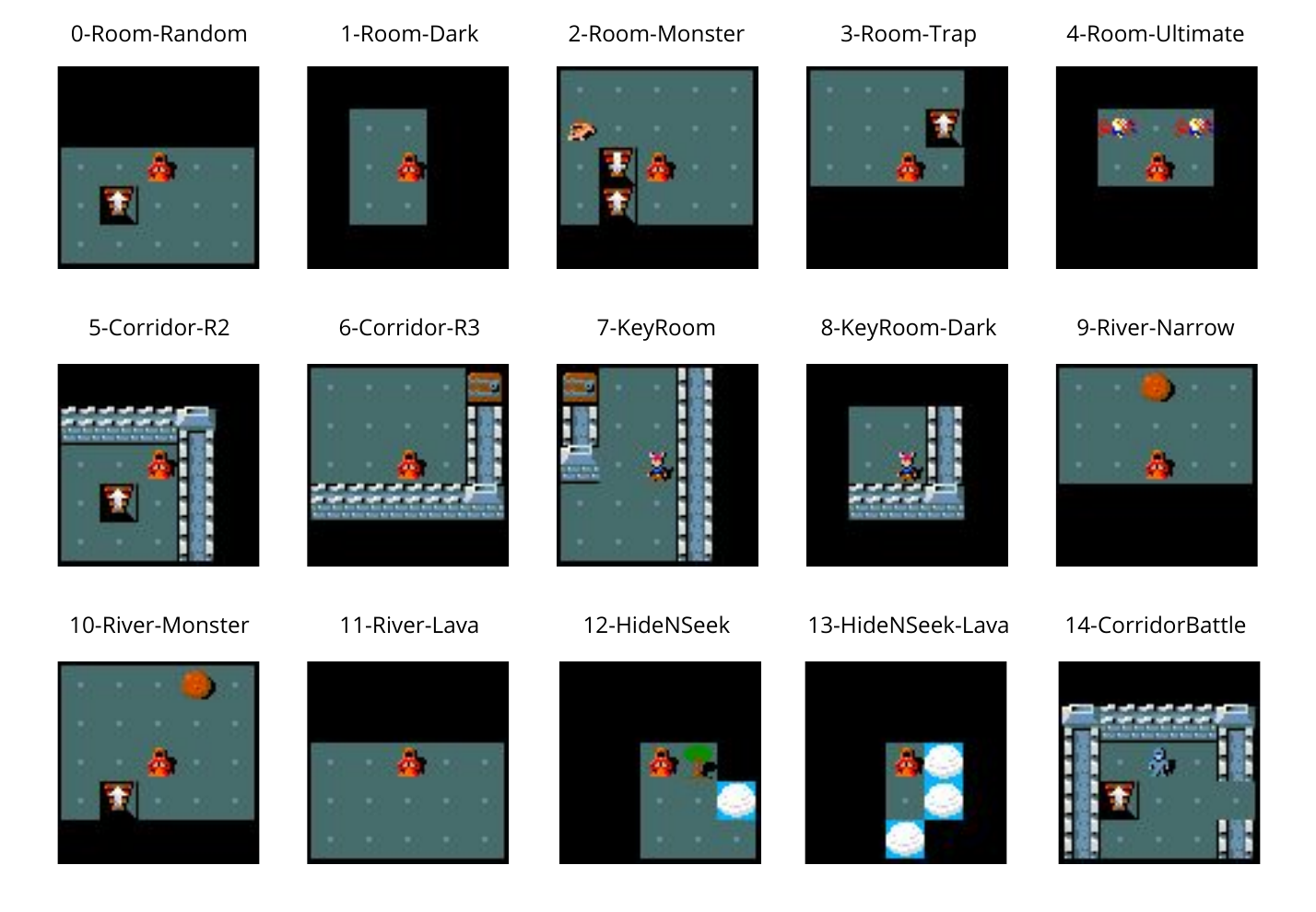}
    \caption{Examples of initial observations for each task in the 15 task MiniHack sequence. Observations are shown for the training task of each pair.}
    \label{fig:figure_crl_env_viz_minihack}
\end{figure}

We show examples of initial observations that an agent may get for each task in the Atari (Figure~\ref{fig:figure_crl_env_viz_atari}), Procgen (Figure~\ref{fig:figure_crl_env_viz_procgen}), and MiniHack (Figure~\ref{fig:figure_crl_env_viz_minihack}) task sequences. Note that visually, it is easy to distinguish the different tasks for Atari and Procgen. However, since MiniHack tasks use the same visual assets, it is more challenging to tell tasks apart from each other. This makes task boundary identification more difficult for algorithms. Furthermore, this also supports why CHORES provides agents with a goal image specifying the task to perform.

\subsection{Baseline implementation details}
\label{appendix:implementation_differences}

After discussion with the authors, the original P\&C~\cite{schwarz2018progress} results collect the task's budgeted number of samples from the environment for each of their progress and compress phases. Instead, we opt to collect half for each, rather than effectively doubling the budget for environment steps compared to other methods. This may be why we observe P\&C underperforming compared to the original paper's results. Note that P\&C evaluates using the knowledge base model, which is only updated in the second half of training on each task. As such, for the first half of each task (while the active column model is being updated), essentially a flat performance curve is reported for P\&C.

Additionally, as stated in Appendix~\ref{results:atari}, the original authors used a pre-release version of IMPALA, while we use the TorchBeast implementation of IMPALA. For CLEAR, we used a 25M frame replay buffer while the original CLEAR paper used a replay buffer of half the number of frames that the agent trains on. For two cycles, this would correspond to 300M frames, while the authors of CLEAR used a replay buffer of 750M frames for their Atari experiment running 5 learning cycles. Due to system constraints, we are unable to run with the larger replay buffer size. Even with this difference, our implementation of CLEAR outperforms the results reported in the original paper. CLEAR uses reservoir sampling~\cite{isele2018selective} to maintain a buffer that stores a uniform sample of all past experience. 

We use a \textit{single} output head across the task sequence, rather than separate heads for each task such as in~\cite{wolczyk2021continual}. We evaluate using the eval mode of policies. For all IMPALA-based methods, eval mode takes the argmax action instead of a stochastic sampling. In the case of P\&C, we evaluate using the output of the knowledge base instead of the active column. While we use existing methods in the manner in which they were designed, we would like to encourage future methods to have parity between train and evaluation policies.

\textbf{Network architecture} For Atari (Appendix~\ref{results:atari}), we use the ``shallow'' model without an LSTM from the IMPALA paper~\cite{espeholt2018impala}. We also use this model for MiniHack (Section~\ref{section:minihack_results}) and CHORES (Section~\ref{section:chores_results}). For Procgen (Section~\ref{section:procgen_results}), we use the ``deep'' residual model without an LSTM from the IMPALA paper, following~\cite{jiang2020_prioritizedlevelreplay, cobbe2020procgen}. We include Procgen results from an earlier version of this paper that used the ``shallow'' model in Appendix~\ref{appendix:old_procgen_results}.

\subsection{Hyperparameters}
\label{appendix:hyperparameters}

\setlength\tabcolsep{1.5pt} 
\begin{table}[H]
    \small
    \centering
    \begin{tabular}{lccccc}
        \toprule
        Hyperparameter &&&&& Shared\\
        \midrule
        Num. actors &&&&& 64 \\
        Learner threads &&&&&  2 \\
        Batch size &&&&&  32 \\
        Unroll length &&&&&  20 \\
        Grad clip &&&&& 40 \\
        Reward clip &&&&& $[-1, 1]$ \\
        Normalize rewards &&&&& No \\
        Baseline cost &&&&& 0.5 \\
        Entropy cost &&&&& 0.01 \\
        Discount factor &&&&& 0.99 \\
        LSTM &&&&& No \\
        Network arch. &&&&& Nature CNN \\
        Learning rate &&&&& 4e${-4}$ \\ 
        Optimizer &&&&& RMSProp \\
        &&&&& $\alpha$ = 0.99 \\
        &&&&& $\epsilon$ = 0.01 \\
        &&&&& $\mu$ = 0 \\
        \bottomrule
        \toprule
        Hyperparameter & EWC & Online EWC & P\&C & CLEAR & \\
        \midrule
        EWC $\lambda$ & 10000 & 175 & 3000 & & \\
        EWC, min. task steps & 2e5 & 2e5 & & &\\
        Fisher samples & 100 & 100 & 100 & & \\
        Normalize Fisher & No & Yes & Yes \\
        Online EWC $\gamma$ & & 0.99 & 0.99 \\
        KL cost & & & 1.0 & & \\
        Batch ratio (novel-replay) & & & & 50-50 & \\
        Policy cloning cost & & & & 0.01 & \\
        Value cloning cost & & & & 0.005 & \\
        Replay buffer size & & & & 25e6 & \\
        \bottomrule
    \end{tabular}
    \caption{Hyperparameters for baselines on the Atari. For Atari, the network architecture is the Nature-CNN model from the DQN paper~\cite{mnih2015human}. For Procgen, we use the same values, except CLEAR's replay buffer size is reduced to 5e6, and across all baselines, we switch the network architecture to the ``deep'' residual model from the original IMPALA paper~\cite{espeholt2018impala}.}
    \label{tab:hyperparams_atari}
\end{table}

\setlength\tabcolsep{1.5pt} 
\begin{table}[H]
    \small
    \centering
    \subfloat[MiniHack]{
    \begin{tabular}{lccccc}
        \toprule
        Hyperparameter &&&&& Shared\\
        \midrule
        Reward clip &&&&& No \\
        Normalize rewards &&&&& Yes \\
        Entropy cost &&&&& 0.001 \\
        Discount factor &&&&& 0.999 \\
        Learning rate &&&&& 2e${-4}$ \\ 
        Optimizer &&&&& RMSProp \\
        &&&&& $\epsilon$ = 1e$-6$ \\
        \bottomrule
    \end{tabular}
    }
    \hspace{3em}
    \subfloat[CHORES]{
    \begin{tabular}{lccccc}
        \toprule
        Hyperparameter &&&&& Shared\\
        \midrule
        Num. actors &&&&& 10 \\
        Batch size &&&&&  10 \\
        Unroll length &&&&&  80 \\
        \bottomrule
    \end{tabular}
    }
    \label{tab:hyperparams_chores}

    \caption{Hyperparameters for baselines on (a) MiniHack, following values used by~\citet{samvelyan2021minihack}; (b) CHORES. We report values changed from those used for Atari.}
    \label{tab:hyperparams_chores_minihack_chores}
\end{table}

We report hyperparameters used by the baselines for Atari in Table~\ref{tab:hyperparams_atari}. Note that the scale of the losses differ between EWC vs. Online EWC and P\&C, see Appendix C.2 of the P\&C paper~\cite{schwarz2018progress}. For EWC, we conducted a hyperparameter search with learning rate from [1e$-5$, 1e$-4$, 4e$-4$, 6e$-4$] and EWC $\lambda$ from [1, 100, 175, 500, 1000, 1500, 3000, 5000, 10000]. For Online EWC and P\&C, we chose EWC $\lambda$ from [175, 3000, 10000] and Online EWC $\gamma$ from [0.95, 0.99, 0.999, 0.9999]. For CLEAR, we chose the replay buffer size from [5e6, 20e6, 25e6].
\\ \\
For Procgen, we use the same hyperparameters as for Atari, except CLEAR's replay buffer size is reduced to 5e6, and across all baselines we switch the network architecture to the ``deep'' residual model from the original IMPALA paper~\cite{espeholt2018impala}. For MiniHack, we use hyperparameters from~\citet{samvelyan2021minihack}, which are shown in Table~\ref{tab:hyperparams_chores_minihack_chores} (a). We also reduce CLEAR's replay buffer size to 10e6. For CHORES, we report the hyperparameters used in Table~\ref{tab:hyperparams_chores_minihack_chores} (b).

\subsection{Experiment runtimes}
\label{sec:experiment_durations}

When we run using a single GPU on a machine with 128 GB of RAM, and 32 or 40 vCPUs, we observe the following runtime averages across the entire experiments: 
\begin{itemize}[itemsep=0.05em, topsep=-0.1em]
    \item MiniHack: 45 hours (1.5 hr/task)
    \item Procgen: 54 hours (1.8 hr/task) 
    \item Atari: 129 hours (10.8 hr/task) -- which would have taken 323 hours to run the original 5-cycle experiment
    \item CHORES: discussed in Section~\ref{sec:chores_design_objectives}, around 5 hours per task to train.
\end{itemize}
Atari, as the standard baseline, is what we aimed to improve upon with the selection of Procgen and MiniHack, and indeed we see a speedup of around 6-7x. CHORES, which uses the visually realistic AI2THOR home simulation environment, is slower, but still 2x as fast as Atari per task.

We note that concurrent with our work, advances~\cite{weng2022envpool, moolib2022} have been made in accelerating the Atari simulator (and other CPU-based environments) using C++ threads and by bypassing the Python GIL. This has lowered the compute requirements for running longer Atari experiments, which were previously only possible with industry-level compute. We are excited for the possibilities opened by these advances, for future work on continual RL incorporating longer task sequences, and the more difficult Atari settings introduced by~\cite{machado2018revisiting, farebrother2018generalization}.

\subsection{Atari results}
\label{results:atari}

\begin{figure}[h]
    \centering
    \hspace{1.5em} 
    \includegraphics[trim=0 3em 18em 0, clip, width=0.25\textwidth]{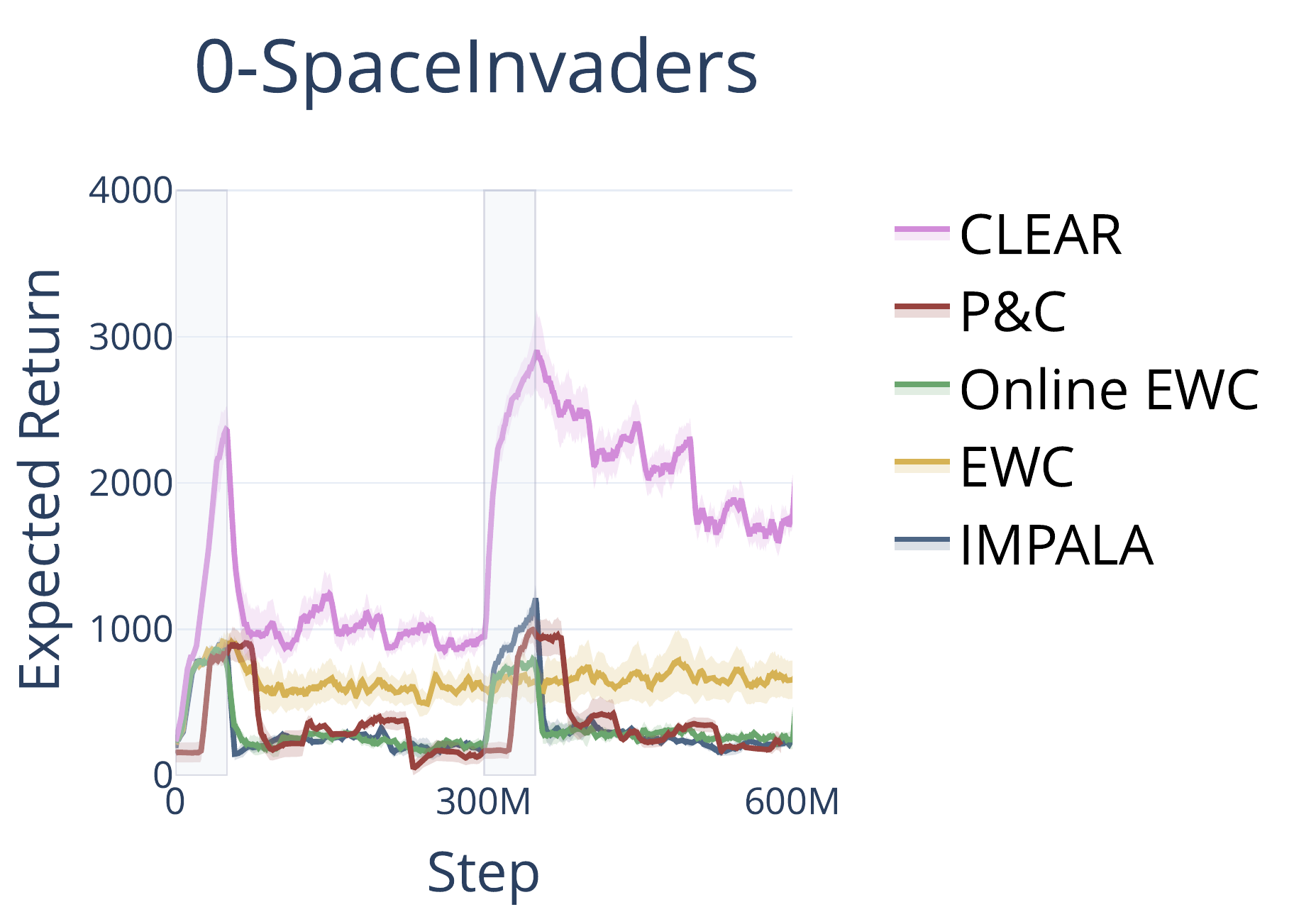}
    \includegraphics[trim=0 3em 18em 0, clip, width=0.25\textwidth]{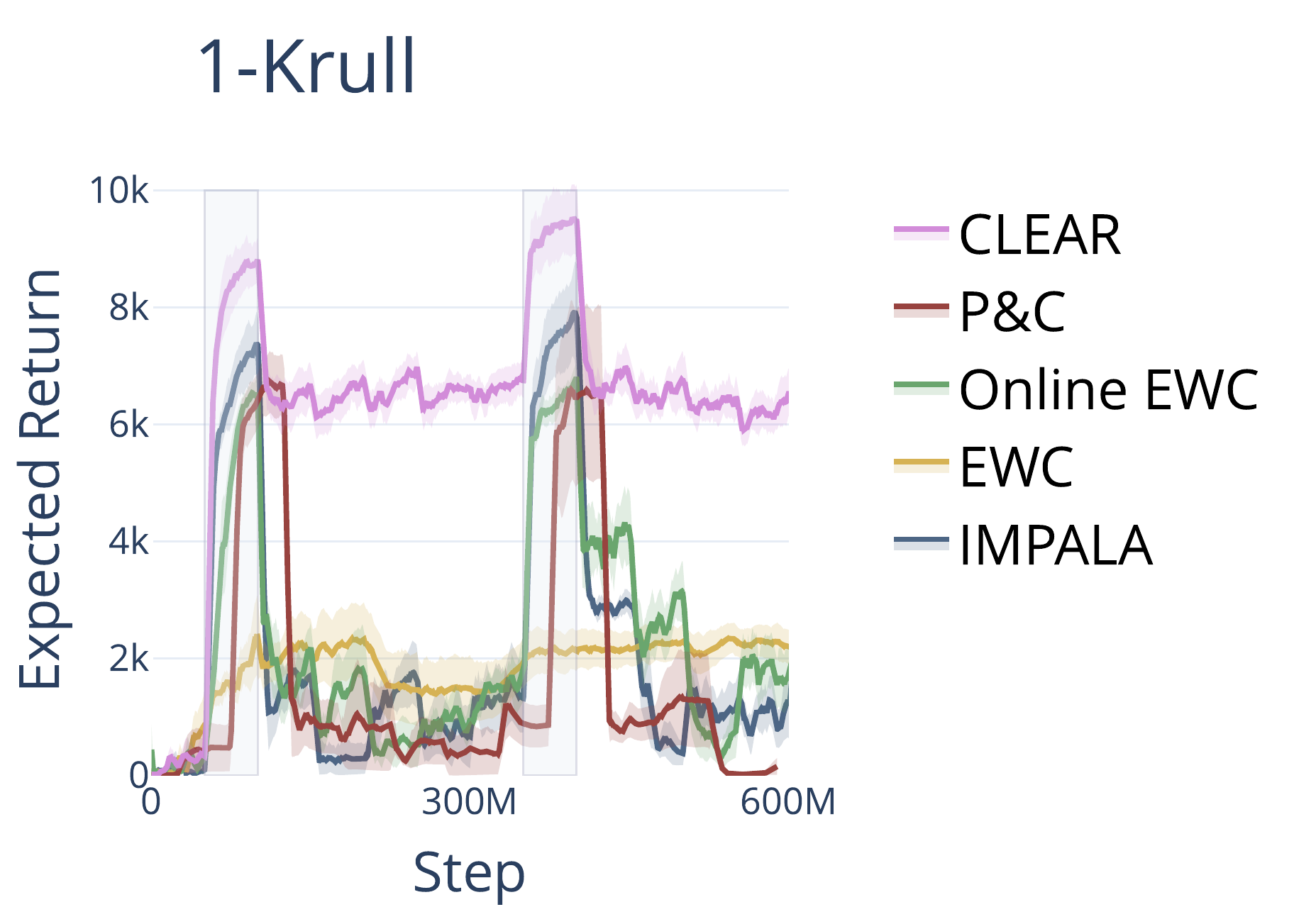}
    \includegraphics[trim=0 3em 0em 0, clip, width=0.38\textwidth]{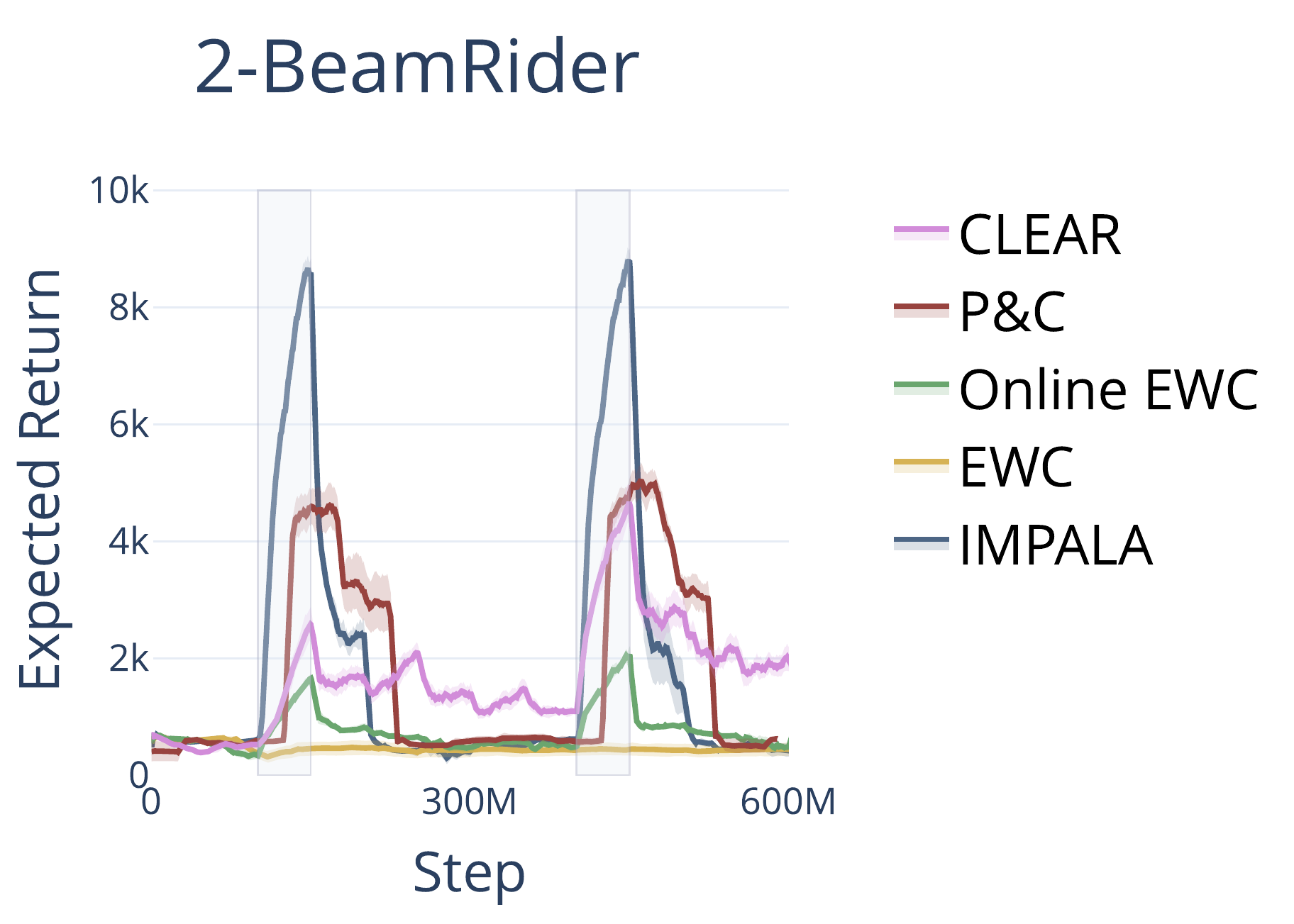} \\
    \includegraphics[trim=0 0em 18em 0, clip, width=0.25\textwidth]{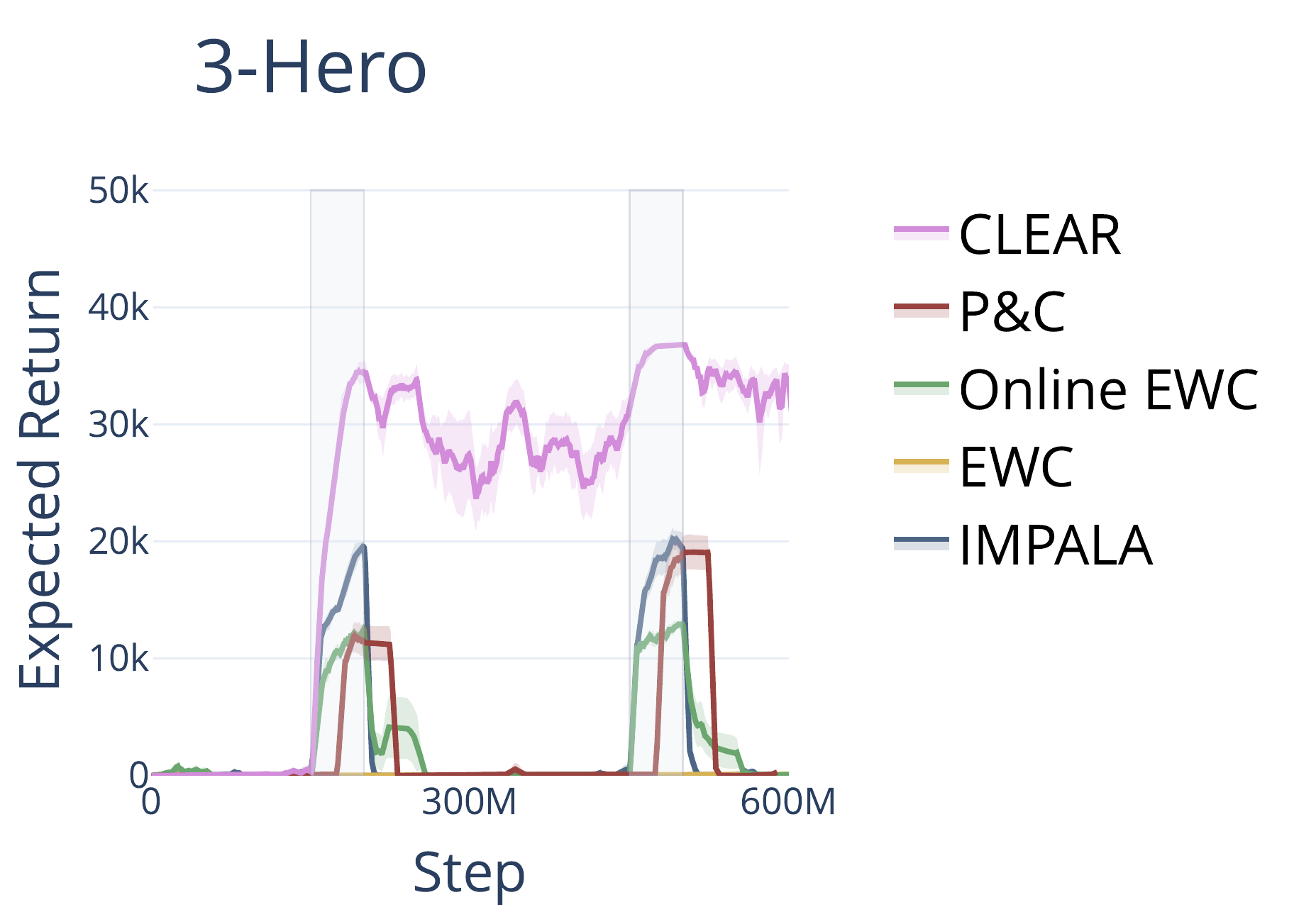}
    \includegraphics[trim=0 0em 18em 0, clip, width=0.25\textwidth]{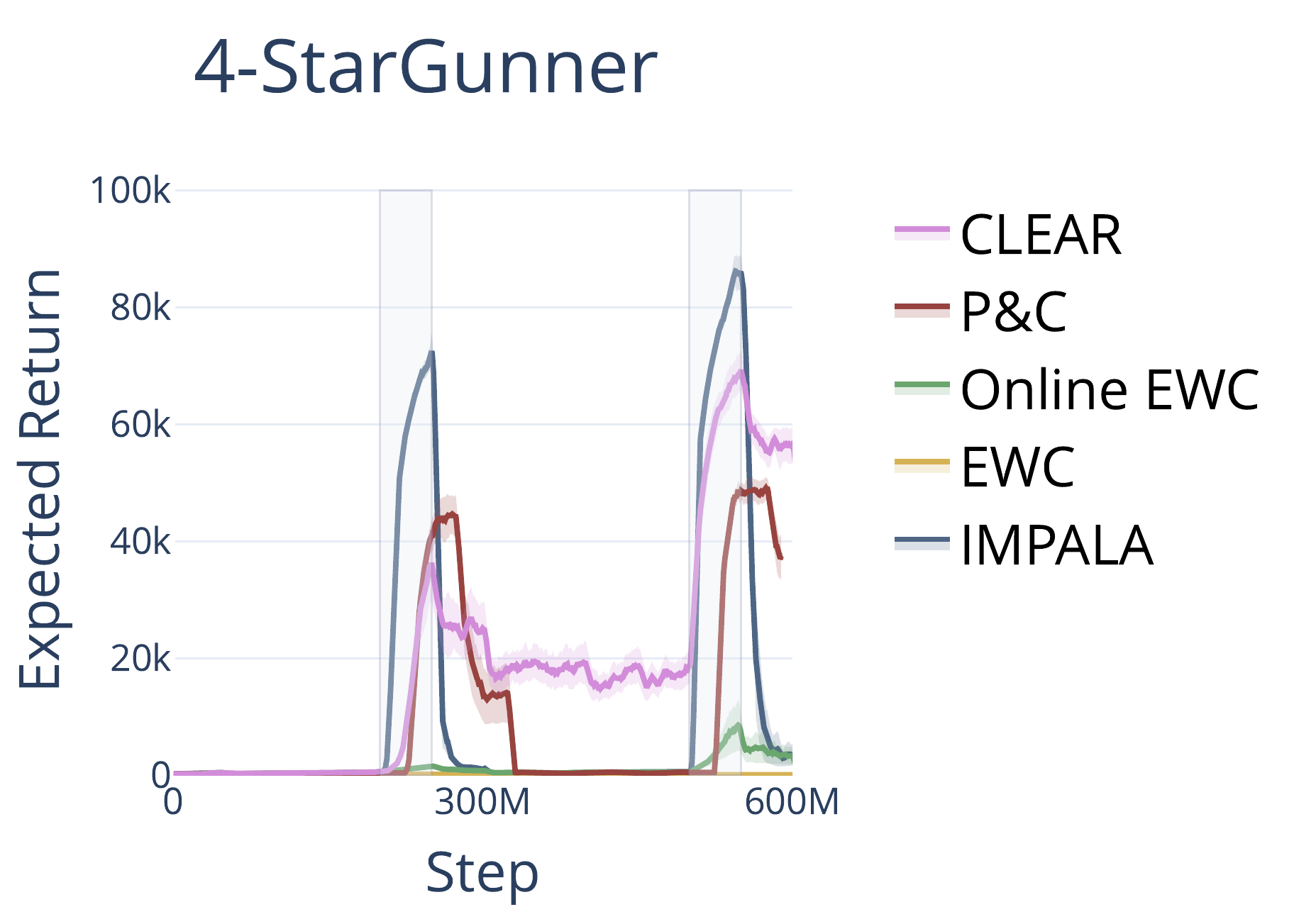}
    \includegraphics[trim=0 0em 18em 0, clip, width=0.25\textwidth]{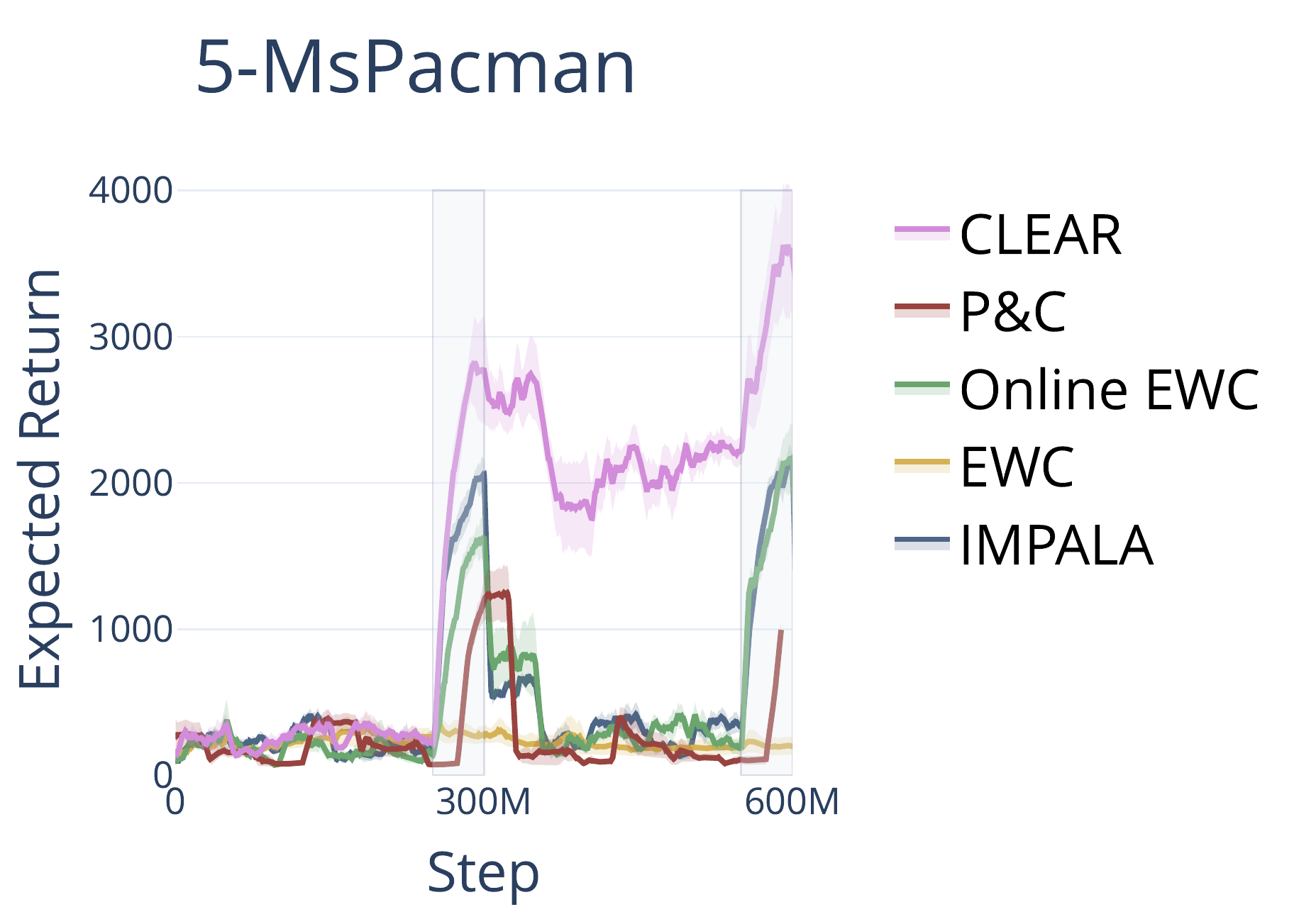}
    \hspace{4em}
    \caption{Results for Continual Evaluation $(\mathcal{C})$ on the 6 Atari task sequence from~\cite{rolnick2018clear, schwarz2018progress}. Due to compute constraints, we only train for 2 cycles compared to the original experiments which used 5 learning cycles. IMPALA is the baseline learning algorithm that the other methods for continual RL build off. Gray shaded rectangles show when the agent trains on each task.}
    \label{fig:reimplement_atari_results}
\end{figure}

We include the standard, proven Atari task sequence as a benchmark, in order to validate our baseline implementations on an existing standard. The reproduction of these Atari results were developed over hundreds of hours, including time spent analyzing papers for algorithm details, corresponding with the original authors, tuning hyperparameters, and running many seeds of Atari experiments, each of which takes hundreds of millions of frames. These results were reproduced using a university server cluster and several thousand dollars of AWS credits, compared to the industry-level compute that the original authors (from DeepMind)~\citet{schwarz2018progress} and~\citet{rolnick2018clear} had access to. This is one of the primary reasons we are advocating for more compute-friendly continual RL benchmarks. It is also the reason that we were only able to run 2 learning cycles for these Atari results instead of the intended 5 cycles.

We use the full 18-dim action space for this task sequence. The observation space is (84, 84) grayscale images, and the agent receives a framestack of 4. The Atari games used are fully deterministic, and following the prior continual RL work on Atari, we do not apply sticky actions~\cite{machado2018revisiting}. 

Atari results are shown in Figure~\ref{fig:reimplement_atari_results}, and we compare them against the results presented in~\citet{rolnick2018clear}. Notably, on almost all Atari tasks, our implementations outperform the results reported in CLEAR. This may be because we use the TorchBeast~\cite{torchbeast2019} implementation of IMPALA, while the results in~\citet{rolnick2018clear} and~\citet{schwarz2018progress} use an earlier, pre-release version of IMPALA. 

\textbf{Benchmark analysis:}
Summary metrics are available in Table~\ref{tab:summary_metrics}. From these, we observe that Atari does effectively test for robustness to catastrophic forgetting, but exhibits nearly no transfer. Looking at the diagnostic transfer in Table~\ref{tab:atari_transfer}, we observe no transfer, likely because the six Atari tasks used are too distinct from each other. 



\textbf{Algorithm design:}
From the continual evaluation results in Figure~\ref{fig:reimplement_atari_results}, we can see that CLEAR outperforms the other baselines at both at recall and plasticity on Atari, which matches the original results by~\cite{rolnick2018clear}. EWC maintains a flat return curve for early tasks, which is consistent, losing plasticity and failing to learn the later tasks. P\&C largely maintains its plasticity, but we observe considerably more forgetting than was reported. We discuss this disparity in Appendix~\ref{appendix:implementation_differences}.




\clearpage

\subsection{Additional Procgen Figures}
\label{appendix:old_procgen_results}

\begin{figure}[h]
    \centering
    \hspace{1.5em} 
    \includegraphics[trim=0 3.5em 19em 0, clip, width=0.25\textwidth]{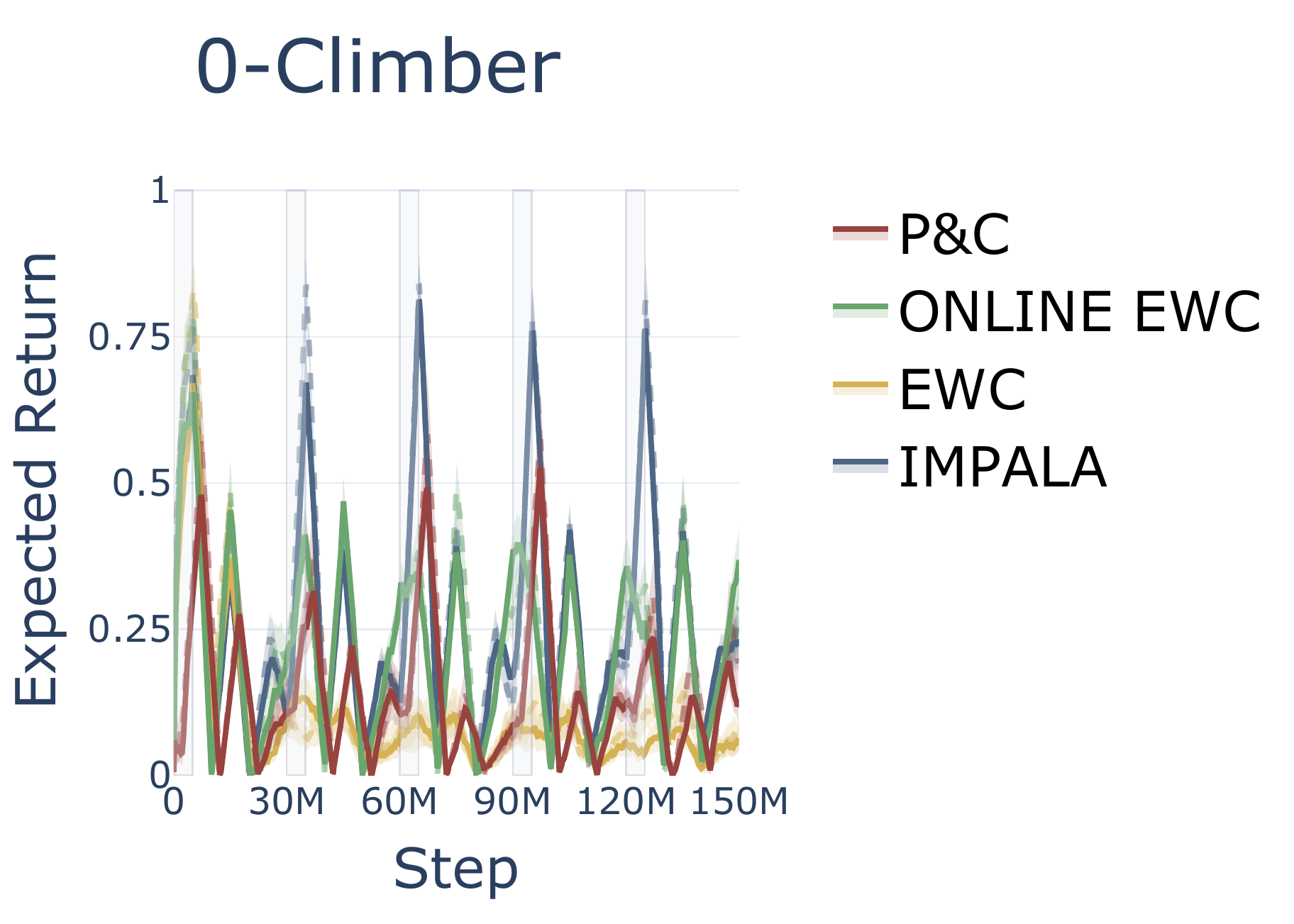}
    \includegraphics[trim=0 3.5em 19em 0, clip, width=0.25\textwidth]{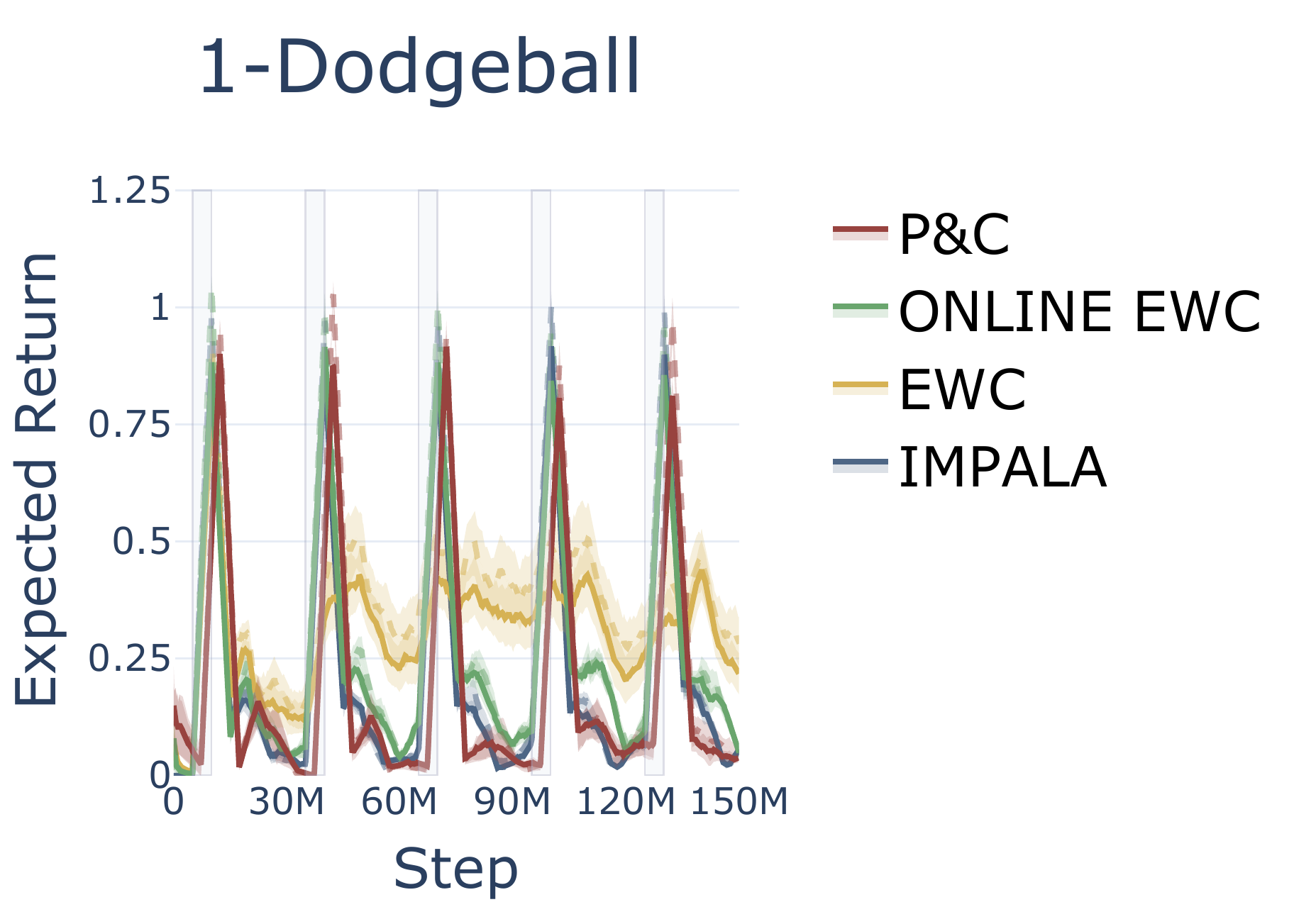}
    \includegraphics[trim=0 3.5em 0em 0, clip, width=0.39\textwidth]{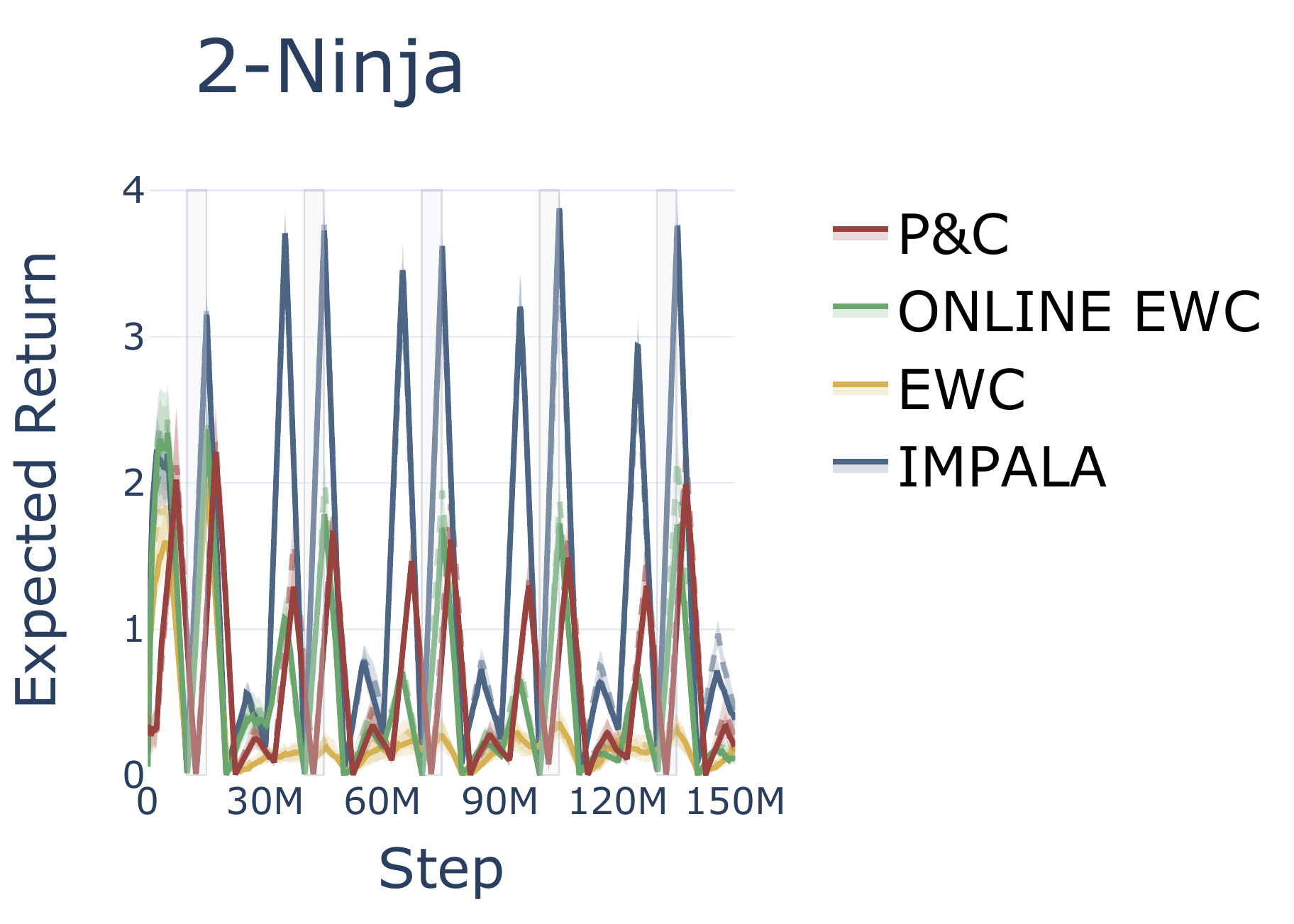} \\
    \includegraphics[trim=0 0em 19em 0, clip, width=0.25\textwidth]{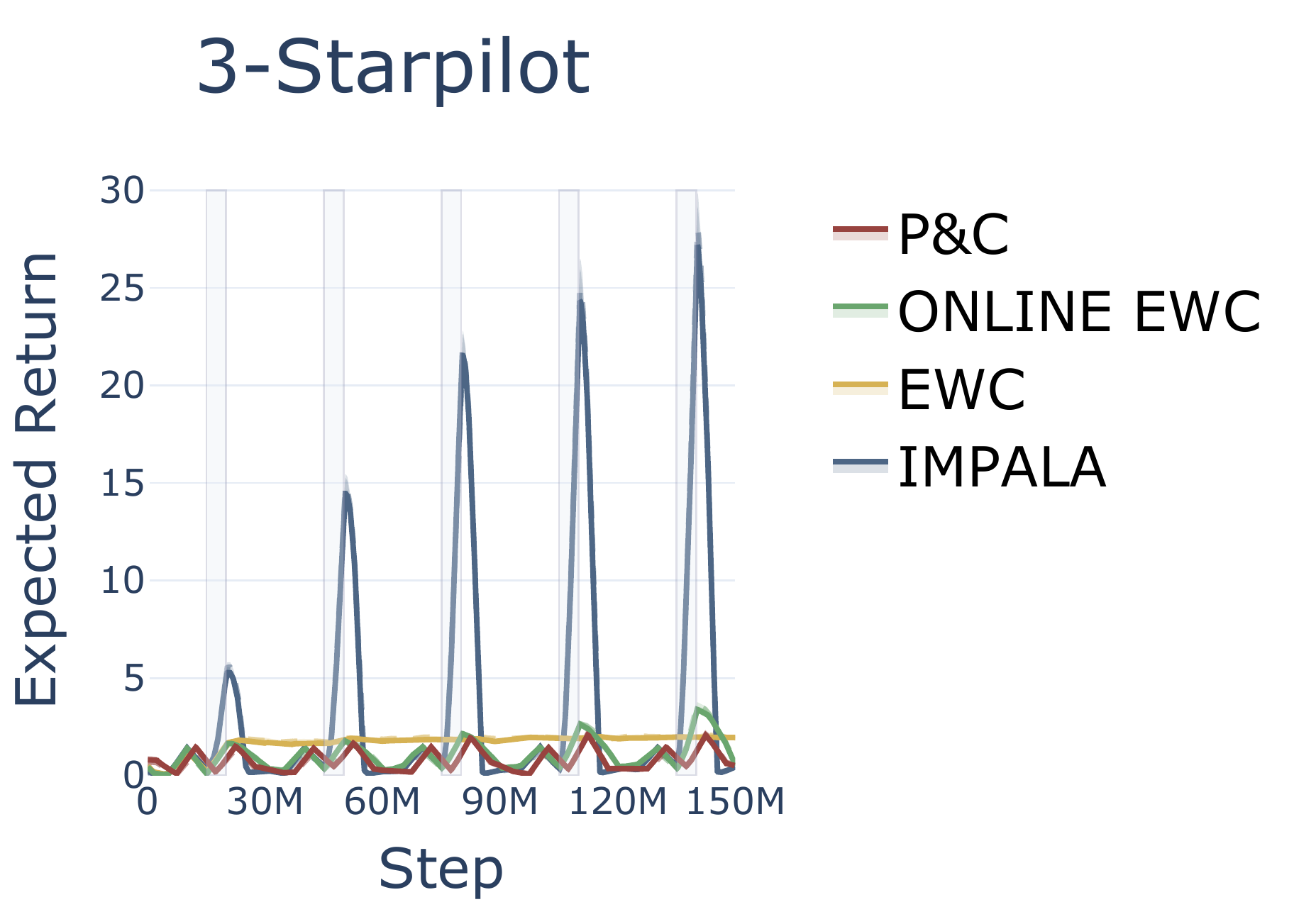}
    \includegraphics[trim=0 0em 19em 0, clip, width=0.25\textwidth]{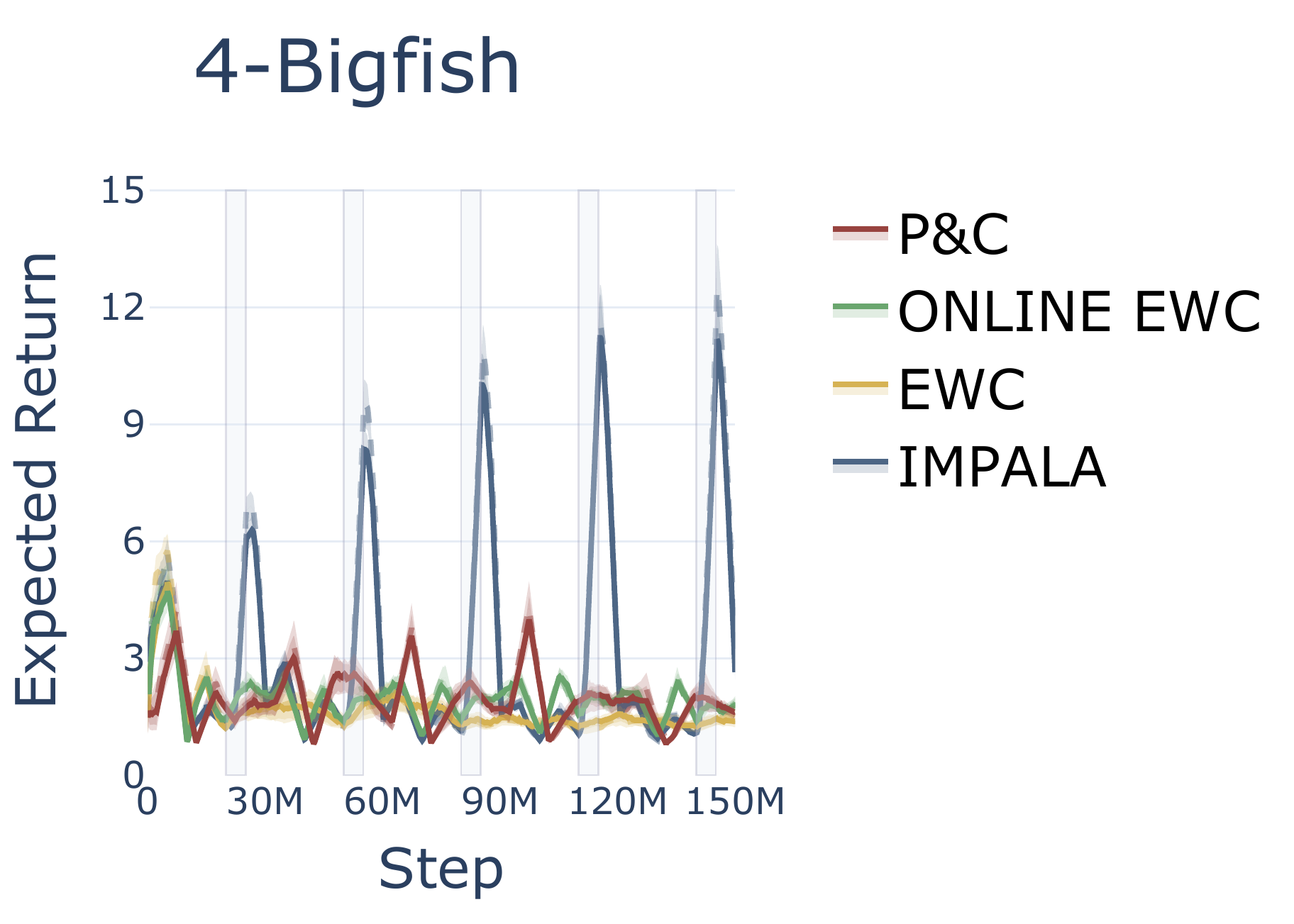}
    \includegraphics[trim=0 0em 19em 0, clip, width=0.25\textwidth]{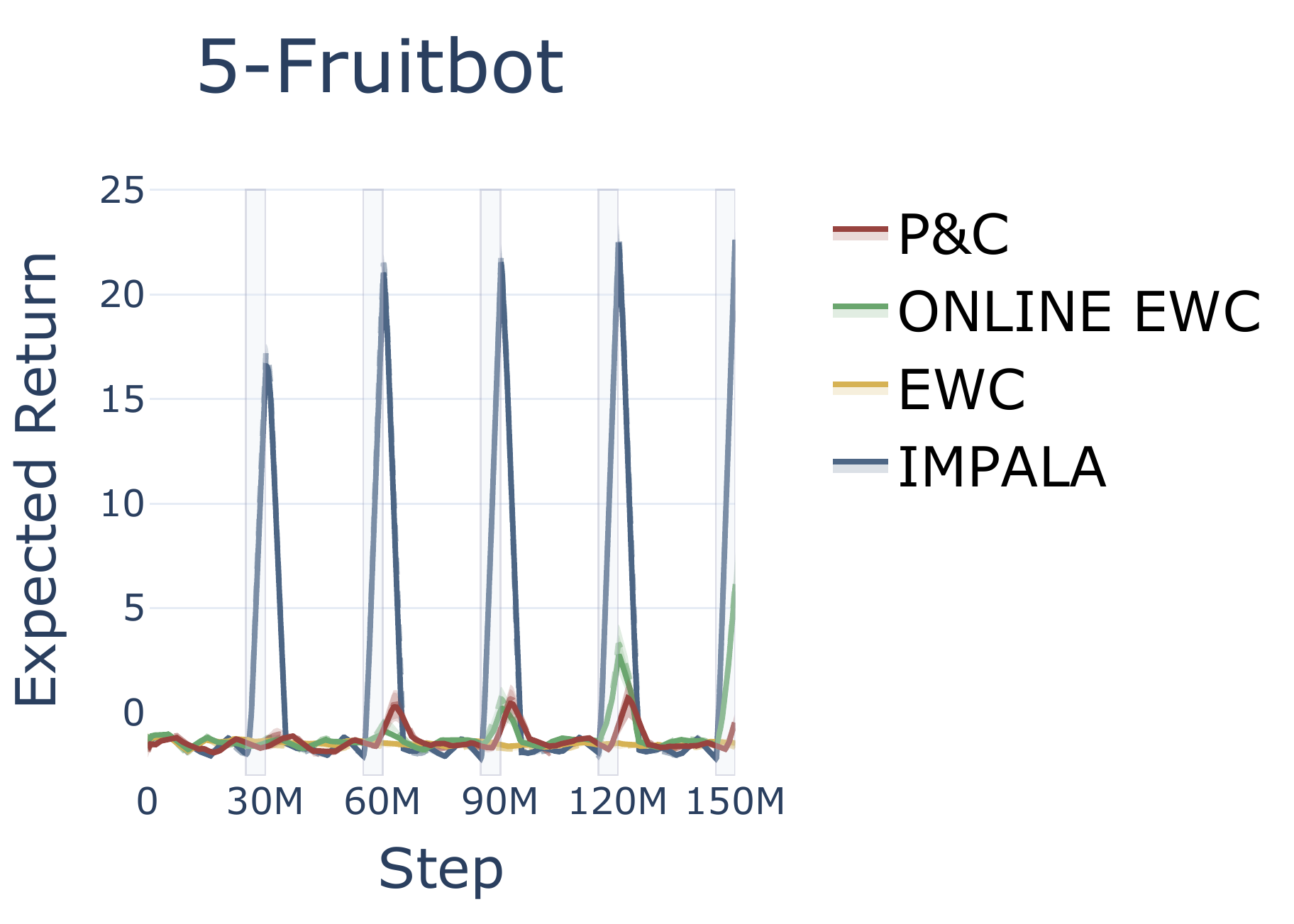}
    \hspace{4.5em}
    \caption{Results for Continual Evaluation $(\mathcal{C})$ on the 6 Procgen tasks. These are the same results as shown in Figure \ref{fig:procgen_results_res}, but with CLEAR removed and rescaled to better visualize the baselines.
    }
    \label{fig:procgen_results_noclear}
\end{figure}

\begin{figure}[h]
    \centering
    \hspace{1.5em} 
    \includegraphics[trim=0 3em 18em 0, clip, width=0.23\textwidth]{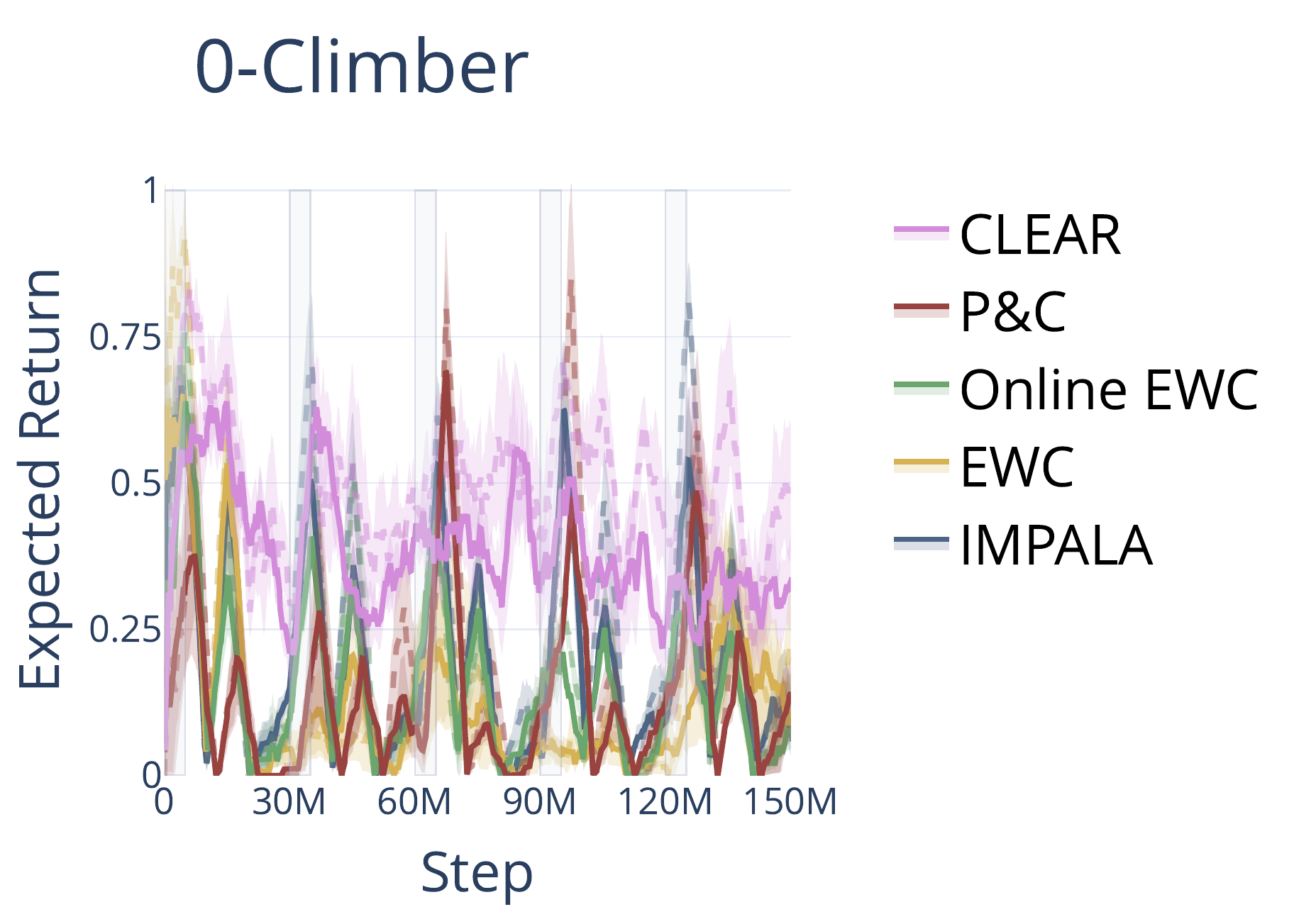}
    \includegraphics[trim=0 3em 18em 0, clip, width=0.23\textwidth]{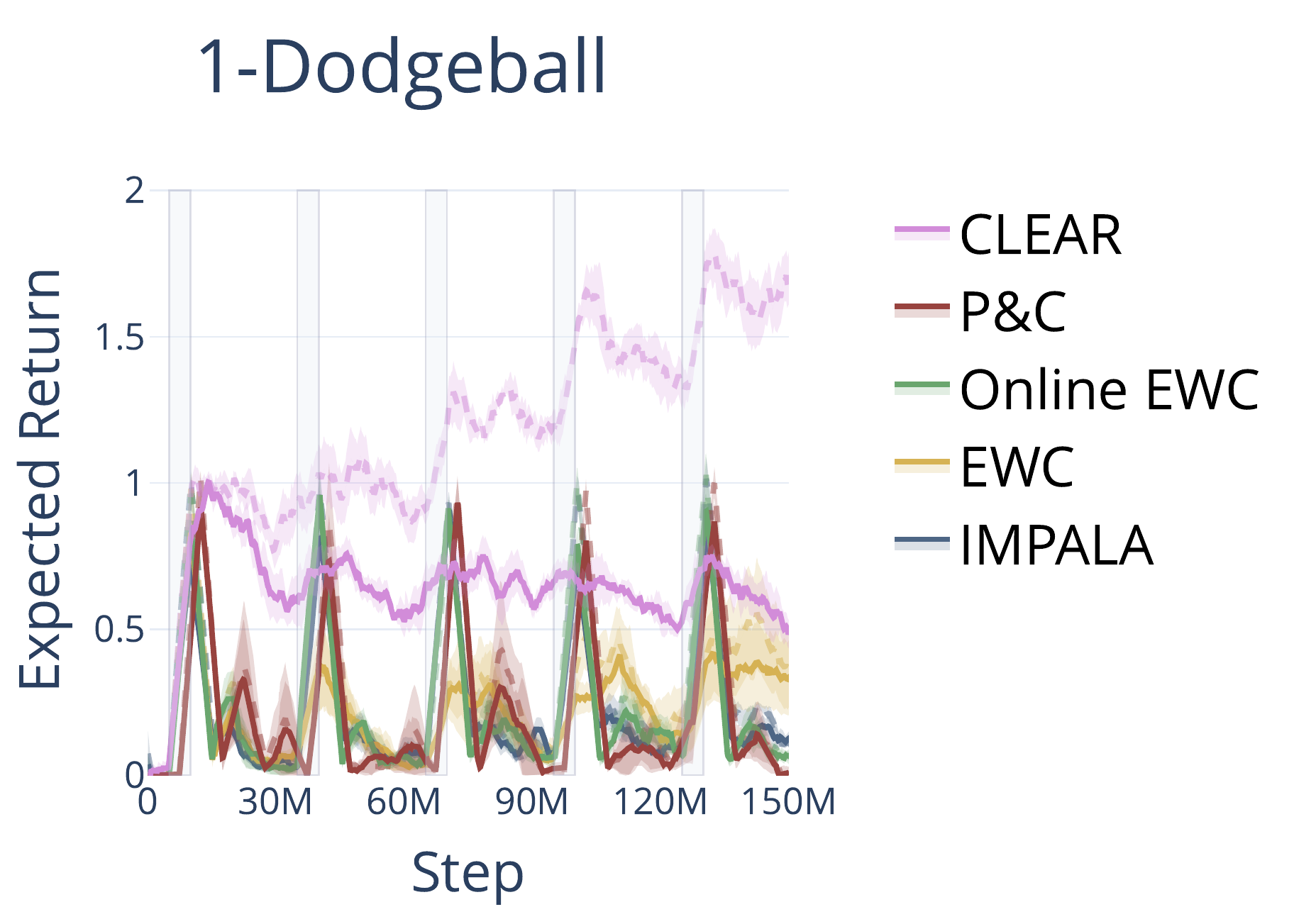}
    \includegraphics[trim=0 3em 0em 0, clip, width=0.35\textwidth]{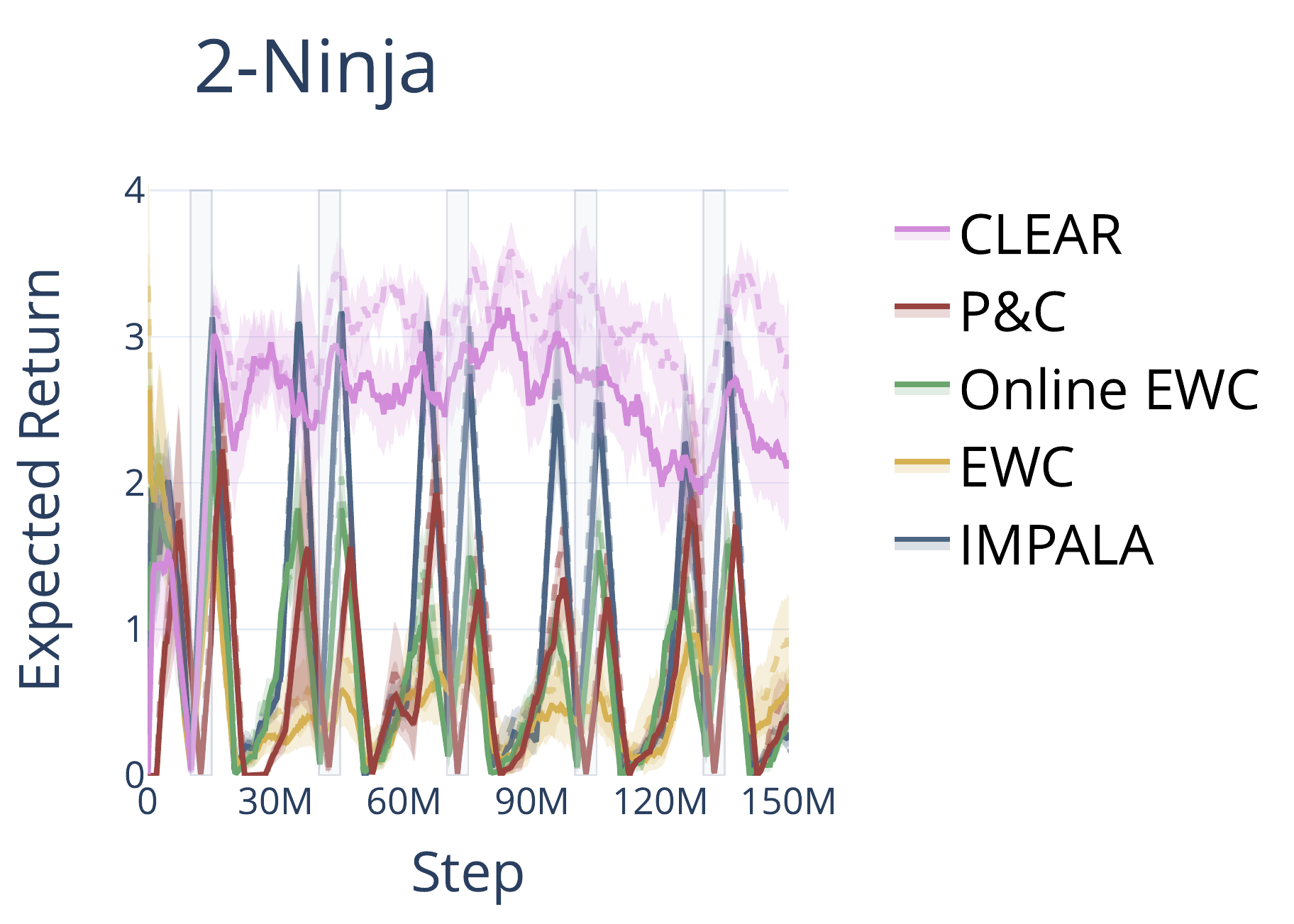} \\
    \includegraphics[trim=0 0em 18em 0, clip, width=0.23\textwidth]{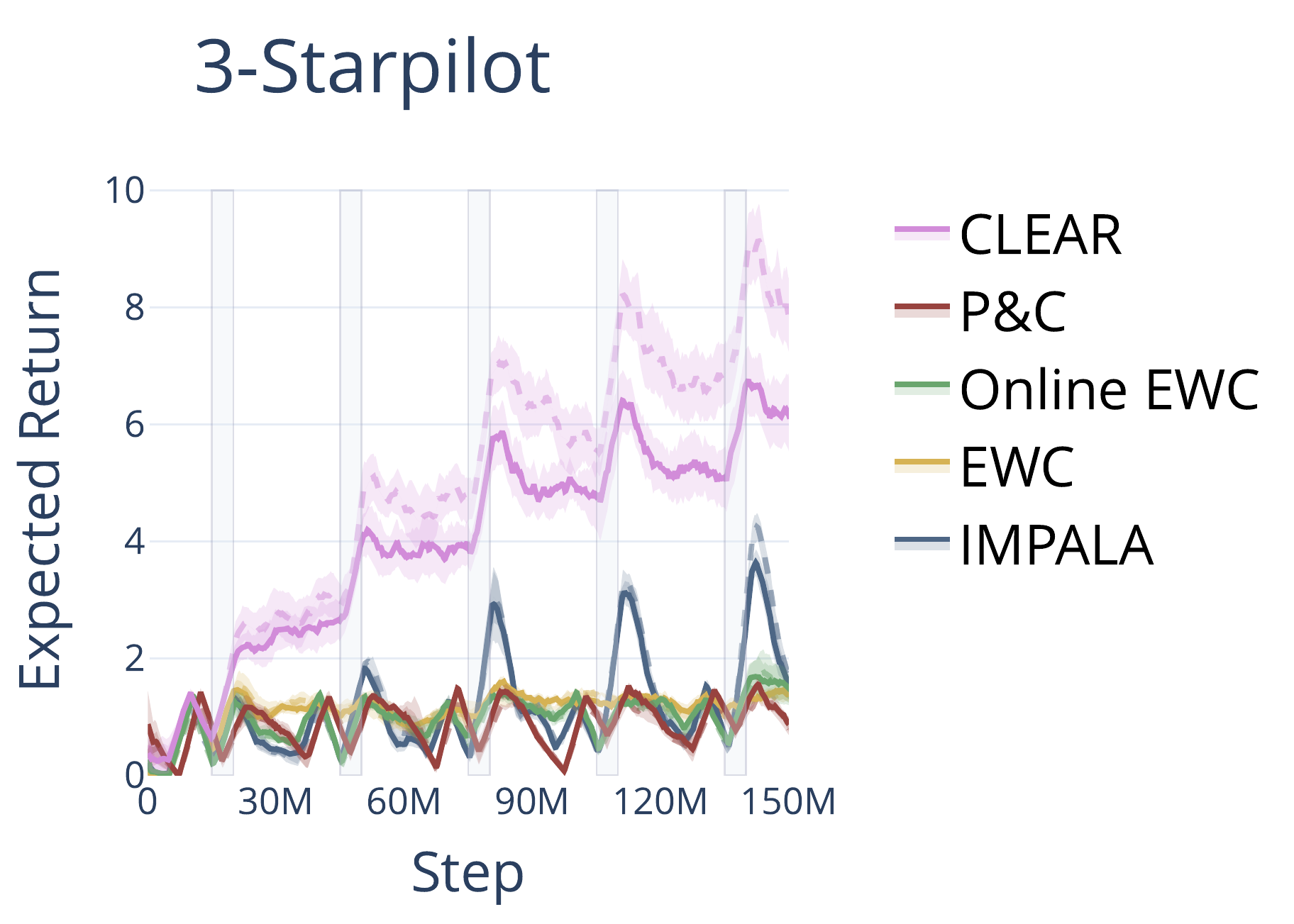}
    \includegraphics[trim=0 0em 18em 0, clip, width=0.23\textwidth]{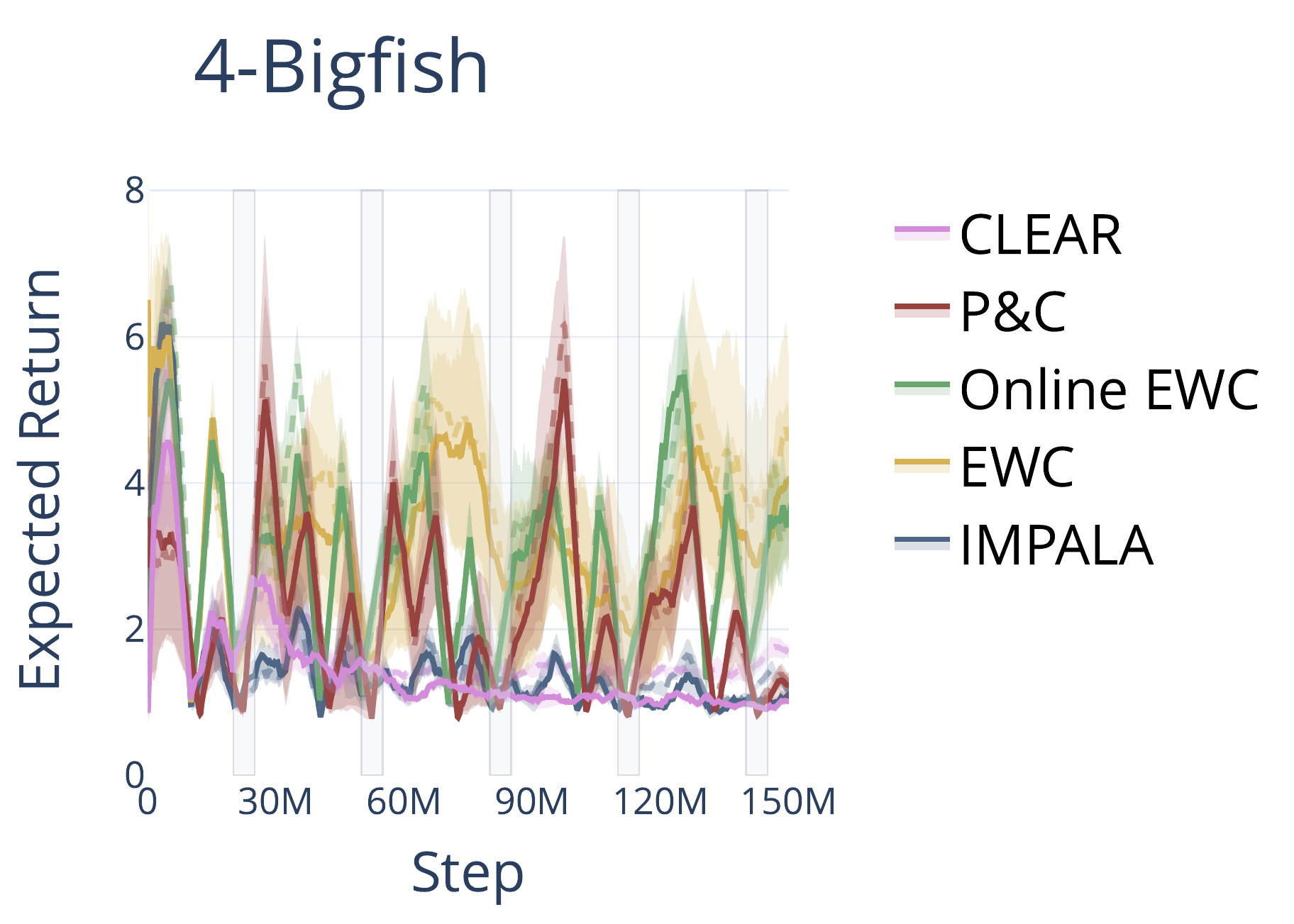}
    \includegraphics[trim=0 0em 18em 0, clip, width=0.23\textwidth]{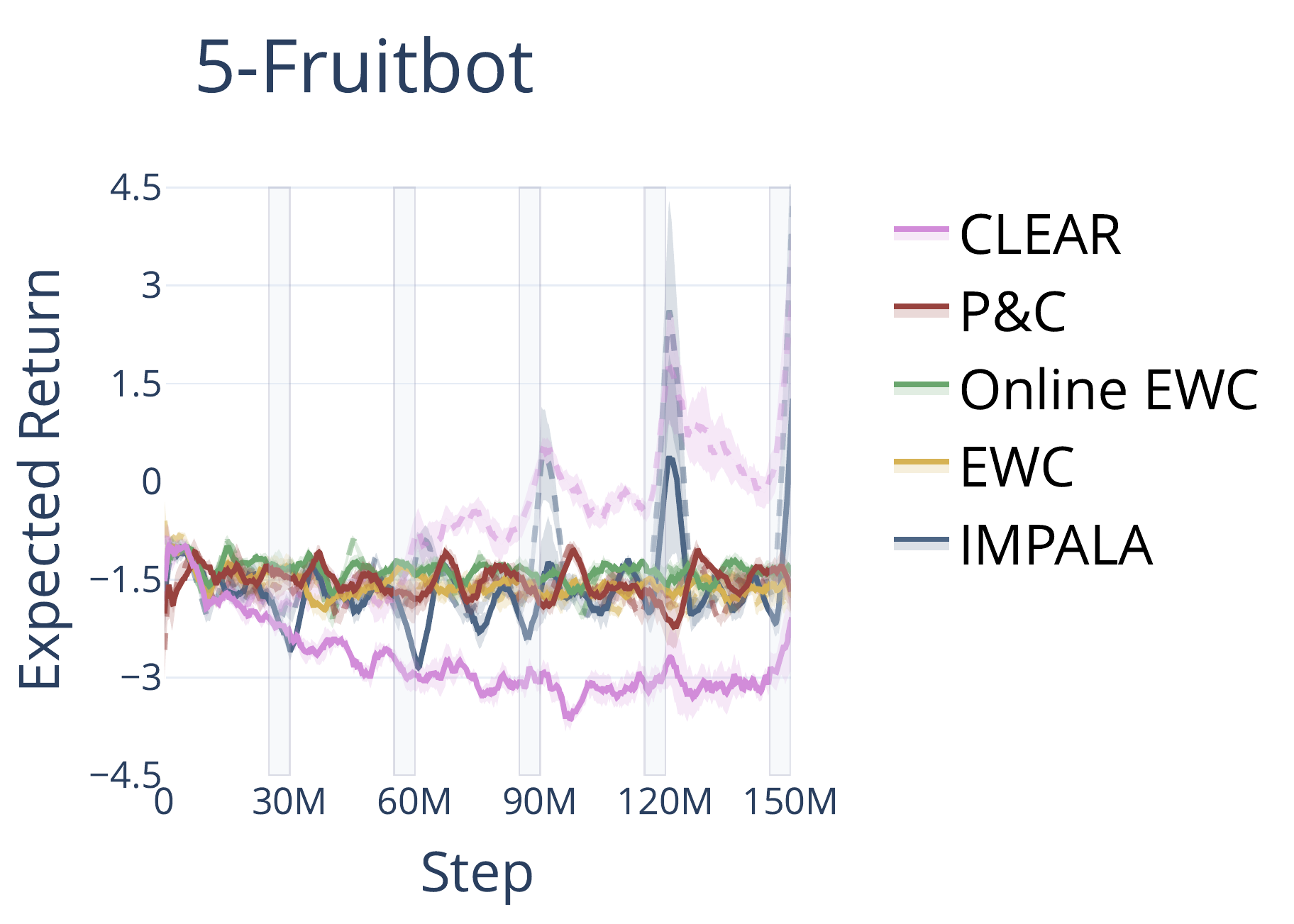}
    \hspace{3.6em}
    \caption{Results for Continual Evaluation $(\mathcal{C})$ on the 6 Procgen tasks, based on recommendations by~\citet{igl2021transient}. The solid line shows evaluation on unseen testing environments; the dashed line shows evaluation on training environments. Gray shaded rectangles show when the agent trains on each task. These results use the ``shallow'' model from the IMPALA paper~\cite{espeholt2018impala}, while the results in Section~\ref{section:procgen_results}, Figure~\ref{fig:procgen_results_res} use the ``deep'' residual model.
    }
    \label{fig:procgen_results}
\end{figure}

For readability, we provide an alternative plot (Figure~\ref{fig:procgen_results_noclear}) of the Procgen results shown in Section~\ref{section:procgen_results}, Figure~\ref{fig:procgen_results_res}, without the CLEAR baseline. Additionally, in Figure~\ref{fig:procgen_results}, we report baseline results on Procgen using the ``shallow'' model. The updated Procgen results in Section~\ref{section:procgen_results}, Figure~\ref{fig:procgen_results_res} use the ``deep'' residual model. 

\subsection{Final performance tables}
\label{sec:final_perf_tables}

\subsubsection{Atari, final performance tables}

\begin{table}[h]
    \tablemarginhack
    \begin{tabular}{lccccc}
         \toprule
         Task & CLEAR & P\&C & Online EWC & EWC & IMPALA  \\
         \midrule
         0-SpaceInvaders & $\bm{1766.75 \pm 89.36}$ & $209.14 \pm 55.91$ & $240.19 \pm 65.36$ & $654.45 \pm 134.16$ & $247.64 \pm 39.18$ \\
         1-Krull        & $\bm{6542.51 \pm 410.25}$ & $156.65 \pm 141.62$ & $1714.31 \pm 532.34$ & $2210.76 \pm 290.60$ & $1249.52 \pm 592.68$ \\
         2-BeamRider    & $\bm{2002.95 \pm 212.14}$ & $627.60 \pm 36.22$ & $491.86 \pm 68.31$ & $458.69 \pm 109.92$ & $425.58 \pm 108.48$ \\
         3-Hero         & $\bm{33604.29 \pm 1487.71}$ & $289.25 \pm 289.25$ & $0.00 \pm 0.00$ & $0.00 \pm 0.00$ & $0.00 \pm 0.00$ \\
         4-StarGunner   & $\bm{56365.93 \pm 2881.83}$ & $37515.49 \pm 3692.96$ & $3421.75 \pm 1453.06$ & $140.03 \pm 87.05$ & $3495.73 \pm 1774.20$ \\
         5-MsPacman     & $\bm{3536.24 \pm 400.45}$ & $996.19 \pm 58.19$ & $2171.94 \pm 229.90$ & $201.91 \pm 58.12$ & $2104.36 \pm 103.78$ \\
         \bottomrule
    \end{tabular}
    \caption{Comparison of final performance (mean$\pm$SEM) between methods in the environments for each Atari task.}
    \label{tab:final_perf_atari_train}
\end{table}

\subsubsection{Procgen, final performance tables}

\begin{table}[h]
    \centering
    \begin{tabular}{lccccc}
         \toprule
         Task & CLEAR & P\&C & Online EWC & EWC & IMPALA  \\
         \midrule
         0-Climber & $\bm{0.74 \pm 0.04}$ & $0.12 \pm 0.02$ & $0.37 \pm 0.06$ & $0.06 \pm 0.02$ & $0.22 \pm 0.04$ \\
         1-Dodgeball & $\bm{1.35 \pm 0.05}$ & $0.04 \pm 0.01$ & $0.05 \pm 0.01$ & $0.22 \pm 0.04$ & $0.06 \pm 0.01$ \\
         2-Ninja & $\bm{3.89 \pm 0.18}$ & $0.20 \pm 0.06$ & $0.11 \pm 0.04$ & $0.20 \pm 0.08$ & $0.38 \pm 0.06$ \\
         3-Starpilot & $\bm{38.29 \pm 1.45}$ & $0.52 \pm 0.09$ & $0.53 \pm 0.10$ & $1.95 \pm 0.06$ & $0.44 \pm 0.09$ \\
         4-Bigfish & $\bm{11.31 \pm 0.58}$ & $1.63 \pm 0.14$ & $1.81 \pm 0.18$ & $1.39 \pm 0.15$ & $2.65 \pm 0.18$ \\
         5-Fruitbot & $\bm{25.81 \pm 0.22}$ & $-0.46 \pm 0.42$ & $6.16 \pm 1.24$ & $-1.44 \pm 0.07$ & $22.62 \pm 0.33$ \\
         \bottomrule
    \end{tabular}
    \caption{Comparison of final performance (mean$\pm$SEM) between methods in the \textit{unseen testing} environments for each Procgen task.}
    \label{tab:final_perf_procgen_test}
\end{table}

\begin{table}[h]
    \centering
    \begin{tabular}{lccccc}
         \toprule
         Task & CLEAR & P\&C & Online EWC & EWC & IMPALA  \\
         \midrule
         0-Climber & $\bm{0.81 \pm 0.05}$ & $0.19 \pm 0.05$ & $0.29 \pm 0.05$ & $0.07 \pm 0.03$ & $0.24 \pm 0.04$ \\
         1-Dodgeball & $\bm{2.19 \pm 0.08}$ & $0.05 \pm 0.01$ & $0.05 \pm 0.01$ & $0.29 \pm 0.05$ & $0.05 \pm 0.01$ \\
         2-Ninja & $\bm{3.92 \pm 0.18}$ & $0.27 \pm 0.08$ & $0.12 \pm 0.03$ & $0.20 \pm 0.07$ & $0.42 \pm 0.09$ \\
         3-Starpilot & $\bm{38.76 \pm 1.52}$ & $0.56 \pm 0.10$ & $0.52 \pm 0.10$ & $1.99 \pm 0.06$ & $0.43 \pm 0.09$ \\
         4-Bigfish & $\bm{13.90 \pm 0.55}$ & $1.52 \pm 0.09$ & $1.86 \pm 0.18$ & $1.60 \pm 0.22$ & $2.88 \pm 0.27$ \\
         5-Fruitbot & $\bm{27.04 \pm 0.21}$ & $-0.49 \pm 0.38$ & $6.56 \pm 1.19$ & $-1.64 \pm 0.05$ & $22.58 \pm 0.41$ \\
         \bottomrule
    \end{tabular}
    \caption{Comparison of final performance (mean$\pm$SEM) between methods in the \textit{training} environments for each Procgen task.}
    \label{tab:final_perf_procgen_train}
\end{table}

\subsubsection{MiniHack, final performance tables}

\begin{table}[H]
    \centering
    \begin{tabular}{lcc}
         \toprule
         Task & CLEAR & IMPALA  \\
         \midrule
         0-Room-Random & $\bm{0.24 \pm 0.03}$ & $-0.02 \pm 0.08$ \\
         1-Room-Dark & $\bm{0.33 \pm 0.08}$ & $-0.11 \pm 0.07$ \\
         2-Room-Monster & $\bm{0.30 \pm 0.04}$ & $-0.02 \pm 0.08$ \\
         3-Room-Trap & $\bm{0.29 \pm 0.03}$ & $0.03 \pm 0.09$ \\
         4-Room-Ultimate & $\bm{0.37 \pm 0.07}$ & $-0.09 \pm 0.05$ \\
         5-Corridor-R2 & $-0.84 \pm 0.02$ & $\bm{-0.69 \pm 0.07}$ \\
         6-Corridor-R3 & $-0.85 \pm 0.03$ & $\bm{-0.70 \pm 0.07}$ \\
         7-KeyRoom & $\bm{-0.32 \pm 0.02}$ & $-0.40 \pm 0.00$ \\
         8-KeyRoom-Dark & $\bm{-0.38 \pm 0.00}$ & $-0.40 \pm 0.00$ \\
         9-River-Narrow & $\bm{-0.18 \pm 0.01}$ & $-0.19 \pm 0.04$ \\
         10-River-Monster & $-0.26 \pm 0.01$ & $\bm{-0.20 \pm 0.04}$ \\
         11-River-Lava & $-0.26 \pm 0.01$ & $\bm{-0.20 \pm 0.04}$ \\
         12-HideNSeek & $\bm{-0.10 \pm 0.03}$ & $-0.17 \pm 0.07$ \\
         13-HideNSeek-Lava & $\bm{-0.09 \pm 0.03}$ & $-0.17 \pm 0.07$ \\
         14-CorridorBattle & $\bm{-0.33 \pm 0.01}$ & $\bm{-0.33 \pm 0.01}$ \\
         \bottomrule
    \end{tabular}
    \caption{Comparison of final performance (mean$\pm$SEM) between methods in the \textit{unseen testing} environments for each MiniHack task.}
    \label{tab:final_perf_minihack_test}
\end{table}

\begin{table}[H]
    \centering
    \begin{tabular}{lcc}
         \toprule
         Task & CLEAR & IMPALA  \\
         \midrule
         0-Room-Random & $\bm{0.91 \pm 0.02}$ & $0.38 \pm 0.12$ \\
         1-Room-Dark & $\bm{0.67 \pm 0.04}$ & $0.23 \pm 0.08$ \\
         2-Room-Monster & $\bm{0.91 \pm 0.01}$ & $0.36 \pm 0.12$ \\
         3-Room-Trap & $\bm{0.92 \pm 0.02}$ & $0.38 \pm 0.11$ \\
         4-Room-Ultimate & $\bm{0.71 \pm 0.03}$ & $0.23 \pm 0.07$ \\
         5-Corridor-R2 & $\bm{-0.14 \pm 0.03}$ & $-0.68 \pm 0.06$ \\
         6-Corridor-R3 & $-0.86 \pm 0.02$ & $\bm{-0.68 \pm 0.07}$ \\
         7-KeyRoom & $\bm{-0.12 \pm 0.02}$ & $-0.20 \pm 0.00$ \\
         8-KeyRoom-Dark & $\bm{-0.18 \pm 0.01}$ & $-0.20 \pm 0.00$ \\
         9-River-Narrow & $\bm{0.05 \pm 0.04}$ & $-0.19 \pm 0.04$ \\
         10-River-Monster & $-0.20 \pm 0.02$ & $\bm{-0.17 \pm 0.05}$ \\
         11-River-Lava & $-0.22 \pm 0.01$ & $\bm{-0.18 \pm 0.04}$ \\
         12-HideNSeek & $\bm{0.17 \pm 0.02}$ & $0.03 \pm 0.06$ \\
         13-HideNSeek-Lava & $\bm{0.16 \pm 0.02}$ & $0.01 \pm 0.06$ \\
         14-CorridorBattle & $\bm{-0.18 \pm 0.04}$ & $-0.31 \pm 0.02$ \\
         \bottomrule
    \end{tabular}
    \caption{Comparison of final performance (mean$\pm$SEM) between methods in the \textit{training} environments for each MiniHack task.}
    \label{tab:final_perf_minihack_train}
\end{table}

\subsubsection{CHORES, final performance table}

\begin{table}[H]
    \tablemarginhack
    \begin{tabular}{lccc}
         \toprule
         Task & CLEAR & P\&C & EWC  \\
         \midrule
         \textbf{Mem-VaryRoom} & & & \\
         Room 402 & $-5.27 \pm 4.73$ & $0.00 \pm 0.00$ & $2.73 \pm 1.47$ \\
         Room 419 & $0.07 \pm 0.07$ & $0.00 \pm 0.00$ & $-6.53 \pm 3.47$ \\
         Room 423 & $-3.33 \pm 3.33$ & $0.00 \pm 0.00$ & $-3.37 \pm 3.17$ \\
         \\
         \textbf{Mem-VaryTask} & & & \\
         Hang TP & $-3.83 \pm 6.17$ & $-6.50 \pm 3.25$ & $-2.37 \pm 2.37$ \\
         Put TP on Counter & $-7.33 \pm 2.47$ & $-6.53 \pm 3.27$ & $-1.06 \pm 4.19$ \\
         Put TP in Cabinet & $12.71 \pm 9.15$ & $-6.26 \pm 3.15$ & $2.10 \pm 2.10$ \\
         \\
         \textbf{Mem-VaryObject} & & & \\
         Clean Fork & $-2.06 \pm 7.53$ & $-2.99 \pm 3.09$ & $0.27 \pm 0.18$ \\
         Clean Knife & $-0.86 \pm 4.18$ & $-3.13 \pm 3.03$ & $3.67 \pm 1.73$ \\
         Clean Spoon & $1.36 \pm 6.71$ & $-6.21 \pm 3.11$ & $-2.97 \pm 3.27$ \\
         \\
         \textbf{Gen-MultiTraj (train)} & & & \\
         Room 19, Cup & $-0.55 \pm 4.34$ & $-9.26 \pm 0.23$ & $-9.59 \pm 0.18$ \\
         Room 13, Potato & $-2.83 \pm 3.44$ & $-8.01 \pm 0.58$ & $-2.43 \pm 3.35$ \\
         Room 2, Lettuce & $-1.78 \pm 3.19$ & $-9.30 \pm 0.37$ & $-5.43 \pm 2.72$ \\
         \\
         \textbf{Gen-MultiTraj (test)} & & & \\
         Room 19, Cup & $-6.39 \pm 3.20$ & $-9.46 \pm 0.13$ & $-3.06 \pm 3.27$ \\
         Room 13, Potato & $-2.57 \pm 2.67$ & $-8.72 \pm 0.23$ & $-5.61 \pm 2.93$ \\
         Room 2, Lettuce & $-3.06 \pm 2.86$ & $-9.10 \pm 0.33$ & $-2.99 \pm 3.40$ \\
         \bottomrule
    \end{tabular}
    \caption{Comparison of final performance (mean$\pm$SEM) between methods in the environments for CHORES.}
    \label{tab:final_perf_chores}
\end{table}

\subsection{Future work}
\label{sec:future_work}

CORA is an extensive, on-going effort which we continue to develop and maintain. We are working on integrating in more baselines, such as a modular method and a continual supervised learning method adapted to the RL domain. We also hope to incorporate other useful metrics, such as for measuring data efficiency. While the set of benchmarks we present already pose considerable difficulty, we see several directions to build off CORA for more challenging evaluations. For instance, Procgen, MiniHack, and CHORES could be scaled up to even longer and more diverse task sequences. Agents could also be trained on tasks from a mix of different environments.

\clearpage
\section{Code Structure}
\label{sec:code_structure}

\subsection{Architecture diagram}
\begin{figure}
    \centering
    \includegraphics[width=0.75\textwidth]{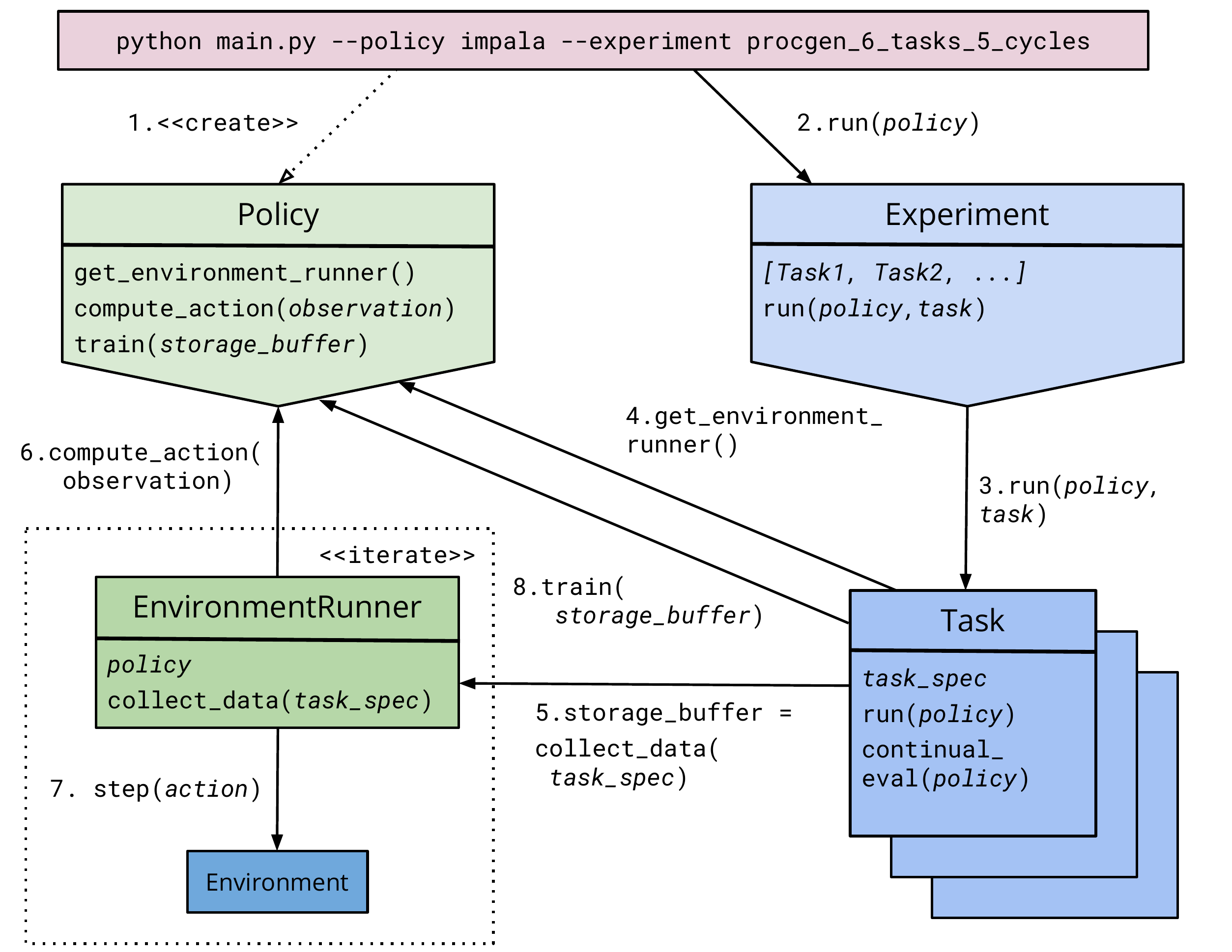}
    \caption{Sequence diagram representing the most basic flow of the \texttt{continual\_rl} package. Blue represents components defined by \texttt{Experiment}, and green represents components defined by \texttt{Policy}.}
    \label{fig:continual_rl_diagram}
\end{figure}

An overview of our code package architecture can be seen in Figure \ref{fig:continual_rl_diagram}. The two fundamental components of the package are \texttt{Experiment} and \texttt{Policy}. \texttt{Experiment} conceptually encapsulates everything that should remain the same between runs, such as task specification, ordering, duration, and observation dimensionality. \texttt{Policy} encapsulates everything an algorithm has control over and can change as tasks are learned. 

To train and evaluate agents on our benchmarks, these two things must be specified. They may be specified either via command line, or by configuration file, along with any hyperparameter changes from the defaults. We recommend referring to the README provided with the source code for more details on running experiments and implementing new policies and experiments.

\subsection{Policies}
Any \texttt{Policy} must implement: (i) computing an action given an observation and (ii) training in response to collected experience. Any existing code that does these can be integrated into the \texttt{continual\_rl} package by implementing a simple adapter wrapper. We provide an example of doing this for PPO based on the \texttt{pytorch-a2c-ppo-acktr} repository~\cite{kostrikov2018rl}. This enables easier integration of agents from outside our codebase, so the experiments and metrics provided by \texttt{continual\_rl} can be leveraged. 



The other thing a \texttt{Policy} must specify is how it should be run (i.e. its training loop), which we encapsulate in modules we refer to as EnvironmentRunners. In most simple cases, an existing EnvironmentRunner will suffice, such as EnvironmentRunnerBatch for standard, synchronous RL. However, in highly asynchronous or distributed cases, a user of CORA may wish to write their own. We provide more details on EnvironmentRunners in Section~\ref{sec:environment_runners}.

The final two steps to using a policy in CORA are the specification of configuration parameters, by extending \texttt{ConfigBase}, and adding the new policy to \texttt{continual\_rl/available\_policies.py}, so it can be used identically to existing ones, either via config file or via command line.



Additionally, since the policies are independent modules, it is also easy to use the provided policy implementations in a separate code base. The framework is installable as a pip package, which can be imported directly.

\subsubsection{EnvironmentRunners}
\label{sec:environment_runners}

EnvironmentRunners have one function they must implement: \texttt{collect\_data()}. Given the task specification, the EnvironmentRunner must collect any number of steps worth of data from the environment
and return the results of what it has collected. The function will be called repeatedly until the total number of steps for the task have been satisfied. Data collection for continual evaluation occurs between calls to \texttt{collect\_data()}, so care should be taken when selecting how much data to collect at a time. If too many timesteps are collected at once, the metrics will not be able to be computed as often as desired.

EnvironmentRunners can also be viewed as a higher-level API for more advanced policies. One example of how this is useful is for IMPALA~\cite{espeholt2018impala}. IMPALA's key feature is how it learns asychronously by decoupling collecting data with actors from training policies, so the simple \texttt{Policy} structure of \texttt{compute\_action()} and \texttt{train()} are insufficient. Instead, we define ImpalaEnvironmentRunner and implement a custom \texttt{collect\_data()} method that returns new results that have accumulated every fixed number of seconds to support the actors and learners working asynchronously.

\subsection{Experiments}
\label{sec:codestruct_experiments}

Any \texttt{Experiment} defines a sequence of tasks. Every task contains full specifications (available in \texttt{continual\_rl/task\_spec.py}) for what environment should be created, how many frames it is given as a budget, and so on. Each task also provides common preprocessing features for convenience. For instance, we can define an ImageTask that scales the observation image, stacks frames, and converts the observation to a PyTorch tensor. 

Experiments use this sequence of tasks to handle collecting metrics such as the Continual Evaluation metric described in Section~\ref{sec:metrics}. The Forgetting and Transfer metrics are computed in a post-processing step using the collected data from continual evaluation.

\clearpage
\section{Metrics}
\label{section:metrics_tables}

Recall that in all cases, each table represents how much training on the task in each column impacts the performance of the task in each row. These metrics are also only computed across the first cycle of the task sequence.

\subsection{Standard error of the mean}
\label{sec:compute_sem}

We present the standard error of the mean in our detailed metric tables. To aggregate metrics across rows and columns, we use the following procedure. First, we define a set of individual metric values $\mathbb{M}_{i,j} = \{ \mathcal{M}_{i,j,s} : s \in S \}$, where S represents the set of seeds used, $\mathcal{M}$ represents the metric to compute (either $\mathcal{F}$ or $\mathcal{Z}$), and $i$ and $j$ are task ids as defined in Section~\ref{sec:metrics}. For each non-aggregate entry in the table we compute $SEM_{i,j} = \frac{\sigma({\mathbb{M}_{i,j}}; ddof=1)}{\sqrt{|\mathbb{M}_{i,j}|}}$ as usual, where $\sigma(\mathbb{M})$ and $\abs{\mathbb{M}}$ are the standard deviation and size of set $\mathbb{M}$.

However, for the aggregate values (row, column, and table averages), it is not the case that $\mathbb{M}_{i,j}$ is independent from $\mathbb{M}_{i+1,j}$ or $\mathbb{M}_{i,j+1}$, so we cannot simply create an aggregate set across the rows or columns. 
Instead, we compute $\mathcal{M}_{j,s} = \frac{1}{|j-i|} \sum_i \mathcal{M}_{i,j,s}$. This metric averages a given seed over all $i$, which yields a metric that is independent and can be aggregated as described above. This allows us to define $SEM_{j} = \frac{\sigma(\mathbb{M}_{j}, ddof=1)}{\sqrt{N}}$, where $N$ is the number of tasks. $SEM_{i}$ is defined symmetrically. Finally, the full-table-aggregate $SEM$ is computed by averaging over the full set of $i\cdot j$ entries in $\mathcal{M}$.

\subsection{Atari Metrics: Forgetting}

\begin{table}[H]
\tablemarginhack
\setlength\tabcolsep{2pt} 
\tiny

\subfloat[IMPALA]{ 
\begin{tabular}{|l|Hlllll|l|}
\hline
 & 0-SpaceInvaders & 1-Krull & 2-BeamRider & 3-Hero & 4-StarGunner & 5-MsPacman & Avg ± SEM \\
\hline 
0-SpaceInvaders & \cellcolor{green!0} -- &
\cellcolor{red!15} 3.8 ± 0.3 & \cellcolor{green!0} -0.1 ± 0.2 & \cellcolor{green!1} -0.3 ± 0.3 & \cellcolor{red!4} 1.0 ± 0.4 & \cellcolor{green!1} -0.3 ± 0.3 & \cellcolor{red!3} 0.8 ± 0.1 \\
1-Krull & \cellcolor{green!0} -- & \cellcolor{green!0} -- & \cellcolor{red!22} 5.6 ± 0.7 & \cellcolor{red!5} 1.4 ± 0.5 & \cellcolor{green!5} -1.4 ± 0.5 & \cellcolor{red!4} 1.0 ± 0.5 & \cellcolor{red!6} 1.7 ± 0.2 \\
2-BeamRider & \cellcolor{green!0} -- & \cellcolor{green!0} -- & \cellcolor{green!0} -- & \cellcolor{red!25} 6.3 ± 0.2 & \cellcolor{red!8} 2.1 ± 0.3 & \cellcolor{red!0} -0.0 ± 0.1 & \cellcolor{red!11} 2.8 ± 0.1 \\
3-Hero & \cellcolor{green!0} -- & \cellcolor{green!0} -- & \cellcolor{green!0} -- & \cellcolor{green!0} -- & \cellcolor{red!34} 8.5 ± 0.2 & \cellcolor{red!0} 0.0 ± 0.0 & \cellcolor{red!17} 4.3 ± 0.1 \\
4-StarGunner & \cellcolor{green!0} -- & \cellcolor{green!0} -- & \cellcolor{green!0} -- & \cellcolor{green!0} -- & \cellcolor{green!0} -- & \cellcolor{red!26} 6.7 ± 0.3 & \cellcolor{red!26} 6.7 ± 0.3 \\
5-MsPacman & \cellcolor{green!0} -- & \cellcolor{green!0} -- & \cellcolor{green!0} -- & \cellcolor{green!0} -- & \cellcolor{green!0} -- & \cellcolor{green!0} -- & \cellcolor{green!0} -- \\
\hline
Avg ± SEM & \cellcolor{green!0} -- & \cellcolor{red!15} 3.8 ± 0.3 & \cellcolor{red!10} 2.7 ± 0.4 & \cellcolor{red!10} 2.5 ± 0.2 & \cellcolor{red!10} 2.6 ± 0.1 & \cellcolor{red!6} 1.5 ± 0.2 & \cellcolor{red!9} 2.3 ± 0.1 \\
\hline
\end{tabular}
}
\subfloat[EWC]{ 
 \begin{tabular}{|l|Hlllll|l|}
\hline
 & 0-SpaceInvaders & 1-Krull & 2-BeamRider & 3-Hero & 4-StarGunner & 5-MsPacman & Avg ± SEM \\
\hline
0-SpaceInvaders & \cellcolor{green!0} -- & \cellcolor{red!9} 2.3 ± 1.8 & \cellcolor{green!1} -0.3 ± 0.5 & \cellcolor{red!0} 0.0 ± 0.5 & \cellcolor{red!0} 0.1 ± 1.0 & \cellcolor{red!1} 0.4 ± 0.8 & \cellcolor{red!2} 0.5 ± 0.3 \\
1-Krull & \cellcolor{green!0} -- & \cellcolor{green!0} -- & \cellcolor{red!2} 0.6 ± 2.3 & \cellcolor{green!2} -0.5 ± 0.4 & \cellcolor{red!7} 1.9 ± 1.5 & \cellcolor{green!0} -0.1 ± 0.4 & \cellcolor{red!2} 0.5 ± 0.6 \\
2-BeamRider & \cellcolor{green!0} -- & \cellcolor{green!0} -- & \cellcolor{green!0} -- & \cellcolor{green!1} -0.3 ± 0.1 & \cellcolor{red!4} 1.0 ± 0.7 & \cellcolor{green!2} -0.5 ± 0.2 & \cellcolor{red!0} 0.1 ± 0.1 \\
3-Hero & \cellcolor{green!0} -- & \cellcolor{green!0} -- & \cellcolor{green!0} -- & \cellcolor{green!0} -- & \cellcolor{red!0} 0.0 ± 0.0 & \cellcolor{red!0} 0.0 ± 0.0 & \cellcolor{red!0} 0.0 ± 0.0 \\
4-StarGunner & \cellcolor{green!0} -- & \cellcolor{green!0} -- & \cellcolor{green!0} -- & \cellcolor{green!0} -- & \cellcolor{green!0} -- & \cellcolor{green!1} -0.3 ± 0.3 & \cellcolor{green!1} -0.3 ± 0.3 \\
5-MsPacman & \cellcolor{green!0} -- & \cellcolor{green!0} -- & \cellcolor{green!0} -- & \cellcolor{green!0} -- & \cellcolor{green!0} -- & \cellcolor{green!0} -- & \cellcolor{green!0} -- \\
\hline
Avg ± SEM & \cellcolor{green!0} -- & \cellcolor{red!9} 2.3 ± 1.8 & \cellcolor{red!0} 0.2 ± 1.1 & \cellcolor{green!1} -0.3 ± 0.1 & \cellcolor{red!3} 0.8 ± 0.6 & \cellcolor{green!0} -0.1 ± 0.2 & \cellcolor{red!1} 0.3 ± 0.3 \\
\hline
\end{tabular}
}

\subfloat[Online EWC]{ 
 \begin{tabular}{|l|Hlllll|l|}
\hline
 & 0-SpaceInvaders & 1-Krull & 2-BeamRider & 3-Hero & 4-StarGunner & 5-MsPacman & Avg ± SEM \\
\hline
0-SpaceInvaders & \cellcolor{green!0} -- & \cellcolor{red!23} 5.9 ± 0.6 & \cellcolor{green!2} -0.7 ± 0.6 & \cellcolor{red!1} 0.3 ± 0.4 & \cellcolor{red!2} 0.6 ± 0.5 & \cellcolor{green!0} -0.2 ± 0.4 & \cellcolor{red!4} 1.2 ± 0.1 \\
1-Krull & \cellcolor{green!0} -- & \cellcolor{green!0} -- & \cellcolor{red!22} 5.7 ± 0.7 & \cellcolor{red!2} 0.5 ± 0.6 & \cellcolor{red!6} 1.6 ± 1.0 & \cellcolor{green!2} -0.5 ± 0.4 & \cellcolor{red!7} 1.8 ± 0.2 \\
2-BeamRider & \cellcolor{green!0} -- & \cellcolor{green!0} -- & \cellcolor{green!0} -- & \cellcolor{red!10} 2.6 ± 0.4 & \cellcolor{red!2} 0.5 ± 0.2 & \cellcolor{red!2} 0.6 ± 0.2 & \cellcolor{red!4} 1.2 ± 0.1 \\
3-Hero & \cellcolor{green!0} -- & \cellcolor{green!0} -- & \cellcolor{green!0} -- & \cellcolor{green!0} -- & \cellcolor{red!24} 6.0 ± 1.6 & \cellcolor{red!6} 1.7 ± 1.5 & \cellcolor{red!15} 3.9 ± 0.1 \\
4-StarGunner & \cellcolor{green!0} -- & \cellcolor{green!0} -- & \cellcolor{green!0} -- & \cellcolor{green!0} -- & \cellcolor{green!0} -- & \cellcolor{red!0} 0.1 ± 0.1 & \cellcolor{red!0} 0.1 ± 0.1 \\
5-MsPacman & \cellcolor{green!0} -- & \cellcolor{green!0} -- & \cellcolor{green!0} -- & \cellcolor{green!0} -- & \cellcolor{green!0} -- & \cellcolor{green!0} -- & \cellcolor{green!0} -- \\
\hline
Avg ± SEM & \cellcolor{green!0} -- & \cellcolor{red!23} 5.9 ± 0.6 & \cellcolor{red!10} 2.5 ± 0.4 & \cellcolor{red!4} 1.1 ± 0.3 & \cellcolor{red!8} 2.2 ± 0.7 & \cellcolor{red!1} 0.3 ± 0.3 & \cellcolor{red!6} 1.6 ± 0.1 \\
\hline
\end{tabular}
}
\subfloat[P\&C]{ 
 \begin{tabular}{|l|Hlllll|l|}
\hline
 & 0-SpaceInvaders & 1-Krull & 2-BeamRider & 3-Hero & 4-StarGunner & 5-MsPacman & Avg ± SEM \\
\hline
0-SpaceInvaders & \cellcolor{green!0} -- & \cellcolor{red!18} 4.6 ± 0.7 & \cellcolor{green!2} -0.7 ± 0.6 & \cellcolor{green!2} -0.6 ± 0.2 & \cellcolor{red!6} 1.6 ± 0.5 & \cellcolor{green!0} -0.1 ± 0.3 & \cellcolor{red!4} 1.0 ± 0.2 \\
1-Krull & \cellcolor{green!0} -- & \cellcolor{green!0} -- & \cellcolor{red!19} 4.9 ± 0.4 & \cellcolor{red!0} 0.1 ± 0.5 & \cellcolor{red!1} 0.3 ± 0.9 & \cellcolor{red!0} 0.1 ± 0.5 & \cellcolor{red!5} 1.3 ± 0.1 \\
2-BeamRider & \cellcolor{green!0} -- & \cellcolor{green!0} -- & \cellcolor{green!0} -- & \cellcolor{red!9} 2.4 ± 0.9 & \cellcolor{red!18} 4.5 ± 0.9 & \cellcolor{green!0} -0.1 ± 0.1 & \cellcolor{red!9} 2.3 ± 0.2 \\
3-Hero & \cellcolor{green!0} -- & \cellcolor{green!0} -- & \cellcolor{green!0} -- & \cellcolor{green!0} -- & \cellcolor{red!20} 5.2 ± 0.6 & \cellcolor{red!0} 0.0 ± 0.0 & \cellcolor{red!10} 2.6 ± 0.3 \\
4-StarGunner & \cellcolor{green!0} -- & \cellcolor{green!0} -- & \cellcolor{green!0} -- & \cellcolor{green!0} -- & \cellcolor{green!0} -- & \cellcolor{red!18} 4.7 ± 0.9 & \cellcolor{red!18} 4.7 ± 0.9 \\
5-MsPacman & \cellcolor{green!0} -- & \cellcolor{green!0} -- & \cellcolor{green!0} -- & \cellcolor{green!0} -- & \cellcolor{green!0} -- & \cellcolor{green!0} -- & \cellcolor{green!0} -- \\
\hline
Avg ± SEM & \cellcolor{green!0} -- & \cellcolor{red!18} 4.6 ± 0.7 & \cellcolor{red!8} 2.1 ± 0.3 & \cellcolor{red!2} 0.6 ± 0.3 & \cellcolor{red!11} 2.9 ± 0.5 & \cellcolor{red!3} 0.9 ± 0.2 & \cellcolor{red!7} 1.8 ± 0.1 \\
\hline
\end{tabular}
}

\subfloat[CLEAR]{ 
 \begin{tabular}{|l|Hlllll|l|}
\hline
 & 0-SpaceInvaders & 1-Krull & 2-BeamRider & 3-Hero & 4-StarGunner & 5-MsPacman & Avg ± SEM \\
\hline
0-SpaceInvaders & \cellcolor{green!0} -- & \cellcolor{red!14} 3.6 ± 0.3 & \cellcolor{green!2} -0.7 ± 0.1 & \cellcolor{red!2} 0.5 ± 0.2 & \cellcolor{red!0} 0.1 ± 0.1 & \cellcolor{red!0} 0.2 ± 0.1 & \cellcolor{red!2} 0.7 ± 0.1 \\
1-Krull & \cellcolor{green!0} -- & \cellcolor{green!0} -- & \cellcolor{red!7} 1.9 ± 0.5 & \cellcolor{green!0} -0.1 ± 0.2 & \cellcolor{green!0} -0.2 ± 0.2 & \cellcolor{red!1} 0.3 ± 0.1 & \cellcolor{red!2} 0.5 ± 0.1 \\
2-BeamRider & \cellcolor{green!0} -- & \cellcolor{green!0} -- & \cellcolor{green!0} -- & \cellcolor{red!6} 1.6 ± 0.3 & \cellcolor{green!3} -0.8 ± 0.1 & \cellcolor{red!5} 1.3 ± 0.3 & \cellcolor{red!2} 0.7 ± 0.2 \\
3-Hero & \cellcolor{green!0} -- & \cellcolor{green!0} -- & \cellcolor{green!0} -- & \cellcolor{green!0} -- & \cellcolor{red!0} 0.2 ± 0.3 & \cellcolor{red!6} 1.7 ± 1.0 & \cellcolor{red!4} 1.0 ± 0.4 \\
4-StarGunner & \cellcolor{green!0} -- & \cellcolor{green!0} -- & \cellcolor{green!0} -- & \cellcolor{green!0} -- & \cellcolor{green!0} -- & \cellcolor{red!5} 1.4 ± 0.3 & \cellcolor{red!5} 1.4 ± 0.3 \\
5-MsPacman & \cellcolor{green!0} -- & \cellcolor{green!0} -- & \cellcolor{green!0} -- & \cellcolor{green!0} -- & \cellcolor{green!0} -- & \cellcolor{green!0} -- & \cellcolor{green!0} -- \\
\hline
Avg ± SEM & \cellcolor{green!0} -- & \cellcolor{red!14} 3.6 ± 0.3 & \cellcolor{red!2} 0.6 ± 0.3 & \cellcolor{red!2} 0.7 ± 0.2 & \cellcolor{green!0} -0.2 ± 0.1 & \cellcolor{red!4} 1.0 ± 0.3 & \cellcolor{red!2} 0.7 ± 0.1 \\
\hline
\end{tabular}
}

\vspace{2em}
\caption{Atari Forgetting metrics with the standard error. We can see that forgetting across the board is quite high, though mitigated by both EWC and CLEAR.}
\label{tab:atari_forgetting_sem}
\end{table}

\clearpage

\subsection{Atari Metrics: Transfer}
\begin{table}[H]
\tablemarginhack
\setlength\tabcolsep{2pt} 
\tiny

\subfloat[IMPALA]{ 
 \begin{tabular}{|l|lllllH|l|}
\hline
 & 0-SpaceInvaders & 1-Krull & 2-BeamRider & 3-Hero & 4-StarGunner & 5-MsPacman & Avg ± SEM \\
\hline
0-SpaceInvaders & \cellcolor{green!0} -- & \cellcolor{green!0} -- & \cellcolor{green!0} -- & \cellcolor{green!0} -- & \cellcolor{green!0} -- & \cellcolor{green!0} -- & \cellcolor{green!0} -- \\
1-Krull & \cellcolor{green!0} 0.1 ± 0.0 & \cellcolor{green!0} -- & \cellcolor{green!0} -- & \cellcolor{green!0} -- & \cellcolor{green!0} -- & \cellcolor{green!0} -- & \cellcolor{green!0} 0.1 ± 0.0 \\
2-BeamRider & \cellcolor{green!0} 0.2 ± 0.2 & \cellcolor{red!0} 0.0 ± 0.0 & \cellcolor{green!0} -- & \cellcolor{green!0} -- & \cellcolor{green!0} -- & \cellcolor{green!0} -- & \cellcolor{green!0} 0.1 ± 0.1 \\
3-Hero & \cellcolor{red!0} 0.0 ± 0.0 & \cellcolor{red!0} -0.0 ± 0.0 & \cellcolor{green!0} 0.2 ± 0.2 & \cellcolor{green!0} -- & \cellcolor{green!0} -- & \cellcolor{green!0} -- & \cellcolor{green!0} 0.1 ± 0.1 \\
4-StarGunner & \cellcolor{red!0} 0.0 ± 0.0 & \cellcolor{red!0} -0.0 ± 0.0 & \cellcolor{red!0} 0.0 ± 0.0 & \cellcolor{red!0} 0.0 ± 0.0 & \cellcolor{green!0} -- & \cellcolor{green!0} -- & \cellcolor{red!0} 0.0 ± 0.0 \\
5-MsPacman & \cellcolor{green!2} 0.6 ± 0.2 & \cellcolor{red!1} -0.4 ± 0.3 & \cellcolor{green!2} 0.7 ± 0.3 & \cellcolor{red!3} -0.8 ± 0.4 & \cellcolor{green!0} 0.2 ± 0.5 & \cellcolor{green!0} -- & \cellcolor{green!0} 0.1 ± 0.1 \\
\hline
Avg ± SEM & \cellcolor{green!0} 0.2 ± 0.1 & \cellcolor{red!0} -0.1 ± 0.1 & \cellcolor{green!1} 0.3 ± 0.1 & \cellcolor{red!1} -0.4 ± 0.2 & \cellcolor{green!0} 0.2 ± 0.5 & \cellcolor{green!0} -- & \cellcolor{green!0} 0.1 ± 0.0 \\
\hline
\end{tabular}
}
\subfloat[EWC]{ 
 \begin{tabular}{|l|lllllH|l|}
\hline
 & 0-SpaceInvaders & 1-Krull & 2-BeamRider & 3-Hero & 4-StarGunner & 5-MsPacman & Avg ± SEM \\
\hline
0-SpaceInvaders & \cellcolor{green!0} -- & \cellcolor{green!0} -- & \cellcolor{green!0} -- & \cellcolor{green!0} -- & \cellcolor{green!0} -- & \cellcolor{green!0} -- & \cellcolor{green!0} -- \\
1-Krull & \cellcolor{green!7} 1.9 ± 1.2 & \cellcolor{green!0} -- & \cellcolor{green!0} -- & \cellcolor{green!0} -- & \cellcolor{green!0} -- & \cellcolor{green!0} -- & \cellcolor{green!7} 1.9 ± 1.2 \\
2-BeamRider & \cellcolor{green!0} 0.2 ± 2.0 & \cellcolor{red!10} -2.7 ± 1.3 & \cellcolor{green!0} -- & \cellcolor{green!0} -- & \cellcolor{green!0} -- & \cellcolor{green!0} -- & \cellcolor{red!5} -1.3 ± 1.4 \\
3-Hero & \cellcolor{red!0} 0.0 ± 0.0 & \cellcolor{red!0} -0.0 ± 0.0 & \cellcolor{red!0} 0.0 ± 0.0 & \cellcolor{green!0} -- & \cellcolor{green!0} -- & \cellcolor{green!0} -- & \cellcolor{red!0} 0.0 ± 0.0 \\
4-StarGunner & \cellcolor{green!13} 3.3 ± 1.2 & \cellcolor{red!6} -1.7 ± 2.0 & \cellcolor{red!0} -0.1 ± 1.1 & \cellcolor{green!1} 0.3 ± 0.6 & \cellcolor{green!0} -- & \cellcolor{green!0} -- & \cellcolor{green!1} 0.4 ± 0.5 \\
5-MsPacman & \cellcolor{green!6} 1.6 ± 0.5 & \cellcolor{red!5} -1.4 ± 0.4 & \cellcolor{green!3} 0.8 ± 0.7 & \cellcolor{red!0} -0.2 ± 0.2 & \cellcolor{green!0} 0.1 ± 0.5 & \cellcolor{green!0} -- & \cellcolor{green!0} 0.2 ± 0.1 \\
\hline
Avg ± SEM & \cellcolor{green!5} 1.4 ± 0.4 & \cellcolor{red!6} -1.5 ± 0.4 & \cellcolor{green!0} 0.2 ± 0.6 & \cellcolor{green!0} 0.1 ± 0.3 & \cellcolor{green!0} 0.1 ± 0.5 & \cellcolor{green!0} -- & \cellcolor{green!0} 0.1 ± 0.2 \\
\hline
\end{tabular}
}

\subfloat[Online EWC]{ 
 \begin{tabular}{|l|lllllH|l|}
\hline
 & 0-SpaceInvaders & 1-Krull & 2-BeamRider & 3-Hero & 4-StarGunner & 5-MsPacman & Avg ± SEM \\
\hline
0-SpaceInvaders & \cellcolor{green!0} -- & \cellcolor{green!0} -- & \cellcolor{green!0} -- & \cellcolor{green!0} -- & \cellcolor{green!0} -- & \cellcolor{green!0} -- & \cellcolor{green!0} -- \\
1-Krull & \cellcolor{red!0} -0.2 ± 0.2 & \cellcolor{green!0} -- & \cellcolor{green!0} -- & \cellcolor{green!0} -- & \cellcolor{green!0} -- & \cellcolor{green!0} -- & \cellcolor{red!0} -0.2 ± 0.2 \\
2-BeamRider & \cellcolor{green!0} 0.1 ± 0.5 & \cellcolor{red!2} -0.7 ± 0.3 & \cellcolor{green!0} -- & \cellcolor{green!0} -- & \cellcolor{green!0} -- & \cellcolor{green!0} -- & \cellcolor{red!1} -0.3 ± 0.2 \\
3-Hero & \cellcolor{green!1} 0.3 ± 0.2 & \cellcolor{red!1} -0.3 ± 0.2 & \cellcolor{red!0} 0.0 ± 0.0 & \cellcolor{green!0} -- & \cellcolor{green!0} -- & \cellcolor{green!0} -- & \cellcolor{red!0} 0.0 ± 0.0 \\
4-StarGunner & \cellcolor{red!0} 0.0 ± 0.0 & \cellcolor{red!0} -0.0 ± 0.0 & \cellcolor{red!0} 0.0 ± 0.0 & \cellcolor{red!0} -0.0 ± 0.0 & \cellcolor{green!0} -- & \cellcolor{green!0} -- & \cellcolor{red!0} 0.0 ± 0.0 \\
5-MsPacman & \cellcolor{green!3} 0.8 ± 0.3 & \cellcolor{red!3} -0.9 ± 0.4 & \cellcolor{green!0} 0.2 ± 0.1 & \cellcolor{green!1} 0.4 ± 0.3 & \cellcolor{red!1} -0.4 ± 0.3 & \cellcolor{green!0} -- & \cellcolor{red!0} 0.0 ± 0.1 \\
\hline
Avg ± SEM & \cellcolor{green!0} 0.2 ± 0.1 & \cellcolor{red!2} -0.5 ± 0.1 & \cellcolor{green!0} 0.1 ± 0.0 & \cellcolor{green!0} 0.2 ± 0.2 & \cellcolor{red!1} -0.4 ± 0.3 & \cellcolor{green!0} -- & \cellcolor{red!0} -0.0 ± 0.0 \\
\hline
\end{tabular}
}
\subfloat[P\&C]{ 
 \begin{tabular}{|l|lllllH|l|}
\hline
 & 0-SpaceInvaders & 1-Krull & 2-BeamRider & 3-Hero & 4-StarGunner & 5-MsPacman & Avg ± SEM \\
\hline
0-SpaceInvaders & \cellcolor{green!0} -- & \cellcolor{green!0} -- & \cellcolor{green!0} -- & \cellcolor{green!0} -- & \cellcolor{green!0} -- & \cellcolor{green!0} -- & \cellcolor{green!0} -- \\
1-Krull & \cellcolor{green!1} 0.4 ± 0.3 & \cellcolor{green!0} -- & \cellcolor{green!0} -- & \cellcolor{green!0} -- & \cellcolor{green!0} -- & \cellcolor{green!0} -- & \cellcolor{green!1} 0.4 ± 0.3 \\
2-BeamRider & \cellcolor{green!1} 0.3 ± 0.3 & \cellcolor{red!0} 0.0 ± 0.1 & \cellcolor{green!0} -- & \cellcolor{green!0} -- & \cellcolor{green!0} -- & \cellcolor{green!0} -- & \cellcolor{green!0} 0.2 ± 0.1 \\
3-Hero & \cellcolor{red!0} 0.0 ± 0.0 & \cellcolor{red!0} 0.0 ± 0.0 & \cellcolor{red!0} 0.0 ± 0.0 & \cellcolor{green!0} -- & \cellcolor{green!0} -- & \cellcolor{green!0} -- & \cellcolor{red!0} 0.0 ± 0.0 \\
4-StarGunner & \cellcolor{red!0} -0.0 ± 0.0 & \cellcolor{red!0} 0.0 ± 0.0 & \cellcolor{red!0} -0.0 ± 0.0 & \cellcolor{red!0} 0.0 ± 0.0 & \cellcolor{green!0} -- & \cellcolor{green!0} -- & \cellcolor{red!0} 0.0 ± 0.0 \\
5-MsPacman & \cellcolor{red!2} -0.7 ± 0.5 & \cellcolor{red!1} -0.4 ± 0.2 & \cellcolor{green!6} 1.6 ± 0.4 & \cellcolor{red!4} -1.0 ± 0.4 & \cellcolor{red!2} -0.6 ± 0.3 & \cellcolor{green!0} -- & \cellcolor{red!0} -0.2 ± 0.1 \\
\hline
Avg ± SEM & \cellcolor{red!0} -0.0 ± 0.2 & \cellcolor{red!0} -0.1 ± 0.0 & \cellcolor{green!2} 0.5 ± 0.1 & \cellcolor{red!2} -0.5 ± 0.2 & \cellcolor{red!2} -0.6 ± 0.3 & \cellcolor{green!0} -- & \cellcolor{red!0} -0.0 ± 0.1 \\
\hline
\end{tabular}
}

\subfloat[CLEAR]{ 
 \begin{tabular}{|l|lllllH|l|}
\hline
 & 0-SpaceInvaders & 1-Krull & 2-BeamRider & 3-Hero & 4-StarGunner & 5-MsPacman & Avg ± SEM \\
\hline
0-SpaceInvaders & \cellcolor{green!0} -- & \cellcolor{green!0} -- & \cellcolor{green!0} -- & \cellcolor{green!0} -- & \cellcolor{green!0} -- & \cellcolor{green!0} -- & \cellcolor{green!0} -- \\
1-Krull & \cellcolor{green!1} 0.3 ± 0.3 & \cellcolor{green!0} -- & \cellcolor{green!0} -- & \cellcolor{green!0} -- & \cellcolor{green!0} -- & \cellcolor{green!0} -- & \cellcolor{green!1} 0.3 ± 0.3 \\
2-BeamRider & \cellcolor{red!1} -0.3 ± 0.3 & \cellcolor{green!0} 0.2 ± 0.2 & \cellcolor{green!0} -- & \cellcolor{green!0} -- & \cellcolor{green!0} -- & \cellcolor{green!0} -- & \cellcolor{red!0} -0.1 ± 0.2 \\
3-Hero & \cellcolor{red!0} 0.0 ± 0.0 & \cellcolor{red!0} 0.0 ± 0.0 & \cellcolor{green!0} 0.1 ± 0.0 & \cellcolor{green!0} -- & \cellcolor{green!0} -- & \cellcolor{green!0} -- & \cellcolor{red!0} 0.0 ± 0.0 \\
4-StarGunner & \cellcolor{red!0} 0.0 ± 0.0 & \cellcolor{red!0} -0.0 ± 0.0 & \cellcolor{red!0} 0.0 ± 0.0 & \cellcolor{red!0} 0.0 ± 0.0 & \cellcolor{green!0} -- & \cellcolor{green!0} -- & \cellcolor{red!0} 0.0 ± 0.0 \\
5-MsPacman & \cellcolor{green!2} 0.5 ± 0.2 & \cellcolor{red!1} -0.3 ± 0.2 & \cellcolor{green!1} 0.3 ± 0.2 & \cellcolor{red!0} -0.1 ± 0.3 & \cellcolor{red!0} -0.1 ± 0.1 & \cellcolor{green!0} -- & \cellcolor{red!0} 0.0 ± 0.0 \\
\hline
Avg ± SEM & \cellcolor{green!0} 0.1 ± 0.1 & \cellcolor{red!0} -0.0 ± 0.0 & \cellcolor{green!0} 0.1 ± 0.1 & \cellcolor{red!0} -0.1 ± 0.1 & \cellcolor{red!0} -0.1 ± 0.1 & \cellcolor{green!0} -- & \cellcolor{red!0} 0.0 ± 0.0 \\
\hline
\end{tabular}
}

\vspace{2em}
\caption{Atari Transfer metrics with standard error. Note that overall there is little forward transfer observed; this is expected because the tasks bear little similarity.}
\label{tab:atari_transfer}
\end{table}

\clearpage

\subsection{Procgen Metrics: Forgetting}
\label{sec:procgen_forgetting}

\begin{table}[H]
\tablemarginhack
\setlength\tabcolsep{2pt} 
\tiny

\subfloat[IMPALA]{ 
 \begin{tabular}{|l|Hlllll|l|}
\hline
 & 0-Climber & 1-Dodgeball & 2-Ninja & 3-Starpilot & 4-Bigfish & 5-Fruitbot & Avg ± SEM \\
\hline
0-Climber & \cellcolor{green!0} -- & \cellcolor{red!10} 2.6 ± 0.2 & \cellcolor{green!4} -1.2 ± 0.2 & \cellcolor{red!5} 1.4 ± 0.2 & \cellcolor{green!2} -0.7 ± 0.1 & \cellcolor{red!1} 0.3 ± 0.1 & \cellcolor{red!2} 0.5 ± 0.0 \\
1-Dodgeball & \cellcolor{green!0} -- & \cellcolor{green!0} -- & \cellcolor{red!21} 5.4 ± 0.2 & \cellcolor{green!0} -0.1 ± 0.3 & \cellcolor{red!3} 0.8 ± 0.2 & \cellcolor{red!0} 0.1 ± 0.1 & \cellcolor{red!6} 1.5 ± 0.1 \\
2-Ninja & \cellcolor{green!0} -- & \cellcolor{green!0} -- & \cellcolor{green!0} -- & \cellcolor{red!24} 6.2 ± 0.2 & \cellcolor{green!3} -0.9 ± 0.2 & \cellcolor{red!2} 0.6 ± 0.1 & \cellcolor{red!8} 2.0 ± 0.1 \\
3-Starpilot & \cellcolor{green!0} -- & \cellcolor{green!0} -- & \cellcolor{green!0} -- & \cellcolor{green!0} -- & \cellcolor{red!4} 1.1 ± 0.1 & \cellcolor{red!0} 0.1 ± 0.0 & \cellcolor{red!2} 0.6 ± 0.0 \\
4-Bigfish & \cellcolor{green!0} -- & \cellcolor{green!0} -- & \cellcolor{green!0} -- & \cellcolor{green!0} -- & \cellcolor{green!0} -- & \cellcolor{red!7} 1.9 ± 0.2 & \cellcolor{red!7} 1.9 ± 0.2 \\
5-Fruitbot & \cellcolor{green!0} -- & \cellcolor{green!0} -- & \cellcolor{green!0} -- & \cellcolor{green!0} -- & \cellcolor{green!0} -- & \cellcolor{green!0} -- & \cellcolor{green!0} -- \\
\hline
Avg ± SEM & \cellcolor{green!0} -- & \cellcolor{red!10} 2.6 ± 0.2 & \cellcolor{red!8} 2.1 ± 0.1 & \cellcolor{red!10} 2.5 ± 0.2 & \cellcolor{red!0} 0.1 ± 0.1 & \cellcolor{red!2} 0.6 ± 0.1 & \cellcolor{red!4} 1.2 ± 0.0 \\
\hline
\end{tabular}
}
\subfloat[EWC]{ 
 \begin{tabular}{|l|Hlllll|l|}
\hline
 & 0-Climber & 1-Dodgeball & 2-Ninja & 3-Starpilot & 4-Bigfish & 5-Fruitbot & Avg ± SEM \\
\hline
0-Climber & \cellcolor{green!0} -- & \cellcolor{red!4} 1.2 ± 0.1 & \cellcolor{green!2} -0.6 ± 0.1 & \cellcolor{red!2} 0.7 ± 0.1 & \cellcolor{red!0} 0.0 ± 0.0 & \cellcolor{green!0} -0.1 ± 0.0 & \cellcolor{red!0} 0.2 ± 0.0 \\
1-Dodgeball & \cellcolor{green!0} -- & \cellcolor{green!0} -- & \cellcolor{red!20} 5.1 ± 0.4 & \cellcolor{green!2} -0.5 ± 0.3 & \cellcolor{red!3} 0.9 ± 0.3 & \cellcolor{red!0} 0.2 ± 0.2 & \cellcolor{red!5} 1.4 ± 0.1 \\
2-Ninja & \cellcolor{green!0} -- & \cellcolor{green!0} -- & \cellcolor{green!0} -- & \cellcolor{red!17} 4.3 ± 0.3 & \cellcolor{red!0} 0.1 ± 0.1 & \cellcolor{green!0} -0.2 ± 0.1 & \cellcolor{red!5} 1.4 ± 0.1 \\
3-Starpilot & \cellcolor{green!0} -- & \cellcolor{green!0} -- & \cellcolor{green!0} -- & \cellcolor{green!0} -- & \cellcolor{green!2} -0.5 ± 0.2 & \cellcolor{red!1} 0.3 ± 0.2 & \cellcolor{green!0} -0.1 ± 0.1 \\
4-Bigfish & \cellcolor{green!0} -- & \cellcolor{green!0} -- & \cellcolor{green!0} -- & \cellcolor{green!0} -- & \cellcolor{green!0} -- & \cellcolor{green!0} -0.2 ± 0.2 & \cellcolor{green!0} -0.2 ± 0.2 \\
5-Fruitbot & \cellcolor{green!0} -- & \cellcolor{green!0} -- & \cellcolor{green!0} -- & \cellcolor{green!0} -- & \cellcolor{green!0} -- & \cellcolor{green!0} -- & \cellcolor{green!0} -- \\
\hline
Avg ± SEM & \cellcolor{green!0} -- & \cellcolor{red!4} 1.2 ± 0.1 & \cellcolor{red!8} 2.2 ± 0.2 & \cellcolor{red!6} 1.5 ± 0.1 & \cellcolor{red!0} 0.1 ± 0.1 & \cellcolor{red!0} -0.0 ± 0.1 & \cellcolor{red!2} 0.7 ± 0.0 \\
\hline
\end{tabular}
}

\subfloat[Online EWC]{ 
 \begin{tabular}{|l|Hlllll|l|}
\hline
 & 0-Climber & 1-Dodgeball & 2-Ninja & 3-Starpilot & 4-Bigfish & 5-Fruitbot & Avg ± SEM \\
\hline
0-Climber & \cellcolor{green!0} -- & \cellcolor{red!18} 4.6 ± 0.2 & \cellcolor{green!13} -3.3 ± 0.3 & \cellcolor{red!13} 3.4 ± 0.3 & \cellcolor{green!2} -0.6 ± 0.2 & \cellcolor{green!3} -0.8 ± 0.3 & \cellcolor{red!2} 0.7 ± 0.1 \\
1-Dodgeball & \cellcolor{green!0} -- & \cellcolor{green!0} -- & \cellcolor{red!24} 6.1 ± 0.3 & \cellcolor{green!3} -0.8 ± 0.2 & \cellcolor{red!4} 1.0 ± 0.2 & \cellcolor{red!0} 0.2 ± 0.1 & \cellcolor{red!6} 1.6 ± 0.1 \\
2-Ninja & \cellcolor{green!0} -- & \cellcolor{green!0} -- & \cellcolor{green!0} -- & \cellcolor{red!18} 4.7 ± 0.3 & \cellcolor{green!2} -0.6 ± 0.2 & \cellcolor{red!0} 0.0 ± 0.2 & \cellcolor{red!5} 1.4 ± 0.1 \\
3-Starpilot & \cellcolor{green!0} -- & \cellcolor{green!0} -- & \cellcolor{green!0} -- & \cellcolor{green!0} -- & \cellcolor{red!2} 0.5 ± 0.2 & \cellcolor{red!5} 1.4 ± 0.1 & \cellcolor{red!3} 0.9 ± 0.1 \\
4-Bigfish & \cellcolor{green!0} -- & \cellcolor{green!0} -- & \cellcolor{green!0} -- & \cellcolor{green!0} -- & \cellcolor{green!0} -- & \cellcolor{red!0} 0.1 ± 0.2 & \cellcolor{red!0} 0.1 ± 0.2 \\
5-Fruitbot & \cellcolor{green!0} -- & \cellcolor{green!0} -- & \cellcolor{green!0} -- & \cellcolor{green!0} -- & \cellcolor{green!0} -- & \cellcolor{green!0} -- & \cellcolor{green!0} -- \\
\hline
Avg ± SEM & \cellcolor{green!0} -- & \cellcolor{red!18} 4.6 ± 0.2 & \cellcolor{red!5} 1.4 ± 0.2 & \cellcolor{red!9} 2.4 ± 0.2 & \cellcolor{red!0} 0.1 ± 0.1 & \cellcolor{red!0} 0.2 ± 0.1 & \cellcolor{red!4} 1.1 ± 0.0 \\
\hline
\end{tabular}
}
\subfloat[P\&C]{ 
 \begin{tabular}{|l|Hlllll|l|}
\hline
 & 0-Climber & 1-Dodgeball & 2-Ninja & 3-Starpilot & 4-Bigfish & 5-Fruitbot & Avg ± SEM \\
\hline
0-Climber & \cellcolor{green!0} -- & \cellcolor{red!2} 0.5 ± 0.3 & \cellcolor{red!4} 1.0 ± 0.5 & \cellcolor{green!1} -0.4 ± 0.2 & \cellcolor{red!4} 1.1 ± 0.4 & \cellcolor{green!2} -0.5 ± 0.2 & \cellcolor{red!1} 0.3 ± 0.1 \\
1-Dodgeball & \cellcolor{green!0} -- & \cellcolor{green!0} -- & \cellcolor{green!0} -0.2 ± 0.2 & \cellcolor{red!12} 3.0 ± 0.2 & \cellcolor{green!0} -0.2 ± 0.2 & \cellcolor{red!1} 0.4 ± 0.2 & \cellcolor{red!2} 0.7 ± 0.1 \\
2-Ninja & \cellcolor{green!0} -- & \cellcolor{green!0} -- & \cellcolor{green!0} -- & \cellcolor{green!2} -0.6 ± 0.1 & \cellcolor{red!8} 2.2 ± 0.3 & \cellcolor{green!0} -0.1 ± 0.0 & \cellcolor{red!2} 0.5 ± 0.1 \\
3-Starpilot & \cellcolor{green!0} -- & \cellcolor{green!0} -- & \cellcolor{green!0} -- & \cellcolor{green!0} -- & \cellcolor{green!2} -0.6 ± 0.2 & \cellcolor{red!6} 1.7 ± 0.2 & \cellcolor{red!2} 0.5 ± 0.1 \\
4-Bigfish & \cellcolor{green!0} -- & \cellcolor{green!0} -- & \cellcolor{green!0} -- & \cellcolor{green!0} -- & \cellcolor{green!0} -- & \cellcolor{green!0} -0.2 ± 0.2 & \cellcolor{green!0} -0.2 ± 0.2 \\
5-Fruitbot & \cellcolor{green!0} -- & \cellcolor{green!0} -- & \cellcolor{green!0} -- & \cellcolor{green!0} -- & \cellcolor{green!0} -- & \cellcolor{green!0} -- & \cellcolor{green!0} -- \\
\hline
Avg ± SEM & \cellcolor{green!0} -- & \cellcolor{red!2} 0.5 ± 0.3 & \cellcolor{red!1} 0.4 ± 0.2 & \cellcolor{red!2} 0.6 ± 0.1 & \cellcolor{red!2} 0.6 ± 0.2 & \cellcolor{red!1} 0.3 ± 0.1 & \cellcolor{red!2} 0.5 ± 0.0 \\
\hline
\end{tabular}
}

\subfloat[CLEAR]{ 
 \begin{tabular}{|l|Hlllll|l|}
\hline
 & 0-Climber & 1-Dodgeball & 2-Ninja & 3-Starpilot & 4-Bigfish & 5-Fruitbot & Avg ± SEM \\
\hline
0-Climber & \cellcolor{green!0} -- & \cellcolor{green!1} -0.4 ± 0.2 & \cellcolor{red!0} 0.2 ± 0.1 & \cellcolor{red!3} 0.9 ± 0.2 & \cellcolor{green!0} -0.1 ± 0.2 & \cellcolor{red!0} 0.1 ± 0.1 & \cellcolor{red!0} 0.2 ± 0.0 \\
1-Dodgeball & \cellcolor{green!0} -- & \cellcolor{green!0} -- & \cellcolor{red!0} -0.0 ± 0.2 & \cellcolor{red!1} 0.3 ± 0.2 & \cellcolor{green!0} -0.1 ± 0.1 & \cellcolor{red!1} 0.3 ± 0.2 & \cellcolor{red!0} 0.1 ± 0.0 \\
2-Ninja & \cellcolor{green!0} -- & \cellcolor{green!0} -- & \cellcolor{green!0} -- & \cellcolor{green!0} -0.1 ± 0.3 & \cellcolor{green!2} -0.5 ± 0.4 & \cellcolor{red!2} 0.6 ± 0.3 & \cellcolor{red!0} 0.0 ± 0.1 \\
3-Starpilot & \cellcolor{green!0} -- & \cellcolor{green!0} -- & \cellcolor{green!0} -- & \cellcolor{green!0} -- & \cellcolor{green!6} -1.6 ± 0.1 & \cellcolor{green!1} -0.4 ± 0.1 & \cellcolor{green!4} -1.0 ± 0.0 \\
4-Bigfish & \cellcolor{green!0} -- & \cellcolor{green!0} -- & \cellcolor{green!0} -- & \cellcolor{green!0} -- & \cellcolor{green!0} -- & \cellcolor{red!0} 0.2 ± 0.1 & \cellcolor{red!0} 0.2 ± 0.1 \\
5-Fruitbot & \cellcolor{green!0} -- & \cellcolor{green!0} -- & \cellcolor{green!0} -- & \cellcolor{green!0} -- & \cellcolor{green!0} -- & \cellcolor{green!0} -- & \cellcolor{green!0} -- \\
\hline
Avg ± SEM & \cellcolor{green!0} -- & \cellcolor{green!1} -0.4 ± 0.2 & \cellcolor{red!0} 0.1 ± 0.1 & \cellcolor{red!1} 0.4 ± 0.1 & \cellcolor{green!2} -0.5 ± 0.1 & \cellcolor{red!0} 0.2 ± 0.1 & \cellcolor{red!0} -0.0 ± 0.0 \\
\hline
\end{tabular}
}

\vspace{2em}
\caption{Procgen Forgetting metrics.}
\label{tab:procgen_forgetting}
\end{table}

\clearpage

\subsection{Procgen Metrics: Transfer}
\label{sec:procgen_transfer}

\begin{table}[H]
\tablemarginhack
\setlength\tabcolsep{2pt} 
\tiny

\subfloat[IMPALA]{ 
 \begin{tabular}{|l|lllllH|l|}
\hline
 & 0-Climber & 1-Dodgeball & 2-Ninja & 3-Starpilot & 4-Bigfish & 5-Fruitbot & Avg ± SEM \\
\hline
0-Climber & \cellcolor{green!0} -- & \cellcolor{green!0} -- & \cellcolor{green!0} -- & \cellcolor{green!0} -- & \cellcolor{green!0} -- & \cellcolor{green!0} -- & \cellcolor{green!0} -- \\
1-Dodgeball & \cellcolor{red!0} 0.0 ± 0.0 & \cellcolor{green!0} -- & \cellcolor{green!0} -- & \cellcolor{green!0} -- & \cellcolor{green!0} -- & \cellcolor{green!0} -- & \cellcolor{red!0} 0.0 ± 0.0 \\
2-Ninja & \cellcolor{green!12} 3.0 ± 0.8 & \cellcolor{red!16} -4.2 ± 0.5 & \cellcolor{green!0} -- & \cellcolor{green!0} -- & \cellcolor{green!0} -- & \cellcolor{green!0} -- & \cellcolor{red!2} -0.6 ± 0.3 \\
3-Starpilot & \cellcolor{red!0} -0.0 ± 0.0 & \cellcolor{green!1} 0.3 ± 0.0 & \cellcolor{red!1} -0.3 ± 0.0 & \cellcolor{green!0} -- & \cellcolor{green!0} -- & \cellcolor{green!0} -- & \cellcolor{red!0} 0.0 ± 0.0 \\
4-Bigfish & \cellcolor{green!4} 1.1 ± 0.4 & \cellcolor{red!8} -2.0 ± 0.2 & \cellcolor{green!1} 0.3 ± 0.1 & \cellcolor{red!0} -0.2 ± 0.1 & \cellcolor{green!0} -- & \cellcolor{green!0} -- & \cellcolor{red!0} -0.2 ± 0.1 \\
5-Fruitbot & \cellcolor{green!0} 0.1 ± 0.1 & \cellcolor{red!0} -0.2 ± 0.0 & \cellcolor{red!0} -0.1 ± 0.0 & \cellcolor{green!1} 0.3 ± 0.0 & \cellcolor{red!1} -0.3 ± 0.0 & \cellcolor{green!0} -- & \cellcolor{red!0} -0.0 ± 0.0 \\
\hline
Avg ± SEM & \cellcolor{green!3} 0.8 ± 0.2 & \cellcolor{red!6} -1.5 ± 0.1 & \cellcolor{red!0} -0.0 ± 0.0 & \cellcolor{red!0} 0.0 ± 0.0 & \cellcolor{red!1} -0.3 ± 0.0 & \cellcolor{green!0} -- & \cellcolor{red!0} -0.1 ± 0.0 \\
\hline
\end{tabular}
}
\subfloat[EWC]{ 
 \begin{tabular}{|l|lllllH|l|}
\hline
 & 0-Climber & 1-Dodgeball & 2-Ninja & 3-Starpilot & 4-Bigfish & 5-Fruitbot & Avg ± SEM \\
\hline
0-Climber & \cellcolor{green!0} -- & \cellcolor{green!0} -- & \cellcolor{green!0} -- & \cellcolor{green!0} -- & \cellcolor{green!0} -- & \cellcolor{green!0} -- & \cellcolor{green!0} -- \\
1-Dodgeball & \cellcolor{red!1} -0.4 ± 0.3 & \cellcolor{green!0} -- & \cellcolor{green!0} -- & \cellcolor{green!0} -- & \cellcolor{green!0} -- & \cellcolor{green!0} -- & \cellcolor{red!1} -0.4 ± 0.3 \\
2-Ninja & \cellcolor{green!9} 2.3 ± 0.7 & \cellcolor{red!13} -3.3 ± 0.4 & \cellcolor{green!0} -- & \cellcolor{green!0} -- & \cellcolor{green!0} -- & \cellcolor{green!0} -- & \cellcolor{red!2} -0.5 ± 0.2 \\
3-Starpilot & \cellcolor{red!5} -1.3 ± 0.4 & \cellcolor{green!14} 3.5 ± 0.2 & \cellcolor{red!12} -3.1 ± 0.2 & \cellcolor{green!0} -- & \cellcolor{green!0} -- & \cellcolor{green!0} -- & \cellcolor{red!1} -0.3 ± 0.2 \\
4-Bigfish & \cellcolor{green!15} 3.8 ± 0.6 & \cellcolor{red!17} -4.3 ± 0.4 & \cellcolor{green!6} 1.6 ± 0.4 & \cellcolor{red!6} -1.5 ± 0.3 & \cellcolor{green!0} -- & \cellcolor{green!0} -- & \cellcolor{red!0} -0.1 ± 0.1 \\
5-Fruitbot & \cellcolor{red!0} -0.1 ± 0.6 & \cellcolor{red!8} -2.1 ± 0.3 & \cellcolor{green!6} 1.6 ± 0.2 & \cellcolor{red!1} -0.4 ± 0.3 & \cellcolor{green!2} 0.5 ± 0.2 & \cellcolor{green!0} -- & \cellcolor{red!0} -0.1 ± 0.1 \\
\hline
Avg ± SEM & \cellcolor{green!3} 0.9 ± 0.2 & \cellcolor{red!6} -1.5 ± 0.1 & \cellcolor{red!0} 0.0 ± 0.1 & \cellcolor{red!4} -1.0 ± 0.3 & \cellcolor{green!2} 0.5 ± 0.2 & \cellcolor{green!0} -- & \cellcolor{red!0} -0.2 ± 0.1 \\
\hline
\end{tabular}
}

\subfloat[Online EWC]{ 
 \begin{tabular}{|l|lllllH|l|}
\hline
 & 0-Climber & 1-Dodgeball & 2-Ninja & 3-Starpilot & 4-Bigfish & 5-Fruitbot & Avg ± SEM \\
\hline
0-Climber & \cellcolor{green!0} -- & \cellcolor{green!0} -- & \cellcolor{green!0} -- & \cellcolor{green!0} -- & \cellcolor{green!0} -- & \cellcolor{green!0} -- & \cellcolor{green!0} -- \\
1-Dodgeball & \cellcolor{red!2} -0.6 ± 0.5 & \cellcolor{green!0} -- & \cellcolor{green!0} -- & \cellcolor{green!0} -- & \cellcolor{green!0} -- & \cellcolor{green!0} -- & \cellcolor{red!2} -0.6 ± 0.5 \\
2-Ninja & \cellcolor{green!18} 4.5 ± 0.5 & \cellcolor{red!17} -4.3 ± 0.5 & \cellcolor{green!0} -- & \cellcolor{green!0} -- & \cellcolor{green!0} -- & \cellcolor{green!0} -- & \cellcolor{green!0} 0.1 ± 0.1 \\
3-Starpilot & \cellcolor{red!2} -0.7 ± 0.2 & \cellcolor{green!8} 2.2 ± 0.0 & \cellcolor{red!8} -2.1 ± 0.1 & \cellcolor{green!0} -- & \cellcolor{green!0} -- & \cellcolor{green!0} -- & \cellcolor{red!0} -0.2 ± 0.1 \\
4-Bigfish & \cellcolor{green!7} 1.8 ± 0.7 & \cellcolor{red!10} -2.7 ± 0.3 & \cellcolor{green!4} 1.1 ± 0.1 & \cellcolor{red!3} -0.8 ± 0.1 & \cellcolor{green!0} -- & \cellcolor{green!0} -- & \cellcolor{red!0} -0.2 ± 0.1 \\
5-Fruitbot & \cellcolor{green!1} 0.3 ± 0.2 & \cellcolor{red!2} -0.5 ± 0.1 & \cellcolor{green!1} 0.3 ± 0.0 & \cellcolor{red!0} -0.1 ± 0.0 & \cellcolor{red!0} -0.1 ± 0.0 & \cellcolor{green!0} -- & \cellcolor{red!0} -0.0 ± 0.0 \\
\hline
Avg ± SEM & \cellcolor{green!4} 1.1 ± 0.2 & \cellcolor{red!5} -1.3 ± 0.1 & \cellcolor{red!0} -0.2 ± 0.0 & \cellcolor{red!2} -0.5 ± 0.1 & \cellcolor{red!0} -0.1 ± 0.0 & \cellcolor{green!0} -- & \cellcolor{red!0} -0.1 ± 0.1 \\
\hline
\end{tabular}
}
\subfloat[P\&C]{ 
 \begin{tabular}{|l|lllllH|l|}
\hline
 & 0-Climber & 1-Dodgeball & 2-Ninja & 3-Starpilot & 4-Bigfish & 5-Fruitbot & Avg ± SEM \\
\hline
0-Climber & \cellcolor{green!0} -- & \cellcolor{green!0} -- & \cellcolor{green!0} -- & \cellcolor{green!0} -- & \cellcolor{green!0} -- & \cellcolor{green!0} -- & \cellcolor{green!0} -- \\
1-Dodgeball & \cellcolor{red!2} -0.7 ± 0.4 & \cellcolor{green!0} -- & \cellcolor{green!0} -- & \cellcolor{green!0} -- & \cellcolor{green!0} -- & \cellcolor{green!0} -- & \cellcolor{red!2} -0.7 ± 0.4 \\
2-Ninja & \cellcolor{green!6} 1.6 ± 0.5 & \cellcolor{red!1} -0.3 ± 0.1 & \cellcolor{green!0} -- & \cellcolor{green!0} -- & \cellcolor{green!0} -- & \cellcolor{green!0} -- & \cellcolor{green!2} 0.7 ± 0.2 \\
3-Starpilot & \cellcolor{red!3} -0.9 ± 0.3 & \cellcolor{green!3} 0.8 ± 0.2 & \cellcolor{green!0} 0.2 ± 0.2 & \cellcolor{green!0} -- & \cellcolor{green!0} -- & \cellcolor{green!0} -- & \cellcolor{red!0} 0.0 ± 0.2 \\
4-Bigfish & \cellcolor{green!5} 1.4 ± 0.4 & \cellcolor{red!2} -0.7 ± 0.3 & \cellcolor{red!2} -0.7 ± 0.4 & \cellcolor{green!1} 0.3 ± 0.2 & \cellcolor{green!0} -- & \cellcolor{green!0} -- & \cellcolor{green!0} 0.1 ± 0.1 \\
5-Fruitbot & \cellcolor{green!0} 0.1 ± 0.1 & \cellcolor{red!0} -0.1 ± 0.1 & \cellcolor{red!0} -0.2 ± 0.1 & \cellcolor{green!0} 0.1 ± 0.1 & \cellcolor{green!0} 0.2 ± 0.1 & \cellcolor{green!0} -- & \cellcolor{red!0} 0.0 ± 0.0 \\
\hline
Avg ± SEM & \cellcolor{green!1} 0.3 ± 0.2 & \cellcolor{red!0} -0.1 ± 0.1 & \cellcolor{red!1} -0.3 ± 0.1 & \cellcolor{green!0} 0.2 ± 0.1 & \cellcolor{green!0} 0.2 ± 0.1 & \cellcolor{green!0} -- & \cellcolor{green!0} 0.1 ± 0.1 \\
\hline
\end{tabular}
}

\subfloat[CLEAR]{ 
 \begin{tabular}{|l|lllllH|l|}
\hline
 & 0-Climber & 1-Dodgeball & 2-Ninja & 3-Starpilot & 4-Bigfish & 5-Fruitbot & Avg ± SEM \\
\hline
0-Climber & \cellcolor{green!0} -- & \cellcolor{green!0} -- & \cellcolor{green!0} -- & \cellcolor{green!0} -- & \cellcolor{green!0} -- & \cellcolor{green!0} -- & \cellcolor{green!0} -- \\
1-Dodgeball & \cellcolor{red!0} -0.2 ± 0.1 & \cellcolor{green!0} -- & \cellcolor{green!0} -- & \cellcolor{green!0} -- & \cellcolor{green!0} -- & \cellcolor{green!0} -- & \cellcolor{red!0} -0.2 ± 0.1 \\
2-Ninja & \cellcolor{green!12} 3.2 ± 0.5 & \cellcolor{red!13} -3.3 ± 0.3 & \cellcolor{green!0} -- & \cellcolor{green!0} -- & \cellcolor{green!0} -- & \cellcolor{green!0} -- & \cellcolor{red!0} -0.0 ± 0.2 \\
3-Starpilot & \cellcolor{red!0} -0.1 ± 0.0 & \cellcolor{green!0} 0.1 ± 0.0 & \cellcolor{red!0} -0.0 ± 0.0 & \cellcolor{green!0} -- & \cellcolor{green!0} -- & \cellcolor{green!0} -- & \cellcolor{red!0} -0.0 ± 0.0 \\
4-Bigfish & \cellcolor{green!2} 0.6 ± 0.5 & \cellcolor{red!4} -1.2 ± 0.2 & \cellcolor{red!0} 0.0 ± 0.1 & \cellcolor{red!0} -0.0 ± 0.1 & \cellcolor{green!0} -- & \cellcolor{green!0} -- & \cellcolor{red!0} -0.1 ± 0.1 \\
5-Fruitbot & \cellcolor{red!0} 0.0 ± 0.1 & \cellcolor{red!0} -0.0 ± 0.0 & \cellcolor{red!0} -0.0 ± 0.0 & \cellcolor{red!0} -0.0 ± 0.0 & \cellcolor{red!0} 0.0 ± 0.0 & \cellcolor{green!0} -- & \cellcolor{red!0} -0.0 ± 0.0 \\
\hline
Avg ± SEM & \cellcolor{green!2} 0.7 ± 0.1 & \cellcolor{red!4} -1.1 ± 0.1 & \cellcolor{red!0} -0.0 ± 0.0 & \cellcolor{red!0} -0.0 ± 0.0 & \cellcolor{red!0} 0.0 ± 0.0 & \cellcolor{green!0} -- & \cellcolor{red!0} -0.1 ± 0.0 \\
\hline
\end{tabular}
}
\vspace{2em}
\caption{Procgen Transfer metrics.}
\label{tab:procgen_transfer}
\end{table}

\clearpage

\subsection{MiniHack Metrics: Forgetting}
\label{sec:minihack_forgetting}

\begin{table}[ht]
\tablemarginhack
\setlength\tabcolsep{2pt} 
\tiny

\subfloat[IMPALA]{ 
 \begin{tabular}{|l|Hllllllllllllll|l|}
\hline
 & 0-R-Ran.. & 1-R-Dar.. & 2-R-Mon.. & 3-R-Tra.. & 4-R-Ult.. & 5-Cor.. & 6-Cor.. & 7-KeyR & 8-KeyR-.. & 9-River-Na.. & 10-River-M.. & 11-River-L.. & 12-HnS.. & 13-HnS.. & 14-Corrido.. & Avg ± SEM \\
\hline
0-R-Ran.. & \cellcolor{green!0} -- & \cellcolor{green!1} -0.3 ± 0.3 & \cellcolor{green!2} -0.5 ± 0.1 & \cellcolor{green!1} -0.4 ± 0.4 & \cellcolor{red!0} 0.0 ± 0.3 & \cellcolor{red!0} -0.0 ± 0.2 & \cellcolor{red!9} 2.3 ± 0.4 & \cellcolor{red!5} 1.4 ± 0.3 & \cellcolor{green!0} -0.2 ± 0.1 & \cellcolor{green!0} -0.2 ± 0.3 & \cellcolor{green!0} -0.2 ± 0.2 & \cellcolor{red!4} 1.0 ± 0.5 & \cellcolor{green!7} -1.8 ± 0.5 & \cellcolor{green!2} -0.6 ± 0.4 & \cellcolor{red!9} 2.4 ± 0.5 & \cellcolor{red!0} 0.2 ± 0.1 \\
1-R-Dar.. & \cellcolor{green!0} -- & \cellcolor{green!0} -- & \cellcolor{green!3} -0.8 ± 0.2 & \cellcolor{red!6} 1.6 ± 0.5 & \cellcolor{green!5} -1.3 ± 0.5 & \cellcolor{green!0} -0.1 ± 0.3 & \cellcolor{red!10} 2.5 ± 0.4 & \cellcolor{red!0} 0.2 ± 0.3 & \cellcolor{green!2} -0.5 ± 0.6 & \cellcolor{red!0} 0.2 ± 0.2 & \cellcolor{red!1} 0.3 ± 0.4 & \cellcolor{red!2} 0.6 ± 0.2 & \cellcolor{green!6} -1.6 ± 0.6 & \cellcolor{green!4} -1.0 ± 0.4 & \cellcolor{red!7} 1.9 ± 0.5 & \cellcolor{red!0} 0.2 ± 0.0 \\
2-R-Mon.. & \cellcolor{green!0} -- & \cellcolor{green!0} -- & \cellcolor{green!0} -- & \cellcolor{green!1} -0.4 ± 0.2 & \cellcolor{green!0} -0.1 ± 0.3 & \cellcolor{green!1} -0.4 ± 0.2 & \cellcolor{red!12} 3.2 ± 0.4 & \cellcolor{red!7} 1.9 ± 0.4 & \cellcolor{green!1} -0.4 ± 0.2 & \cellcolor{red!0} 0.2 ± 0.3 & \cellcolor{green!0} -0.2 ± 0.4 & \cellcolor{red!5} 1.4 ± 0.6 & \cellcolor{green!8} -2.2 ± 0.5 & \cellcolor{green!7} -1.8 ± 0.7 & \cellcolor{red!10} 2.6 ± 0.7 & \cellcolor{red!1} 0.3 ± 0.1 \\
\hline
\end{tabular}
}

\subfloat[CLEAR]{ 
 \begin{tabular}{|l|Hllllllllllllll|l|}
\hline
 & 0-R-Ran.. & 1-R-Dar.. & 2-R-Mon.. & 3-R-Tra.. & 4-R-Ult.. & 5-Cor.. & 6-Cor.. & 7-KeyR & 8-KeyR-.. & 9-River-Na.. & 10-River-M.. & 11-River-L.. & 12-HnS.. & 13-HnS.. & 14-Corrido.. & Avg ± SEM \\
\hline
0-R-Ran.. & \cellcolor{green!0} -- & \cellcolor{green!1} -0.4 ± 0.2 & \cellcolor{green!2} -0.6 ± 0.2 & \cellcolor{green!1} -0.4 ± 0.2 & \cellcolor{red!0} 0.1 ± 0.2 & \cellcolor{green!2} -0.6 ± 0.3 & \cellcolor{green!0} -0.2 ± 0.3 & \cellcolor{red!1} 0.4 ± 0.2 & \cellcolor{red!0} 0.0 ± 0.2 & \cellcolor{red!3} 0.9 ± 0.4 & \cellcolor{red!5} 1.3 ± 0.5 & \cellcolor{red!1} 0.3 ± 0.2 & \cellcolor{green!2} -0.5 ± 0.2 & \cellcolor{red!1} 0.3 ± 0.2 & \cellcolor{red!2} 0.5 ± 0.2 & \cellcolor{red!0} 0.1 ± 0.0 \\
1-R-Dar.. & \cellcolor{green!0} -- & \cellcolor{green!0} -- & \cellcolor{green!10} -2.7 ± 0.3 & \cellcolor{green!2} -0.5 ± 0.3 & \cellcolor{red!0} 0.2 ± 0.2 & \cellcolor{green!2} -0.6 ± 0.4 & \cellcolor{green!4} -1.2 ± 0.4 & \cellcolor{green!0} -0.2 ± 0.4 & \cellcolor{red!0} 0.0 ± 0.2 & \cellcolor{green!0} -0.2 ± 0.2 & \cellcolor{red!0} 0.1 ± 0.3 & \cellcolor{green!0} -0.2 ± 0.2 & \cellcolor{red!6} 1.5 ± 0.4 & \cellcolor{red!6} 1.5 ± 0.6 & \cellcolor{green!2} -0.5 ± 0.4 & \cellcolor{green!0} -0.2 ± 0.1 \\
2-R-Mon.. & \cellcolor{green!0} -- & \cellcolor{green!0} -- & \cellcolor{green!0} -- & \cellcolor{green!2} -0.6 ± 0.2 & \cellcolor{green!1} -0.3 ± 0.2 & \cellcolor{green!2} -0.5 ± 0.3 & \cellcolor{red!0} 0.0 ± 0.3 & \cellcolor{red!1} 0.3 ± 0.3 & \cellcolor{green!0} -0.2 ± 0.2 & \cellcolor{red!4} 1.0 ± 0.4 & \cellcolor{red!5} 1.3 ± 0.5 & \cellcolor{red!2} 0.5 ± 0.2 & \cellcolor{green!3} -0.8 ± 0.2 & \cellcolor{green!0} -0.1 ± 0.3 & \cellcolor{red!4} 1.1 ± 0.2 & \cellcolor{red!0} 0.2 ± 0.1 \\
\hline
\end{tabular}
}

\vspace{2em}
\caption{MiniHack Forgetting metrics.}
\label{tab:minihack_forgetting}
\end{table}

\clearpage
\subsection{MiniHack Metrics: Transfer}
\label{sec:minihack_transfer}

\begin{table}[ht]
\tablemarginhack
\setlength\tabcolsep{2pt} 
\tiny

\subfloat[IMPALA]{ 
 \begin{tabular}{|l|llllllllllllllH|l|}
\hline
 & 0-R-Ran.. & 1-R-Dar.. & 2-R-Mon.. & 3-R-Tra.. & 4-R-Ult.. & 5-Cor.. & 6-Cor.. & 7-KeyR & 8-KeyR-.. & 9-River-Na.. & 10-River-M.. & 11-River-L.. & 12-HideNSe.. & 13-HideNSe.. & 14-Corrido.. & Avg ± SEM \\
\hline
0-R-Ran.. & \cellcolor{green!0} -- & \cellcolor{green!0} -- & \cellcolor{green!0} -- & \cellcolor{green!0} -- & \cellcolor{green!0} -- & \cellcolor{green!0} -- & \cellcolor{green!0} -- & \cellcolor{green!0} -- & \cellcolor{green!0} -- & \cellcolor{green!0} -- & \cellcolor{green!0} -- & \cellcolor{green!0} -- & \cellcolor{green!0} -- & \cellcolor{green!0} -- & \cellcolor{green!0} -- & \cellcolor{green!0} -- \\
1-R-Dar.. & \cellcolor{green!16} 4.0 ± 0.3 & \cellcolor{green!0} -- & \cellcolor{green!0} -- & \cellcolor{green!0} -- & \cellcolor{green!0} -- & \cellcolor{green!0} -- & \cellcolor{green!0} -- & \cellcolor{green!0} -- & \cellcolor{green!0} -- & \cellcolor{green!0} -- & \cellcolor{green!0} -- & \cellcolor{green!0} -- & \cellcolor{green!0} -- & \cellcolor{green!0} -- & \cellcolor{green!0} -- & \cellcolor{green!16} 4.0 ± 0.3 \\
2-R-Mon.. & \cellcolor{green!34} 8.6 ± 1.1 & \cellcolor{green!0} 0.2 ± 0.3 & \cellcolor{green!0} -- & \cellcolor{green!0} -- & \cellcolor{green!0} -- & \cellcolor{green!0} -- & \cellcolor{green!0} -- & \cellcolor{green!0} -- & \cellcolor{green!0} -- & \cellcolor{green!0} -- & \cellcolor{green!0} -- & \cellcolor{green!0} -- & \cellcolor{green!0} -- & \cellcolor{green!0} -- & \cellcolor{green!0} -- & \cellcolor{green!17} 4.4 ± 0.5 \\
3-R-Tra.. & \cellcolor{green!30} 7.5 ± 0.9 & \cellcolor{green!3} 0.9 ± 0.4 & \cellcolor{green!4} 1.2 ± 0.5 & \cellcolor{green!0} -- & \cellcolor{green!0} -- & \cellcolor{green!0} -- & \cellcolor{green!0} -- & \cellcolor{green!0} -- & \cellcolor{green!0} -- & \cellcolor{green!0} -- & \cellcolor{green!0} -- & \cellcolor{green!0} -- & \cellcolor{green!0} -- & \cellcolor{green!0} -- & \cellcolor{green!0} -- & \cellcolor{green!12} 3.2 ± 0.4 \\
\hline
\end{tabular}
}

\subfloat[CLEAR]{ 
 \begin{tabular}{|l|llllllllllllllH|l|}
\hline
 & 0-R-Ran.. & 1-R-Dar.. & 2-R-Mon.. & 3-R-Tra.. & 4-R-Ult.. & 5-Cor.. & 6-Cor.. & 7-KeyR & 8-KeyR-.. & 9-River-Na.. & 10-River-M.. & 11-River-L.. & 12-HideNSe.. & 13-HideNSe.. & 14-Corrido.. & Avg ± SEM \\
\hline
0-R-Ran.. & \cellcolor{green!0} -- & \cellcolor{green!0} -- & \cellcolor{green!0} -- & \cellcolor{green!0} -- & \cellcolor{green!0} -- & \cellcolor{green!0} -- & \cellcolor{green!0} -- & \cellcolor{green!0} -- & \cellcolor{green!0} -- & \cellcolor{green!0} -- & \cellcolor{green!0} -- & \cellcolor{green!0} -- & \cellcolor{green!0} -- & \cellcolor{green!0} -- & \cellcolor{green!0} -- & \cellcolor{green!0} -- \\
1-R-Dar.. & \cellcolor{green!10} 2.5 ± 0.5 & \cellcolor{green!0} -- & \cellcolor{green!0} -- & \cellcolor{green!0} -- & \cellcolor{green!0} -- & \cellcolor{green!0} -- & \cellcolor{green!0} -- & \cellcolor{green!0} -- & \cellcolor{green!0} -- & \cellcolor{green!0} -- & \cellcolor{green!0} -- & \cellcolor{green!0} -- & \cellcolor{green!0} -- & \cellcolor{green!0} -- & \cellcolor{green!0} -- & \cellcolor{green!10} 2.5 ± 0.5 \\
2-R-Mon.. & \cellcolor{green!28} 7.1 ± 0.6 & \cellcolor{green!0} 0.2 ± 0.2 & \cellcolor{green!0} -- & \cellcolor{green!0} -- & \cellcolor{green!0} -- & \cellcolor{green!0} -- & \cellcolor{green!0} -- & \cellcolor{green!0} -- & \cellcolor{green!0} -- & \cellcolor{green!0} -- & \cellcolor{green!0} -- & \cellcolor{green!0} -- & \cellcolor{green!0} -- & \cellcolor{green!0} -- & \cellcolor{green!0} -- & \cellcolor{green!14} 3.6 ± 0.4 \\
3-R-Tra.. & \cellcolor{green!28} 7.0 ± 0.8 & \cellcolor{green!2} 0.7 ± 0.2 & \cellcolor{green!3} 0.9 ± 0.2 & \cellcolor{green!0} -- & \cellcolor{green!0} -- & \cellcolor{green!0} -- & \cellcolor{green!0} -- & \cellcolor{green!0} -- & \cellcolor{green!0} -- & \cellcolor{green!0} -- & \cellcolor{green!0} -- & \cellcolor{green!0} -- & \cellcolor{green!0} -- & \cellcolor{green!0} -- & \cellcolor{green!0} -- & \cellcolor{green!11} 2.9 ± 0.3 \\
\hline
\end{tabular}
}
\vspace{2em}
\caption{MiniHack Transfer metrics. 
}
\label{tab:minihack_transfer}
\end{table}

\clearpage

\subsection{CHORES Metrics: Forgetting}
\label{sec:chores_forgetting}

\begin{table}[H]
\tablemarginhack
\setlength\tabcolsep{2pt} 
\tiny

\subfloat[EWC]{ 
 \begin{tabular}{|l|lll|l|}
\hline
 & R402 & R419 & R423 & Avg ± SEM \\
\hline
R402 & \cellcolor{green!0} -- & \cellcolor{red!0} 0.0 ± 0.0 & \cellcolor{red!5} 1.4 ± 1.4 & \cellcolor{red!2} 0.7 ± 0.7 \\
R419 & \cellcolor{green!0} -- & \cellcolor{green!0} -- & \cellcolor{green!13} -3.3 ± 3.3 & \cellcolor{green!13} -3.3 ± 3.3 \\
R423 & \cellcolor{green!0} -- & \cellcolor{green!0} -- & \cellcolor{green!0} -- & \cellcolor{green!0} -- \\
\hline
Avg ± SEM & \cellcolor{green!0} -- & \cellcolor{red!0} 0.0 ± 0.0 & \cellcolor{green!4} -1.0 ± 1.0 & \cellcolor{green!2} -0.6 ± 0.6 \\
\hline
\end{tabular}
}
\hspace{1em}
\subfloat[P\&C]{ 
 \begin{tabular}{|l|lll|l|}
\hline
 & R402 & R419 & R423 & Avg ± SEM \\
\hline
R402 & \cellcolor{green!0} -- & \cellcolor{red!0} 0.0 ± 0.0 & \cellcolor{red!0} 0.0 ± 0.0 & \cellcolor{red!0} 0.0 ± 0.0 \\
R419 & \cellcolor{green!0} -- & \cellcolor{green!0} -- & \cellcolor{red!0} 0.0 ± 0.0 & \cellcolor{red!0} 0.0 ± 0.0 \\
R423 & \cellcolor{green!0} -- & \cellcolor{green!0} -- & \cellcolor{green!0} -- & \cellcolor{green!0} -- \\
\hline
Avg ± SEM & \cellcolor{green!0} -- & \cellcolor{red!0} 0.0 ± 0.0 & \cellcolor{red!0} 0.0 ± 0.0 & \cellcolor{red!0} 0.0 ± 0.0 \\
\hline
\end{tabular}
}
\hspace{1em}
\subfloat[CLEAR]{ 
 \begin{tabular}{|l|lll|l|}
\hline
 & R402 & R419 & R423 & Avg ± SEM \\
\hline
R402 & \cellcolor{green!0} -- & \cellcolor{red!16} 4.1 ± 2.4 & \cellcolor{green!24} -6.1 ± 3.1 & \cellcolor{green!4} -1.0 ± 1.0 \\
R419 & \cellcolor{green!0} -- & \cellcolor{green!0} -- & \cellcolor{green!12} -3.1 ± 3.1 & \cellcolor{green!12} -3.1 ± 3.1 \\
R423 & \cellcolor{green!0} -- & \cellcolor{green!0} -- & \cellcolor{green!0} -- & \cellcolor{green!0} -- \\
\hline
Avg ± SEM & \cellcolor{green!0} -- & \cellcolor{red!16} 4.1 ± 2.4 & \cellcolor{green!18} -4.6 ± 2.8 & \cellcolor{green!6} -1.7 ± 1.7 \\
\hline
\end{tabular}
}

\vspace{2em}
\caption{CHORES: Mem-VaryRoom Forgetting metrics.}
\label{tab:chores_vary_env_forgetting}
\end{table}

\begin{table}[H]
\tablemarginhack
\setlength\tabcolsep{2pt} 
\tiny

\subfloat[EWC]{ 
 \begin{tabular}{|l|lll|l|}
\hline
 & Hang TP & Counter & Cabinet & Avg ± SEM \\
\hline
Hang TP & \cellcolor{green!0} -- & \cellcolor{red!0} -0.0 ± 0.0 & \cellcolor{red!0} 0.0 ± 0.0 & \cellcolor{red!0} -0.0 ± 0.0 \\
Counter & \cellcolor{green!0} -- & \cellcolor{green!0} -- & \cellcolor{red!25} 6.3 ± 4.3 & \cellcolor{red!25} 6.3 ± 4.3 \\
Cabinet & \cellcolor{green!0} -- & \cellcolor{green!0} -- & \cellcolor{green!0} -- & \cellcolor{green!0} -- \\
\hline
Avg ± SEM & \cellcolor{green!0} -- & \cellcolor{red!0} -0.0 ± 0.0 & \cellcolor{red!12} 3.1 ± 2.2 & \cellcolor{red!8} 2.1 ± 1.4 \\
\hline
\end{tabular}
}
\hspace{1em}
\subfloat[P\&C]{ 
 \begin{tabular}{|l|lll|l|}
\hline
 & Hang TP & Counter & Cabinet & Avg ± SEM \\
\hline
Hang TP & \cellcolor{green!0} -- & \cellcolor{red!0} 0.1 ± 0.1 & \cellcolor{green!13} -3.3 ± 3.2 & \cellcolor{green!6} -1.6 ± 1.6 \\
Counter & \cellcolor{green!0} -- & \cellcolor{green!0} -- & \cellcolor{green!14} -3.5 ± 3.3 & \cellcolor{green!14} -3.5 ± 3.3 \\
Cabinet & \cellcolor{green!0} -- & \cellcolor{green!0} -- & \cellcolor{green!0} -- & \cellcolor{green!0} -- \\
\hline
Avg ± SEM & \cellcolor{green!0} -- & \cellcolor{red!0} 0.1 ± 0.1 & \cellcolor{green!13} -3.4 ± 3.2 & \cellcolor{green!8} -2.2 ± 2.2 \\
\hline
\end{tabular}
}
\hspace{1em}
\subfloat[CLEAR]{ 
 \begin{tabular}{|l|lll|l|}
\hline
 & Hang TP & Counter & Cabinet & Avg ± SEM \\
\hline
Hang TP & \cellcolor{green!0} -- & \cellcolor{red!19} 4.8 ± 0.4 & \cellcolor{red!1} 0.3 ± 0.2 & \cellcolor{red!10} 2.5 ± 0.3 \\
Counter & \cellcolor{green!0} -- & \cellcolor{green!0} -- & \cellcolor{green!2} -0.5 ± 0.5 & \cellcolor{green!2} -0.5 ± 0.5 \\
Cabinet & \cellcolor{green!0} -- & \cellcolor{green!0} -- & \cellcolor{green!0} -- & \cellcolor{green!0} -- \\
\hline
Avg ± SEM & \cellcolor{green!0} -- & \cellcolor{red!19} 4.8 ± 0.4 & \cellcolor{green!0} -0.1 ± 0.2 & \cellcolor{red!6} 1.5 ± 0.2 \\
\hline
\end{tabular}
}

\vspace{2em}
\caption{CHORES: Mem-VaryTasks Forgetting metrics.}
\label{tab:chores_vary_tasks_forgetting}
\end{table}

\begin{table}[H]
\tablemarginhack
\setlength\tabcolsep{2pt} 
\tiny

\subfloat[EWC]{ 
 \begin{tabular}{|l|lll|l|}
\hline
 & Fork & Knife & Spoon & Avg ± SEM \\
\hline
Fork & \cellcolor{green!0} -- & \cellcolor{red!13} 3.4 ± 3.0 & \cellcolor{green!12} -3.2 ± 3.2 & \cellcolor{red!0} 0.1 ± 0.1 \\
Knife & \cellcolor{green!0} -- & \cellcolor{green!0} -- & \cellcolor{green!12} -3.2 ± 3.2 & \cellcolor{green!12} -3.2 ± 3.2 \\
Spoon & \cellcolor{green!0} -- & \cellcolor{green!0} -- & \cellcolor{green!0} -- & \cellcolor{green!0} -- \\
\hline
Avg ± SEM & \cellcolor{green!0} -- & \cellcolor{red!13} 3.4 ± 3.0 & \cellcolor{green!12} -3.2 ± 3.2 & \cellcolor{green!4} -1.0 ± 1.1 \\
\hline
\end{tabular}
}
\hspace{1em}
\subfloat[P\&C]{ 
 \begin{tabular}{|l|lll|l|}
\hline
 & Fork & Knife & Spoon & Avg ± SEM \\
\hline
Fork & \cellcolor{green!0} -- & \cellcolor{red!0} -0.0 ± 0.0 & \cellcolor{red!12} 3.1 ± 3.1 & \cellcolor{red!6} 1.5 ± 1.5 \\
Knife & \cellcolor{green!0} -- & \cellcolor{green!0} -- & \cellcolor{red!12} 3.0 ± 3.0 & \cellcolor{red!12} 3.0 ± 3.0 \\
Spoon & \cellcolor{green!0} -- & \cellcolor{green!0} -- & \cellcolor{green!0} -- & \cellcolor{green!0} -- \\
\hline
Avg ± SEM & \cellcolor{green!0} -- & \cellcolor{red!0} -0.0 ± 0.0 & \cellcolor{red!12} 3.0 ± 3.0 & \cellcolor{red!8} 2.0 ± 2.0 \\
\hline
\end{tabular}
}
\hspace{1em}
\subfloat[CLEAR]{ 
 \begin{tabular}{|l|lll|l|}
\hline
 & Fork & Knife & Spoon & Avg ± SEM \\
\hline
Fork & \cellcolor{green!0} -- & \cellcolor{red!26} 6.7 ± 3.3 & \cellcolor{red!1} 0.4 ± 6.0 & \cellcolor{red!14} 3.6 ± 1.8 \\
Knife & \cellcolor{green!0} -- & \cellcolor{green!0} -- & \cellcolor{red!11} 2.9 ± 2.9 & \cellcolor{red!11} 2.9 ± 2.9 \\
Spoon & \cellcolor{green!0} -- & \cellcolor{green!0} -- & \cellcolor{green!0} -- & \cellcolor{green!0} -- \\
\hline
Avg ± SEM & \cellcolor{green!0} -- & \cellcolor{red!26} 6.7 ± 3.3 & \cellcolor{red!6} 1.7 ± 1.9 & \cellcolor{red!13} 3.4 ± 0.2 \\
\hline
\end{tabular}
}

\vspace{2em}
\caption{CHORES: Mem-VaryObjects Forgetting metrics.}
\label{tab:chores_vary_objects_forgetting}
\end{table}

\begin{table}[H]
\tablemarginhack
\setlength\tabcolsep{2pt} 
\tiny

\subfloat[EWC]{ 
 \begin{tabular}{|l|lll|l|}
\hline
 & R19, Cup & R13, Potat.. & R02, Lettu.. & Avg ± SEM \\
\hline
R19, Cup & \cellcolor{green!0} -- & \cellcolor{red!7} 1.9 ± 3.2 & \cellcolor{green!0} -0.2 ± 0.1 & \cellcolor{red!3} 0.9 ± 1.5 \\
R13, Potat.. & \cellcolor{green!0} -- & \cellcolor{green!0} -- & \cellcolor{red!2} 0.5 ± 0.4 & \cellcolor{red!2} 0.5 ± 0.4 \\
R02, Lettu.. & \cellcolor{green!0} -- & \cellcolor{green!0} -- & \cellcolor{green!0} -- & \cellcolor{green!0} -- \\
\hline
Avg ± SEM & \cellcolor{green!0} -- & \cellcolor{red!7} 1.9 ± 3.2 & \cellcolor{red!0} 0.2 ± 0.1 & \cellcolor{red!2} 0.7 ± 1.2 \\
\hline
\end{tabular}
}
\hspace{1em}
\subfloat[P\&C]{ 
 \begin{tabular}{|l|lll|l|}
\hline
 & R19, Cup & R13, Potat.. & R02, Lettu.. & Avg ± SEM \\
\hline
R19, Cup & \cellcolor{green!0} -- & \cellcolor{red!0} 0.0 ± 0.1 & \cellcolor{green!0} -0.2 ± 0.1 & \cellcolor{green!0} -0.1 ± 0.1 \\
R13, Potat.. & \cellcolor{green!0} -- & \cellcolor{green!0} -- & \cellcolor{green!0} -0.1 ± 0.1 & \cellcolor{green!0} -0.1 ± 0.1 \\
R02, Lettu.. & \cellcolor{green!0} -- & \cellcolor{green!0} -- & \cellcolor{green!0} -- & \cellcolor{green!0} -- \\
\hline
Avg ± SEM & \cellcolor{green!0} -- & \cellcolor{red!0} 0.0 ± 0.1 & \cellcolor{green!0} -0.1 ± 0.1 & \cellcolor{green!0} -0.1 ± 0.1 \\
\hline
\end{tabular}
}
\hspace{1em}
\subfloat[CLEAR]{ 
 \begin{tabular}{|l|lll|l|}
\hline
 & R19, Cup & R13, Potat.. & R02, Lettu.. & Avg ± SEM \\
\hline
R19, Cup & \cellcolor{green!0} -- & \cellcolor{red!5} 1.3 ± 1.3 & \cellcolor{green!5} -1.3 ± 1.2 & \cellcolor{red!0} -0.0 ± 1.0 \\
R13, Potat.. & \cellcolor{green!0} -- & \cellcolor{green!0} -- & \cellcolor{green!3} -0.9 ± 4.6 & \cellcolor{green!3} -0.9 ± 4.6 \\
R02, Lettu.. & \cellcolor{green!0} -- & \cellcolor{green!0} -- & \cellcolor{green!0} -- & \cellcolor{green!0} -- \\
\hline
Avg ± SEM & \cellcolor{green!0} -- & \cellcolor{red!5} 1.3 ± 1.3 & \cellcolor{green!4} -1.1 ± 2.9 & \cellcolor{green!1} -0.3 ± 2.1 \\
\hline
\end{tabular}
}

\vspace{2em}
\caption{CHORES: Gen-MultiTraj Forgetting metrics.}
\label{tab:chores_multi_traj_forgetting}
\end{table}

\clearpage

\subsection{CHORES Metrics: Transfer}
\label{sec:chores_transfer}

\begin{table}[H]
\tablemarginhack
\setlength\tabcolsep{2pt} 
\tiny

\subfloat[EWC]{ 
 \begin{tabular}{|l|lll|l|}
\hline
 & R402 & R419 & R423 & Avg ± SEM \\
\hline
R402 & \cellcolor{green!0} -- & \cellcolor{green!0} -- & \cellcolor{green!0} -- & \cellcolor{green!0} -- \\
R419 & \cellcolor{red!0} 0.0 ± 0.0 & \cellcolor{green!0} -- & \cellcolor{green!0} -- & \cellcolor{red!0} 0.0 ± 0.0 \\
R423 & \cellcolor{red!0} -0.1 ± 0.1 & \cellcolor{red!0} 0.0 ± 0.0 & \cellcolor{green!0} -- & \cellcolor{red!0} -0.1 ± 0.0 \\
\hline
Avg ± SEM & \cellcolor{red!0} -0.1 ± 0.0 & \cellcolor{red!0} 0.0 ± 0.0 & \cellcolor{green!0} -- & \cellcolor{red!0} -0.0 ± 0.0 \\
\hline
\end{tabular}
}
\hspace{1em}
\subfloat[P\&C]{ 
 \begin{tabular}{|l|lll|l|}
\hline
 & R402 & R419 & R423 & Avg ± SEM \\
\hline
R402 & \cellcolor{green!0} -- & \cellcolor{green!0} -- & \cellcolor{green!0} -- & \cellcolor{green!0} -- \\
R419 & \cellcolor{green!19} 4.8 ± 2.9 & \cellcolor{green!0} -- & \cellcolor{green!0} -- & \cellcolor{green!19} 4.8 ± 2.9 \\
R423 & \cellcolor{green!19} 4.8 ± 2.9 & \cellcolor{red!0} 0.0 ± 0.0 & \cellcolor{green!0} -- & \cellcolor{green!9} 2.4 ± 1.4 \\
\hline
Avg ± SEM & \cellcolor{green!19} 4.8 ± 2.9 & \cellcolor{red!0} 0.0 ± 0.0 & \cellcolor{green!0} -- & \cellcolor{green!12} 3.2 ± 1.9 \\
\hline
\end{tabular}%
}
\vspace{2em}
\caption{CHORES: Mem-VaryRoom Transfer metrics.}
\label{tab:chores_vary_env_transfer}
\end{table}

\begin{table}[H]
\tablemarginhack
\setlength\tabcolsep{2pt} 
\tiny

\subfloat[EWC]{ 
 %
}

\vspace{2em}
\caption{CHORES: Mem-VaryTasks Transfer metrics.}
\label{tab:chores_vary_tasks_transfer}
\end{table}

\begin{table}[H]
\tablemarginhack
\setlength\tabcolsep{2pt} 
\tiny

\subfloat[EWC]{ 
 %
}

\vspace{2em}
\caption{CHORES: Mem-VaryObjects Transfer metrics.}
\label{tab:chores_vary_objects_transfer}
\end{table}

\begin{table}[H]
\tablemarginhack
\setlength\tabcolsep{2pt} 
\tiny

\subfloat[EWC]{ 
 %
}

\vspace{2em}
\caption{CHORES: Gen-MultiTraj Transfer metrics.}
\label{tab:chores_multi_traj_transfer}
\end{table}

\end{document}